\documentclass[preprint,11pt]{elsarticle}
\usepackage{times}
\usepackage{url}
\usepackage[hidelinks]{hyperref}
\usepackage[small]{caption}
\usepackage{booktabs}
\usepackage{multirow}
\usepackage{colortbl}
\usepackage{xspace}
\usepackage{bm}

\usepackage[utf8]{inputenc}
\usepackage[english]{babel}

\makeatletter
\def\ps@pprintTitle{%
  \let\@oddhead\@empty
  \let\@evenhead\@empty
  \def\@oddfoot{}%
  \let\@evenfoot\@oddfoot}
\makeatother

\newcommand{\citepossessive}[1]{\citeauthor{#1}'s [\citeyear{#1}]}

\newcommand{\fbf}{\textbf}

\newcommand{\fsl}{\textsl}

\usepackage{array}
\newcolumntype{H}{>{\setbox0=\hbox\bgroup}c<{\egroup}@{}}
\newcolumntype{R}[1]{>{\raggedright\arraybackslash}p{#1}}
\usepackage{subcaption}

\usepackage{float}
\usepackage{placeins}
\usepackage{graphicx}

\usepackage{siunitx} 
\newcommand{\qtyP}[1]{\qty{#1}{\percent}}

\usepackage[normalem]{ulem}
\newcommand{\textinred}[1]{#1}

\newcommand{\ssig}{\textcolor{green!60!black}{\faCheck}}
\newcommand{\snos}{\textcolor{orange!80!black}{$\ddagger$}}
\newcommand{\snose}{\textcolor{orange!80!black}{$\dagger$}}
\newcommand{\sig}[1]{\textcolor{green!60!black}{\faCheck}\;\;\textcolor{black}{\small(#1)}}
\newcommand{\sigv}[1]{\textcolor{green!60!black}{\faCheck}\;\;\textcolor{black}{\footnotesize(#1)}}
\newcommand{\nosig}{\textcolor{red!60!black}{\faTimes}}
\newcommand{\nos}[1]{{\;\textcolor{orange!80!black}{$\ddagger$}}\;\;\;\textcolor{black}{\small(#1)}}
\newcommand{\nose}[1]{{\;\textcolor{orange!80!black}{$\dagger$}}\;\;\;\textcolor{black}{\small(#1)}}
\newcommand{\whitecell}[1]{\cellcolor{white}#1}

\usepackage{standalone}
\usepackage{tikz}
\usetikzlibrary{positioning}
\usetikzlibrary{decorations.pathreplacing}
\usetikzlibrary{arrows.meta}
\usetikzlibrary{fit}
\usetikzlibrary{backgrounds}

\usepackage{xstring}
\usepackage{fontawesome5}
\usepackage{pgfplots}
\pgfplotsset{compat=newest}
\usetikzlibrary{calc}
\usetikzlibrary{shapes}
\usetikzlibrary{external}

\definecolor{ncsuBlue}{RGB}{0,142, 203} 
\definecolor{ncsuTeal}{RGB}{0, 176, 183} 
\definecolor{ncsuRed}{RGB}{199, 53, 45} 
\definecolor{ncsuOrange}{RGB}{231, 135, 35} 
\definecolor{ncsuGreen}{RGB}{124, 185, 43} 
\definecolor{ncsuPurple}{RGB}{141, 109, 172} 
\definecolor{ncsuYellow}{RGB}{237, 179, 31} 
\definecolor{ncsuGray}{RGB}{134, 135, 139} 
\definecolor{ncsuLightBlue}{RGB}{152, 221, 243} 
\definecolor{ncsuLightTeal}{RGB}{172, 223, 223} 
\definecolor{ncsuLightRed}{RGB}{247, 203, 190} 
\definecolor{ncsuLightOrange}{RGB}{242, 211,159} 
\definecolor{ncsuLightGreen}{RGB}{212, 230, 173} 
\definecolor{ncsuLightPurple}{RGB}{225, 205, 227} 
\definecolor{ncsuLightYellow}{RGB}{249, 223, 139} 
\definecolor{ncsuLightGray}{RGB}{205, 206, 211} 
\definecolor{ncsuDarkBlue}{RGB}{33, 70, 109} 
\definecolor{ncsuDarkTeal}{RGB}{40, 92, 103} 
\definecolor{ncsuDarkRed}{RGB}{141, 28, 36} 
\definecolor{ncsuDarkOrange}{RGB}{158, 77, 32} 
\definecolor{ncsuDarkGreen}{RGB}{71, 99, 51} 
\definecolor{ncsuDarkPurple}{RGB}{86, 52, 127} 
\definecolor{ncsuDarkYellow}{RGB}{152, 124, 51} 
\definecolor{ncsuDarkGray}{RGB}{60, 61, 65} 

\usepackage[inline]{enumitem}

\usepackage{xcolor}
\usepackage{soul}

\newcounter{textboxno}
\usepackage[many]{tcolorbox}
\newtcolorbox{mybox}[2][]{%
boxsep=3pt,left=2pt,right=2pt,bottom=5pt,
width=\textwidth,
boxrule=1pt,
attach boxed title to top center = {yshift=-\tcboxedtitleheight/2},
colbacktitle=white,coltitle=black,
boxed title style={size=normal,colframe=white,boxrule=0pt}, 
interior style={white},
title={\refstepcounter{textboxno}\label{#1}
Example \arabic{textboxno}: {#2}
\def\@currentlabel{\p@textboxno\thetextboxno}},
enhanced,
float,
}

\definecolor{ncsuBlue}{RGB}{0,142, 203} 
\definecolor{ncsuTeal}{RGB}{0, 176, 183} 
\definecolor{ncsuRed}{RGB}{199, 53, 45} 
\definecolor{ncsuOrange}{RGB}{231, 135, 35} 
\definecolor{ncsuGreen}{RGB}{124, 185, 43} 
\definecolor{ncsuPurple}{RGB}{141, 109, 172} 
\definecolor{ncsuYellow}{RGB}{237, 179, 31} 
\definecolor{ncsuGray}{RGB}{134, 135, 139} 
\definecolor{ncsuLightBlue}{RGB}{152, 221, 243} 
\definecolor{ncsuLightTeal}{RGB}{172, 223, 223} 
\definecolor{ncsuLightRed}{RGB}{247, 203, 190} 
\definecolor{ncsuLightOrange}{RGB}{242, 211,159} 
\definecolor{ncsuLightGreen}{RGB}{212, 230, 173} 
\definecolor{ncsuLightPurple}{RGB}{225, 205, 227} 
\definecolor{ncsuLightYellow}{RGB}{249, 223, 139} 
\definecolor{ncsuLightGray}{RGB}{205, 206, 211} 
\definecolor{ncsuDarkBlue}{RGB}{33, 70, 109} 
\definecolor{ncsuDarkTeal}{RGB}{40, 92, 103} 
\definecolor{ncsuDarkRed}{RGB}{141, 28, 36} 
\definecolor{ncsuDarkOrange}{RGB}{158, 77, 32} 
\definecolor{ncsuDarkGreen}{RGB}{71, 99, 51} 
\definecolor{ncsuDarkPurple}{RGB}{86, 52, 127} 
\definecolor{ncsuDarkYellow}{RGB}{152, 124, 51} 
\definecolor{ncsuDarkGray}{RGB}{60, 61, 65} 
\definecolor{ncsuGold}{RGB}{255, 184, 28} 
\definecolor{ncsuBanana}{RGB}{254,221,0} 
\definecolor{barDarkBlue}{RGB}{0,0,255} 
\definecolor{barLightBlue}{RGB}{179,179,225} 
\definecolor{barDarkRed}{RGB}{255,0,0} 
\definecolor{barLightRed}{RGB}{255,179,179} 
\definecolor{barDarkGreen}{RGB}{0,255,0} 
\definecolor{barLightGreen}{RGB}{179,255,179} 
\definecolor{pumpkin}{RGB}{255, 116, 23} 
\definecolor{mint}{RGB}{16, 185, 129} 
\definecolor{GoodGreen}{RGB}{76, 175, 80}
\definecolor{EvilRed}{RGB}{229, 57, 53}
\definecolor{PhaseBlue}{RGB}{232, 241, 250}
\definecolor{PhaseYellow}{RGB}{255, 253, 231}
\definecolor{PhaseRed}{RGB}{255, 235, 238}
\definecolor{LeaderGold}{RGB}{255, 215, 0}

\newtcolorbox[auto counter, number within=section]{promptbox}[3][]{
  colback=gray!10, 
  colframe=black, 
  boxrule=0.5mm, 
  sharp corners=south, 
  width=\textwidth, 
  fontupper=\color{black}\footnotesize,
  left=4pt, 
  right=3pt,
  boxsep=0pt,
  title={\faIcon{#2}~#3}, 
  coltitle=white, 
  colbacktitle=black, 
  fonttitle=\bfseries\color{white}\small,
  toptitle=2pt,
  bottomtitle=2pt,
  #1
}

\usepackage{fancyvrb}
\DefineVerbatimEnvironment{MyVerbatim}{Verbatim}
  {fontsize=\small, baselinestretch=0.9, fontfamily=txtt}

\newcommand{\fqt}[1]{``#1''}

\journal{Artificial Intelligence}

\begin{document}

\begin{frontmatter}

\title{Theory of Mind in Action: The Instruction Inference Task in Dynamic Human-Agent Collaboration}

\author[ncsu]{Fardin Saad}
\author[tud]{Pradeep K. Murukannaiah}
\author[ncsu]{Munindar P. Singh}

\affiliation[ncsu]{organization={North Carolina State University},
            city={Raleigh},
            state={NC},
            country={USA}
}
\affiliation[tud]{organization={Delft University of Technology},
            city={Delft},
            country={The Netherlands}
}

\begin{abstract}        
\textinred{Successful human-agent teaming relies on an \emph{agent} being able to understand instructions given by a (human) \emph{principal}. 
In many cases, an instruction may be incomplete or ambiguous. 
In such cases, the agent must infer the unspoken intentions from their shared context, that is, it must exercise the principal's Theory of Mind (ToM) and infer the mental states of its principal.} 
We consider the prospects of effective human-agent collaboration using large language models (LLMs). 
To assess ToM in a dynamic, goal-oriented, and collaborative environment, we introduce a novel task, \emph{Instruction Inference}, in which an \emph{agent} assists a \emph{principal} in reaching a goal by interpreting incomplete or ambiguous instructions. 

We present \emph{Tomcat}, an LLM-based agent, designed to exhibit ToM reasoning in interpreting and responding to the principal's instructions.
We implemented two variants of Tomcat. 
One, dubbed Fs-CoT \textinred{(Fs for few-shot, CoT for chain-of-thought)}, is based on a small number of examples demonstrating the requisite structured reasoning.
One, dubbed CP \textinred{(commonsense prompt)}, relies on commonsense knowledge and information about the problem.
We realized both variants of Tomcat on three leading LLMs, namely, GPT-4o, DeepSeek-R1, and Gemma-3-27B. 
To evaluate the effectiveness of Tomcat, we conducted a study with 52 human participants in which we provided participants with the same information as the CP variant. 
We computed intent accuracy, action optimality, and planning optimality to measure the ToM capabilities of Tomcat and our study participants. We found that Tomcat with Fs-CoT, particularly with GPT-4o and DeepSeek-R1, achieves performance comparable to the human participants, underscoring its ToM potential for human-agent collaboration.
\end{abstract}


\end{frontmatter}
    
\section{Introduction}
\label{sec:introduction}
\textinred{Natural communication is rarely limited to literal language \cite{Grice75,Searle75}. Instead, collaborators frequently use indirect expressions whose intended meaning must be inferred from contextual cues and shared task understanding \cite{Zhi-Xuan+24:CLIPS, AAMAS-25:Gricean, Qian+24:tell-me-more}.} 
Successful collaboration \textinred{therefore} presupposes an ability to interpret \textinred{incomplete} and ambiguous language, which in turn presupposes an understanding of the \textinred{collaborator's} mental state \cite{Tomasello+05:sharing-intentions,Duranti-15:intentions}.
The capability of modeling another entity as endowed with a \emph{mind}, e.g., beliefs and intentions, is referred to as having a Theory of Mind (ToM) \cite{Premack+Woodruff-78:ToM-chimps, Frith-05:ToM}. 
ToM enables people to infer unspoken intentions and respond appropriately to indirect cues \cite{Erdogan+25:TOMA-hybrid-intelligence}. 
\begin{figure}[h]
    \centering
    \includestandalone[width=\linewidth, mode=image]{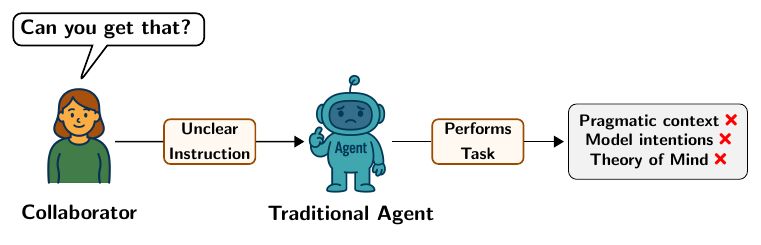}
    \vspace{1em}   
    \includestandalone[mode=image,width=\linewidth]{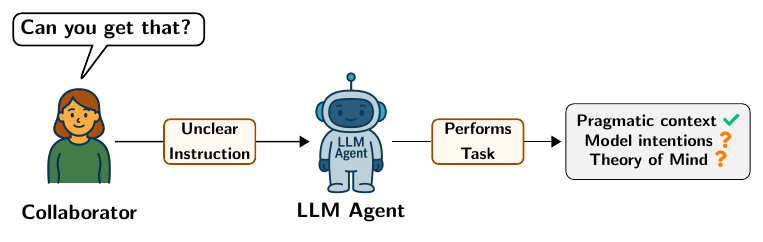}
    \caption{\textinred{Interpreting unclear instructions by traditional vs. LLM agents where the latter performs the task considering the pragmatic context.}}
    \label{fig:motivation}
\end{figure}

 \textinred{For instance, a collaborator may say \fsl{Can you get that?} while pointing to an object (e.g., a key) in the shared environment.
 Interpreting such an instruction requires inferring which object was requested based on contextual cues and shared task understanding, thereby requiring ToM reasoning. 
 At the first-order ToM level, this may involve reasoning such as \fsl{I believe that you want the red key}. 
In more complex situations, collaborators may engage in higher-order reasoning, i.e., second-order ToM, where one reasons about another party's beliefs, for instance, \fsl{I believe that you believe that I know which key is needed}.} 
\textinred{Such higher-order reasoning is often observed in strategic interactions involving persuasion, negotiation, or deception \cite{Chong+Ho+Camerer-16:CognitiveHierarchy, DeWeerd+22:Hi-ToM-Negotiation, Sarkadi-23:ArmsRaceToM}.
However, in cooperative settings, resolving unclear instructions often relies on reasoning about the collaborator's immediate goals and intentions, which can be captured through first-order ToM \cite{AAMAS-25:Gricean, Tomasello+05:sharing-intentions, Wu+21:TooManyCooks}.}
 
Equipping AI agents with ToM can foster \textinred{more} natural and effective human-agent interaction \cite{Richards+24:Code-ToM}, \textinred{particularly in settings where communication is incomplete or ambiguous \cite{AAMAS-25:Gricean}}.
\textinred{However, traditional autonomous agents have historically struggled to interpret such language.
Prior work in instruction following and human-agent collaboration shows that agents lacking pragmatic context \cite{Wan+22:HandMeThat, Singh+23:ProgPrompt, Ichter+23:Robotic-Affordance} tend to interpret instructions literally and thus fail when instructions are incomplete or ambiguous \cite{Zhi-Xuan+24:CLIPS, Qian+24:tell-me-more}. 
As a result, these agents often cannot infer a collaborator's intended goal from contextual cues in dynamic task environments, as shown in Figure~\ref{fig:motivation}.}

\textinred{The advent of large language models (LLMs) offers a promising avenue to bridge this gap. 
LLMs have demonstrated an emergent capability to account for pragmatic context and perform sophisticated linguistic reasoning \cite{Strachan+24:testing-ToM-LLM, Kosinski-24:evaluating-ToM-LLMs, Gandhi+23:BigToM, Wu+23:Hi-ToM}.
Recent studies suggest that these models may exhibit ToM capabilities by modeling the mental states of others \citep{Strachan-24:ToM-LLM-humans, Kosinski-24:evaluating-ToM-LLMs}.
However, evaluations of the ToM capabilities of LLMs rely on static benchmarks such as false-belief, faux-pas, or conversational reasoning tasks.} 
For example, \citet{Strachan+24:testing-ToM-LLM} measure the ToM capabilities of LLMs by providing a short conversational vignette followed by targeted questions to assess how LLMs reason about another party's mental states.
\textinred{These settings evaluate reasoning about mental states but do not test whether agents can use such reasoning to interpret \emph{unclear instructions} and act appropriately in \emph{collaborative task environments}, as shown by the contrast in Figure~\ref{fig:motivation}.}

\textinred{To address this challenge, we introduce a new ToM task, \textbf{\emph{Instruction Inference}}, consisting of 20 distinct problem scenarios.
Each scenario evaluates an agent's capability to infer a principal's (simulated human) intended goal from clear and unclear instructions and act appropriately in a collaborative environment.} 
We use the term \emph{instructions} to refer to commands, requests, or directives issued by \textinred{the principal}.
\textinred{
To ground the design of our dataset, we draw on Grice's theory of conversational implicature \cite{Grice75}.
Grice identifies that cooperative communication is guided by four conversational maxims: Quantity, Quality, Relation, and Manner. 
Violations of these maxims trigger pragmatic inference, requiring listeners to infer the speaker's intended meaning beyond the literal utterance.
We distinguish two types of unclear instructions grounded in this framework.
\emph{Incomplete} instructions omit critical information required to determine the intended action; they correspond to violations of Quantity maxim.
\emph{Ambiguous} instructions yield two or more possible interpretations; they correspond to violations of Manner maxim. 
We focus on these two categories because violations of Quantity and Manner maxims commonly arise in cooperative settings where collaborators assume shared context and task understanding \cite{Zhi-Xuan+24:CLIPS, AAMAS-25:Gricean, Qian+24:tell-me-more}. 
By contrast, violations of the Quality and Relation maxims are less characteristic of cooperative collaboration, as they typically correspond to deception or conversational breakdown.}

\begin{figure}
    \centering
    \includegraphics[width=\linewidth, trim=3.5cm 5.2cm 2.5cm 3.5cm, clip]{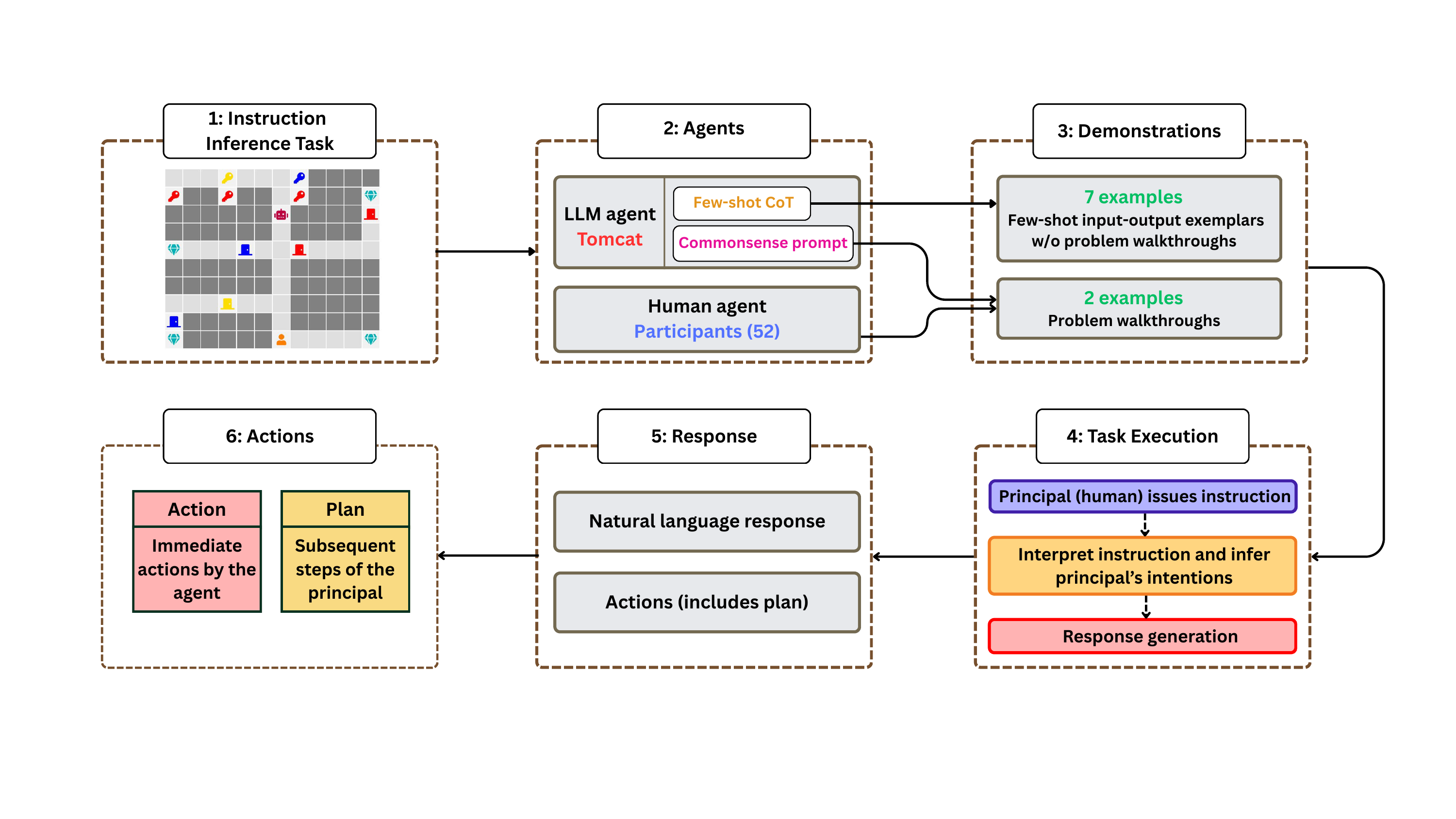}
    \caption{Pipeline of the Instruction Inference task with Tomcat and participants.}
    \label{fig:flow-tomcat}
\end{figure}

We \textinred{implement this task using} \emph{Tomcat}, an LLM-based agent deployed in a collaborative environment called Doors, Keys, and Gems \cite{Zhi-Xuan+24:CLIPS}, shown in Figure~\ref{fig:doors-keys-gems}. 
\textinred{In this environment,} Tomcat assists its principal in achieving their \textinred{objective of collecting a gem by retrieving keys or unlocking doors.
Tomcat interprets both clear and unclear instructions by reasoning about the task environment and the principal's movements.}
Our study employs two variants of Tomcat, \emph{Few-shot Chain-of-thought (Fs-CoT) Prompt} and \emph{Commonsense Prompt (CP)} (Figure~\ref{fig:flow-tomcat}).
\textinred{Tomcat with Fs-CoT leverages} in-context learning \cite{Brown+20:few-shot-learners} and few-shot chain-of-thought (Fs-CoT) prompting \cite{Wei+22:CoT-reasoning} \textinred{to incorporate task knowledge and contextual cues to infer the principal's intended goal}. 
\textinred{Tomcat with CP relies on commonsense knowledge about the Instruction Inference task with two detailed problem walkthroughs.}

\textinred{We realized both variants of Tomcat on three LLMs---GPT-4o, DeepSeek-R1, and Gemma-3-27B---and compare their ToM capabilities with human participants. 
To establish a benchmark for comparison, we conduct an empirical study with 52 human participants who perform the same Instruction Inference task} under identical conditions as Tomcat with CP, i.e., using the same information and objectives.
This evaluation is framed by the following research questions.

\begin{description}
    \item [RQ$_1$:] How do Tomcat's ToM capabilities on Instruction Inference compare to human performance? 
 
    \item [RQ$_2$:] \textinred{Does Tomcat with Fs-CoT outperform Tomcat with CP on Instruction Inference?} 

\end{description}

\textinred{To summarize, our \textbf{contributions} are as follows:}
\begin{enumerate}

\item \textbf{Instruction Inference task}
We introduce Instruction Inference, a novel task for evaluating ToM in a collaborative environment where instructions may be clear or unclear (incomplete or ambiguous).
Unlike existing ToM benchmarks that rely primarily on narrative or conversational scenarios, Instruction Inference evaluates whether an agent can infer a principal's intended goal from clear and unclear instructions and act appropriately in a shared task environment.

\item \textbf{Tomcat agent for instruction inference.}
We develop \emph{Tomcat}, an LLM-based agent that performs instruction inference in a dynamic environment by reasoning about pragmatic context and the principal's actions.
We investigate two variants of Tomcat: a \emph{Few-shot Chain-of-Thought (Fs-CoT)} variant using seven reasoning exemplars, and a \emph{Commonsense Prompt (CP)} variant that relies on task knowledge with two problem walkthroughs.

\item \textbf{Empirical evaluation with human participants.}
We conduct an empirical evaluation comparing both Tomcat variants' ToM capabilities, realized on three LLMs (GPT-4o, DeepSeek-R1, and Gemma-3-27B), with those of 52 human participants on our Instruction Inference task.
Our results show that the Fs-CoT variants of GPT-4o and DeepSeek-R1 achieve performance comparable to human participants.

\end{enumerate}
The subsequent sections provide a systematic analysis of each module in our pipeline, as shown in Figure~\ref{fig:flow-tomcat}.

\section{Related Work}
\label{sec:Related}
\paragraph{\textbf{\textinred{ToM Benchmarks}}} Building on the recent success in LLMs, \citet{Strachan+24:testing-ToM-LLM} evaluate models like GPT-4, GPT-3.5, and Llama2-70B on ToM tasks including false-beliefs, faux pas, irony, strange stories, and hinting tasks. 
The tasks adopt practical conversational social scenarios in contrast to our goal-driven collaborative setting. 
\citet{Strachan+24:testing-ToM-LLM} report that GPT-4 outperforms humans in irony, hinting, and strange stories, but not in faux pas. 
Similarly, \citet{Kosinski-24:evaluating-ToM-LLMs} tests foundational ToM concepts, such as belief attribution and reasoning about mental states, using structured cognitive frameworks, revealing that advanced LLMs like GPT-4 exhibit human-like reasoning in certain tasks but not in nuanced scenarios. 
\textinred{\citet{Kim+23:FANToM} introduce \emph{FANToM}, a conversational benchmark that evaluates mental state reasoning under asymmetric information. 
\citet{Chan+24:NegotiationToM} present \emph{NegotiationToM}, a benchmark grounded in the belief-desire-intention (BDI) framework that evaluates ToM reasoning in negotiation dialogues derived from the CaSiNo dataset \cite{Chawla+21:CaSiNo}.
\citet{Gandhi+23:BigToM} introduce BigToM, a large scale benchmark that extends classical narrative-based ToM tasks to evaluate belief reasoning across diverse social scenarios. 
}

\paragraph{\textbf{\textinred{Higher-order ToM Reasoning}}} \citet{Street-24:llm-ToM} posits that LLMs' ToM capabilities can help improve human-agent collaboration. 
\citet{Street+24:LLM-HO-ToM} develop the Multi-Order Theory of Mind Q\&A (MoToMQA) test suite to evaluate LLMs on ToM tasks that require recursive reasoning about nested mental states (second- to sixth-order). 
They show GPT-4's capability to handle the deep recursive reasoning that is critical for understanding complex social interactions. 
\citet{Jamali+23:LLM-human-ToM} analyze hidden embeddings in LLMs and identify neuron-like units responsive to true- and false-belief tasks, drawing parallels to ToM processing \textinred{by humans}.

\paragraph{\textbf{\textinred{ToM in Multiagent Systems}}} \textinred{Recent work has expanded ToM into more complex domains, such as multiagent systems, automated planning, and multimodal environments. 
\citet{Rocha+23:ToM-MAS-Review} review the use of ToM in multiagent systems, noting its value for improving coordination, generating explanations, providing assistance, and modeling deception.
In planning, \citet{Shvo+22:Misconceptions-ToM-Plan} use ToM to help agents identify conflicting beliefs about a plan's validity and correct these misunderstandings through direct communication or physical actions.
\citet{Shi+25:MuMA-ToM} introduce MuMA-ToM, a multimodal multiagent benchmark for inferring an agent's beliefs and social goals in everyday household activities using video and textual context. 
\citet{Yim+24:Guandan-ToM} adopt the cooperative card game \emph{Guandan} and show that ToM-based planning helps LLM agents better coordinate with allies and adapt strategies against opponents.} 

\begin{table}[htb]
\centering
\rowcolors{2}{orange!20}{white}
\resizebox{\textwidth}{!}{
\begin{tabular}{R{0.3\textwidth} R{0.25\textwidth} R{0.13\textwidth} R{0.20\textwidth} p{0.17\textwidth} R{0.20\textwidth}}
\toprule
\textbf{ToM Task} &
\textbf{Format} &
\textbf{Modality} &
\textbf{Collaborative} &
\textbf{Instruction} &
\textbf{Scenario} \\
\midrule
False-Belief \cite{Strachan+24:testing-ToM-LLM, Kosinski-24:evaluating-ToM-LLMs, Jamali+23:LLM-human-ToM} & Narrative QA & Text & No & -- & Social vignette \\

Faux Pas, Irony, Hinting \cite{Strachan+24:testing-ToM-LLM} & Narrative QA & Text & No & -- & Social vignette \\

RMET (Reading the Mind in Eye Test) \cite{Strachan+24:GPT-4o-Read} &
Visual QA & Visual + Text & No & -- & Visual vignette \\

SocialIQa, ToMi \cite{Sap+22:LLM-neural-ToM} &
Multiple choice QA & Text & No & -- & Social vignette\\

Altered ToM \cite{Ullman-23:LLM-alter-ToM} & Narrative QA & Text & No & -- & Social vignette \\

Robot Intent Classification \cite{Verma+24:LLM-HRI-ToM} & Nested QA & Text & No & -- & Robot behavior \\

ChangeMyView \cite{Amirizaniani+24:LLM-likehuman-ToM} & Open-ended QA & Text & No & -- & Social vignette \\


MoToMQA (Multi-Order ToM Q\&A) \cite{Street+24:LLM-HO-ToM} &
Nested QA & Text & No & -- & Social vignette  \\

FANToM \cite{Kim+23:FANToM} & Conversational QA & Text & No & -- & Conversational interaction \\

NegotiationToM \cite{Chan+24:NegotiationToM} & Multiple choice QA & Text & No & -- & Negotiation dialogue \\

BigToM \cite{Gandhi+23:BigToM} & Narrative QA & Text & No & -- & Social vignette \\

Hi-ToM \cite{Wu+23:Hi-ToM} & Nested QA & Text & No & -- & Social vignette \\

ToM Text Games \cite{Li+23:LLM-MAC-ToM} &
Text based game & Text & Yes (Agent--agent) & Explicit & Cooperative gameplay \\

MuMA-ToM \cite{Shi+25:MuMA-ToM} &
Multimodal QA & Video + Text & Yes (Agent--agent) & Explicit & Household interaction \\

Guandan ToM \cite{Yim+24:Guandan-ToM} &
Card game & Text & Yes (Agent--agent) & Explicit & Strategic gameplay \\

Shared Workspace Task \cite{Zhang+24:LLM-workspace-ToM} & Action execution & Text & Yes (Human--agent) & Explicit & Collaborative gameplay \\

Code-ToM Assistant \cite{Richards+24:Code-ToM} &
Interactive assistant & Text & Yes (Human--agent) & Explicit & Code comprehension \\

\textbf{\emph{Instruction Inference} (Ours)} & Plan and action execution & Text + Spatial Grid & Yes (Human--agent) & Incomplete or Ambiguous & Collaboration under uncertainty \\

\bottomrule
\end{tabular}
}
\caption{\textinred{Comparison of ToM benchmarks across format, modality, collaboration setting, and instruction type.}
}
\label{tab:ToM_tasks_comparison}
\end{table}

\paragraph{\textbf{\textinred{ToM in Multiagent Reinforcement Learning}}} \textinred{\citet{Oguntola+23:ToM-Intrinsic-MARL} explore ToM as an intrinsic reward signal for multiagent reinforcement learning (MARL) through second-order belief prediction and find success in improving multiagent collaboration in mixed cooperative-competitive environments.} 
\textinred{\citet{Li+23:LLM-MAC-ToM} compare LLMs' ToM capabilities against MARL and planning in multiagent text-based games. They} demonstrate emergent collaborative behaviors and higher-order ToM reasoning in LLMs and highlight challenges due to long-horizon planning and belief tracking. 

\paragraph{\textbf{\textinred{Limitations of LLMs for ToM}}} \textinred{\citet{Zhang+24:LLM-workspace-ToM} find that ToM capabilities improve understanding, but do not enhance overall team performance in a real-time shared workspace task. 
\citet{Amirizaniani+24:LLM-likehuman-ToM} using open-ended questions derived from Reddit's \emph{ChangeMyView} subreddit, note that prompt-tuning of GPT-3.5 and GPT-4 improves performance but that LLMs do not achieve human-like reasoning in nuanced social contexts.}
\citet{Sap+22:LLM-neural-ToM} evaluate GPT-3's ToM capabilities using the SocialIQa and ToMi (Theory of Mind Inventory) tasks, highlighting performance significantly lower than humans and emphasizing the limitations of LLMs in social intelligence. 
Extending this line of inquiry, \citet{Verma+24:LLM-HRI-ToM} examine GPT-3.5 and GPT-4 in Human-Robot Interaction tasks. Performance declines significantly under minor contextual perturbations, suggesting that LLMs rely on superficial reasoning rather than robust ToM capabilities. 
Similarly, \citet{Ullman-23:LLM-alter-ToM} demonstrates that GPT-3 and GPT-3.5 struggle with false-belief tasks when trivial alterations, such as rephrasing or irrelevant context, are introduced, indicating their dependence on surface-level patterns. 
However, \citet{Strachan+24:testing-ToM-LLM} report that humans \emph{also} fail perturbed false-belief tasks, possibly due to their tendency to maintain original beliefs despite contradictory evidence. 

\paragraph{\textbf{\textinred{Instruction Inference (Our Task)}}} Table~\ref{tab:ToM_tasks_comparison} contrasts these benchmarks across key dimensions and shows that our Instruction Inference task uniquely combines human-agent, goal-driven collaboration in a changing environment with inherently \textinred{incomplete and} ambiguous instructions.

\section{Instruction Inference}
\label{sec:instruction-inference}
\textinred{This section describes the task setting and terminology used in our study, the Doors, Keys, and Gems environment, the design rationale behind the instruction types used in our dataset, and the pipeline through which agents interpret instructions and generate actions.}

\subsection{Task Setting and Terminology}
\label{sec:task-terminology}
The Instruction Inference task helps assess an agent's ToM capabilities in a dynamic, goal-oriented environment involving human-agent collaboration. 
The agent executes actions by reasoning about the principal's mental model based on their instructions \cite{Yao+23:ReAct-LLMs}.
The task involves two roles: a \emph{principal} who issues instructions and an \emph{agent} that interprets the instructions and performs actions. 
In the environment, the principal represents a simulated human collaborator whose movements and instructions provide contextual cues about their intended goal. 

In our empirical evaluation, human participants assume the role of the \emph{agent}. 
Thus, participants must interpret the principal's instructions and perform actions accordingly in the environment. 
Since the study involves two distinct human roles, we explicitly distinguish between them:
\begin{enumerate}[label=(\roman*), itemsep=2pt]
    \item \textbf{Human as Principal:} The simulated human principal that appears in the environment (prompts) and issues instructions.
    \item \textbf{Human as Participant (Agent):} A human study participant who assumes the role of the agent during evaluation.
\end{enumerate}
This distinction clarifies the use of the term \fqt{human} in the prompts and environment descriptions, where it refers to the principal rather than the participants.
The principal is simulated so that both human participants and LLM-based agents interact with the same instructions and environment under identical task conditions.
\begin{figure}[!htb]
\subcaptionsetup{width=0.9\linewidth}
\begin{subfigure}[t]{0.32\textwidth}
\centering
\includestandalone[width=0.99\linewidth]{tikz/grid-world}
\caption{Initial state of our scenario. Dark gray cells are walls and light gray cells are passages; the icons are a principal, an agent, doors, keys, and gems.}
\label{fig:dkg_1}
\end{subfigure}
\begin{subfigure}[t]{0.32\textwidth}
\centering
\newcommand*{\makeCell}[2]{%
    \IfEqCase{#2}{%
        {m}{\node at #1 [cellNode,fill=gray!10, text=purple] {\faRobot};}
        {h}{\node at #1 [cellNode, fill=gray!10, text=orange] {\faUser};}%
        {g}{\node at #1 [cellNode, fill=gray!10, text=ncsuTeal] {\faIcon{gem}};}%
        {r}{\node at #1 [cellNode, fill=gray!10, text=red] {\faIcon{key}};}%
        {y}{\node at #1 [cellNode, fill=gray!10, text=ncsuBanana] {\faIcon{key}};}%
        {R}{\node at #1 [cellNode, fill=gray!10, text=red] {\faIcon{door-closed}};}
        {Y}{\node at #1 [cellNode, fill=gray!10, text=ncsuBanana] {\faIcon{door-closed}};}%
        {W}{\node at #1 [cellNode, fill=gray!99] {};}%
        {.}{\node at #1 [cellNode, fill=gray!25, draw=none] {};}%
    }[\PackageError{tree}{Undefined option to tree: #2}{}]%
  }

\begin{tikzpicture}[scale=0.45]
\usetikzlibrary{calc}

\tikzstyle{cellNode}=[draw=white, inner sep=1, minimum width=12, minimum height=12, text=white, font=\footnotesize\sffamily, text centered, draw=none]
\tikzstyle{labelNode}=[fill=none, draw=none, text=gray, font=\small\sffamily]

\foreach \y [count=\n] in {
 {r, ., ., ., m, W, W, g},
 {y, ., W, W, ., W, W, .},
 {W, W, W, W, ., W, W, .},
 {., R, h, ., ., ., ., .},
 {., W, ., W, W, W, W, .},
 {., W, ., W, W, W, W, Y},
 {Y, W, ., W, W, W, W, .},
 {g, W, g, W, W, W, W, g}, 
  } {
      \foreach \x [count=\m] in \y {
       \makeCell{(\m,-\n)}{\x};
    }
}

  \pgfmathsetmacro{\z}{7} 
  \foreach \a [count=\i] in {0,...,\z} {
    \node[labelNode,anchor=center,yshift=2] at (\i,-\z-2) {\a};
  }
  \foreach \a [count=\i] in {0,...,\z} {
    \node[labelNode,anchor=east,xshift=-4] at (1,-\i) {\a};
}

\draw[->,dotted,red,thick] (7,-4) -- (4,-4);


\end{tikzpicture}
\caption{The principal moves toward the red door and asks for the red key. Tomcat figures the principal wants the gem at (7, 0) and will also need the yellow key.}
\label{fig:dkg_2}
\end{subfigure}
\begin{subfigure}[t]{0.32\textwidth}
\centering
\includestandalone[width=0.99\linewidth]{tikz/final-grid}
\caption{Tomcat fetches the red (0, 0) \emph{and} yellow (1, 0) keys to the principal, so they can unlock the red (3, 1) and yellow (6, 0) doors to reach the gem at (7, 0).}
\label{fig:dkg_3}
\end{subfigure}
\caption{Example scenario in the Doors, Keys, and Gems world, showing an agent inferring a (human) principal's mental state.}
\label{fig:doors-keys-gems}
\end{figure}
\subsection{\textinred{Doors, Keys, and Gems Environment}}
\label{sec:dkg}
Figure~\ref{fig:doors-keys-gems} illustrates a representative scenario for Instruction Inference. 
Figure~\ref{fig:dkg_1} depicts an initial grid configuration of the Doors, Keys, and Gems environment. We express coordinates as (row, column), thus the principal is at (3, 6). 
Figure~\ref{fig:dkg_2} demonstrates the observed grid configuration after the principal moves from (3, 6) to (3, 2) and says ``Can you get the red key?'' to the agent. The agent observes the principal moving leftward and stopping adjacent to the red door. Based on this movement, the instruction, and the grid configuration, the agent infers that the principal's desired gem is likely situated beyond the red door at (7, 0). Since access to the gem is further blocked by a yellow door, the agent collects both the red and yellow keys and delivers them to the principal. The final grid layout is illustrated in Figure~\ref{fig:dkg_3}, where the principal unlocks both the red and yellow doors and retrieves their desired gem at (7, 0).

\subsection{Instruction Types and Design Rationale}
\label{sec:types-rationale}
\begin{figure}[t]
    \centering \includestandalone[width=\linewidth,mode=image]{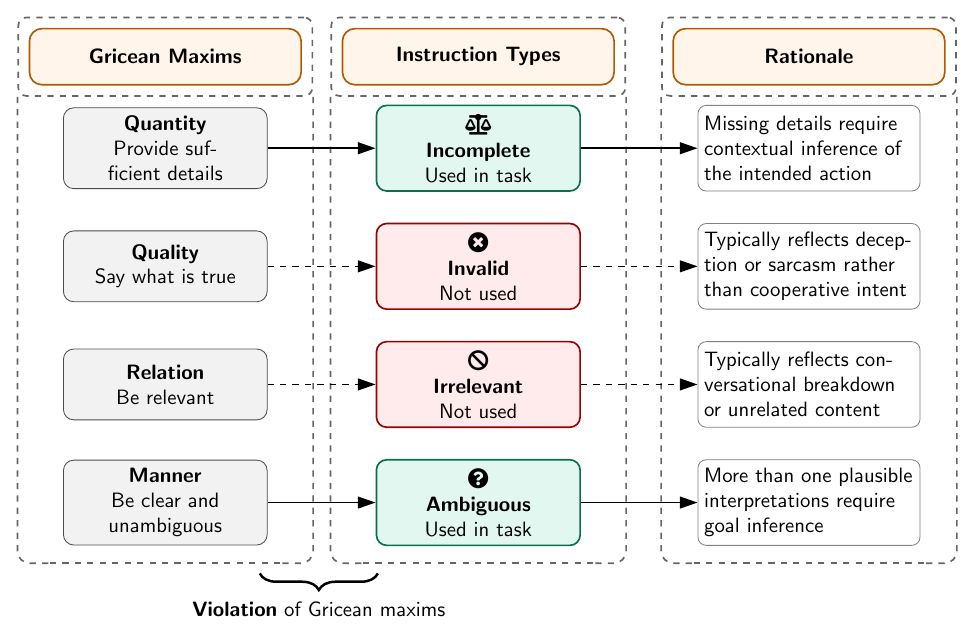}
    \caption{\textinred{Design rationale of unclear instruction types derived from violations of Gricean maxims. The Instruction Inference task categorizes \emph{incomplete} (Quantity violation) and \emph{ambiguous} (Manner violation) as unclear instructions.}}
    \label{fig:motivation-design}
\end{figure}

We ground the design of our instruction types in Grice's \emph{Cooperative Principle} \cite{Grice75}, which holds that effective communication requires participants to cooperate.
Grice defined four conversational maxims under this principle: Quantity, Quality, Relation, and Manner, summarized in Figure~\ref{fig:motivation-design}.
When a maxim is violated, listeners use pragmatic inference (termed \emph{implicature}) to infer the implied meaning from shared context \cite{Grice75}.

\citet{AAMAS-25:Gricean} classify unclear instructions as violations of the Gricean maxims.
Violations of Quantity, Quality, Relation, and Manner maxims yield incomplete, invalid, irrelevant, and ambiguous instructions, respectively. 
Incomplete instructions lack sufficient information needed to determine the intended action.
Invalid instructions contain false information about actions unachievable within the task constraints.
Irrelevant instructions have no connection to the task or to the speaker's goals.
Ambiguous instructions admit more than one plausible interpretation.
Clear instructions do not violate any maxim.

We use \emph{incomplete} and \emph{ambiguous} instructions among the four unclear categories because they are especially characteristic of cooperative collaboration, where speakers often assume shared context and therefore do not fully specify their intentions \cite{Zhi-Xuan+24:CLIPS, AAMAS-25:Gricean}.  
Conversely, invalid and irrelevant instructions are less representative of cooperative collaboration and would shift the task away from goal inference toward deception detection, sarcasm identification, or conversational breakdown.

\begin{table}[htb]
\centering
\rowcolors{2}{white}{orange!15}
\resizebox{\textwidth}{!}{
\begin{tabular}{l l l c c}
\toprule
 & \textbf{Instruction Type} & \textbf{Gricean Maxim Violation} & \textbf{Used in Setup} & \textbf{\# Instructions} \\
\midrule

\multicolumn{2}{l}{Clear} & None & Yes & 8 \\

& Incomplete & Quantity & Yes & 6 \\

\cellcolor{white}& Ambiguous & Manner & Yes & 6 \\

& Invalid & Quality & No & -- \\

\cellcolor{white}\parbox[t]{2mm}{\multirow{-4}{*}{\rotatebox[origin=c]{90}{Unclear}}} & Irrelevant & Relation & No & -- \\

\bottomrule
\end{tabular}
}
\caption{\textinred{Instruction types in relation to violations of Gricean maxims and their distribution in the Instruction Inference task}.}
\label{tab:instruction_types}
\end{table}

\textinred{Overall, our dataset contains three instruction categories: clear, incomplete, and ambiguous. 
The Instruction Inference task evaluates whether an agent can correctly interpret each category and infer the principal's intended goal. Table~\ref{tab:instruction_types} reports the distribution of these categories in our dataset.}
 
\subsection{\textinred{Instruction Inference Pipeline}}
\label{sec:pipeline}
We show our pipeline in Figure~\ref{fig:flow-tomcat}, which outlines the key components of the Instruction Inference task. In this setting, agents must interpret instructions in context, infer the principal's underlying intentions, and execute a sequence of optimal actions. The responses generated by the agents include a natural language response, an action plan, and an explanation of the reasoning behind each step. These responses collectively capture the agent's interpretation of the instruction and provide a structured basis for evaluating its ToM capabilities. By requiring agents to model latent goals and reason over \textinred{incomplete or ambiguous} instructions, the Instruction Inference task serves as a robust framework to assess goal-oriented ToM in a collaborative environment. 

\section{The Tomcat Framework}
\label{sec:Tomcat}
Tomcat is an LLM-based agent, designed to collaborate with a principal on the Instruction Inference task. Tomcat leverages the common ground \cite{Clark-15:common-ground}, including shared objectives and contextual interpretation of the environment. Further, Tomcat's decision making employs in-context learning and chain-of-thought reasoning to infer implicit instructions and execute the optimal actions. The prompt design comprises three key components:

\paragraph{\textbf{Common Ground}} It contains the background information, objectives, and rules of the collaborative game. It establishes common ground between Tomcat and the principal, including knowledge of the environment. The full prompt template is provided in \ref{sec:CGC}. 

\begin{figure}[htb]
    \begin{promptbox}[colback=red!20]{robot}{Response Generation}
   Based on your understanding of the game rules, current grid layout, human actions, instruction, and the labeled examples provided below, generate the instruction type (clear or unclear), response, and optimal actions for the following human actions and instruction, delimited by triple backticks:\\
\texttt{```}\\
\texttt{\textbf{Human Action:} The human moves to the right from  their current position at (5,0) until they reach at (5,4) and gives the instruction.\\
\textbf{Instruction:} Can you pass me the blue key?\\
\textbf{Type:} \textless Identify the instruction type and provide chain-of-thought reasoning for the type based on the current grid configuration\textgreater\\
\textbf{Response:} \textless Generate chain-of-thought reasoning for 
the response based on the current grid configuration\textgreater\\
\textbf{Actions:} \textless Provide optimal actions from the 
response based on the instruction and the current 
grid configuration\textgreater\\}
\texttt{```}    
\end{promptbox}
    \caption{Excerpt of the Response Generation component.}
    \label{fig:RG}
\end{figure}

\paragraph{\textbf{Response Generation}} This incorporates the principal's actions (movements as displayed in the observed grid in Figure~\ref{fig:dkg_2}), the instruction, and the current grid (configuration). Using this information, Tomcat is tasked with determining the type of instruction, generating the natural language response, and the sequence of actions to achieve the shared objective. Figure~\ref{fig:RG} shows an excerpt of the prompt (detailed prompt templates are available in \ref{sec:RGC}). The responses generated by Tomcat are as follows: 

\begin{description}
    
    \item [Type:] Tomcat classifies the instruction as either \emph{clear} or \emph{unclear} by analyzing the principal's actions given the grid and the instruction, as shown in Figure~\ref{fig:RG}. Additionally, Tomcat provides a rationale for the classification, facilitating the inference of the principal's intent by identifying the desired gem. 
    
    For instance, \textsl{Type: Unclear. Based on the human's movement, instruction, and the current grid configuration, I infer that the human moved to (8,4) to collect the gem at (8,9), which is located beyond the blue door and red door at (8,6) and (8,8), respectively. Since the human  asked for a blue key, it indicates they want to unlock the blue door first. However, the instruction does not mention the need for a red key to unlock the subsequent red door, making it unclear.}
 
    \item [(Natural Language) Response:] Tomcat generates a natural language response detailing the actions it executes to assist the principal, along with the anticipated next steps of the principal. We refer to Tomcat's executed actions as \emph{actions} and the complete response, which includes both Tomcat's action and the principal's subsequent steps, as a \emph{plan}. This generated plan enables us to evaluate Tomcat's ToM capabilities.
    
    For instance, \textsl{Response: The instruction asks to collect a blue key, and based on the human's movements and the grid configuration, the desired gem they want to collect is placed at (8,9). However, this gem is beyond both a blue door at (8,6) and a red door at (8,8). Therefore, I will collect one blue key and one red key located at (3,8) and (5,8), respectively. To minimize movement, I will first pick up the blue key at (3,8), followed by the red key at (5,8), as this strategy reduces the number of steps from my current position at (1,8) to the human at (8,4). Once both keys are collected, I will pass them to the human, allowing them to unlock the blue door and the red door at (8,6) and (8,8) respectively and retrieve the gem at (8,9).}  

    \item [Actions:] Tomcat extracts the sequence of optimal actions from its response, generating them in an enumerated format. This sequence includes both Tomcat's executed actions and the principal's subsequent steps. Structuring responses this way facilitates efficient annotation and evaluation. For instance,

   \begin{enumerate}[leftmargin=3em]
    \item \tikz[remember picture] \node[coordinate, yshift=1.1em] (n11) {};Collect: blue\_key at (3,8). \tikz[remember picture] \node[coordinate, yshift=1.1em] (n1) {};
    \item Collect: red\_key at (5,8). \tikz[remember picture] \node[coordinate, yshift=-0.5em, xshift=0.5em] (n2) {};
    \item Pass: blue\_key and red\_key to the human at (8,4).
    \item Unlock: human unlocks the Blue\_door at (8,6) and the Red\_door at (8,8).
    \item \tikz[remember picture] \node[coordinate, yshift=-0.5em] (n5) {};Retrieve: human retrieves gem at (8,9).
\end{enumerate}
\begin{tikzpicture}[overlay, remember picture]
    \node[coordinate, right=0em of n1] (n3){};
    \node[coordinate, right=0em of n2] (n4){};
    \draw[thick,decorate,decoration={brace,amplitude=5pt,mirror}] (n4) -- (n3) node[midway, right=5pt] {\textbf{Action}};
    \node[coordinate, left=1.5em of n11] (n1r){};
    \node[coordinate, left=1.5em of n5] (n5r){};
    \draw[thick,decorate,decoration={brace,amplitude=5pt,mirror}] (n1r) -- (n5r) node[midway, left=5pt] {\textbf{Plan}};
\end{tikzpicture}
\end{description}
\begin{figure}[htb]
  \centering
    \begin{promptbox}[colback=ncsuLightBlue!60]{robot}{Demonstration Exemplars: CP}
    Use the following problems and examples, delimited by triple quotes, to understand how to generate the appropriate type, response, and actions for each instruction and human actions.\\
\texttt{'''}\\
   ---\textbf{Problem (1)} demonstrating initial, observed, completed grids with figures and the example. \\\\
\textbf{Figure 2:} Initial Grid Configuration \\
\ldots \\
   \textbf{--Initial Grid:} This grid in Figure 2 \ldots
   \\\\
   \textbf{Figure 3:} Observed Grid Configuration \\
   \ldots\\
   \textbf{--Observed Grid:} The human moves left to \ldots
     \\\\
   \textbf{Figure 4:} Completed Grid Configuration \\
   \ldots\\
   \textbf{--Completed Grid:} You infer that the human \ldots\\\\
   \texttt{\textbf{Example} from this grid configuration.\\
   \textbf{Human Action:} The human moves left from their\ldots\\
   \textbf{Instruction:} Can you pass me the red key?\\
   \textbf{Type:} Unclear. Based on the instruction and \ldots\\
   \textbf{Response:} There are two yellow doors, one \ldots\\
   \textbf{Actions:} 1) Collect red\_key at (0,0) \\\ldots
   }
\end{promptbox}
  \caption{Excerpt of the Demonstration Exemplar component for Tomcat with CP.}
  \label{fig:dc_pap}
\end{figure}
\paragraph{\textbf{Demonstration Exemplars}} \textinred{These} provide solved problem instances to illustrate correct reasoning and execution. The problems correspond to the grid scenarios depicted in Figures~\ref{fig:dkg_1},~\ref{fig:dkg_2}, and~\ref{fig:dkg_3}, whereas the demonstrations refer to the input-output pairs used by LLMs for in-context learning \cite{Brown+20:few-shot-learners}. The input may include the grid, principal's preceding actions, and the instruction, whereas the output includes the identification of the instruction type with the principal's intent, the optimal response, and the corresponding sequence of actions. The input-output structure is defined as follows:
\begin{align}
\text{Input}  &= \text{(Optional) Grid States} + \text{Principal's Actions} + \text{Instruction}  \label{eq:input} \\ 
\text{Output} &= \text{Type} + \text{Response} + \text{Actions} \label{eq:output}
\end{align}

The Tomcat framework has \emph{two} variants: 
\begin{enumerate*}[label=(\arabic*)]
    \item the Commonsense Prompt (CP) and
    \item the Few-shot Chain-of-Thought (Fs-CoT) prompt.
\end{enumerate*}
 These variants retain the three core components described earlier, but differ in the Demonstration Exemplar component with minor modifications in the common ground component. In CP (Figure~\ref{fig:dc_pap}), the \emph{input} within the Demonstration Exemplar includes the grid states (initial, observed, completed), whereas in Fs-CoT, it does not \textinred{(Equation~\ref{eq:input})}.  

\subsection{Commonsense Prompt (CP)}
\label{sec:pap}
This variant of Tomcat aligns closely with the task presented to the participants: the demonstration exemplar consists of two problem scenarios shared with the participants. Each scenario features the initial, observed, and completed grid states, accompanied by explanations of the events that occur in those states in Figure~\ref{fig:doors-keys-gems}. Figure~\ref{fig:dc_pap} presents an excerpt from the demonstration component of CP (detailed prompt templates are available in \ref{sec:DCP}). The input-output format corresponds to Equations~\ref{eq:input} and \ref{eq:output}. The other components of this prompt, i.e., the common ground and the response generation, are the same as those provided to the participants, with only minor modifications, as described in Section~\ref{sec:exp}.

\begin{figure}[htb]
    \begin{promptbox}[colback=green!10]{robot}{Additional Rules for Fs-CoT}
   \textbf{\textit{Rules of the Game:}} \\
   \ldots \\
\ldots\\
- Always generate your response by cross-checking and verifying it against the current grid configuration to ensure accuracy. \\
- When traversing the grid, minimize and optimize your path, avoiding collisions with walls (W) or other obstacles. \\
- Infer the human's desired gem based on their movement, instruction, and grid configuration. Despite unclear instructions, generate a response that captures the human's intent and ensures successful retrieval of their desired gem. \\
- Apply theory of mind principles to infer the human's desired gem based on their actions and instructions.
 
\end{promptbox}
    \caption{Additional rules of Common Ground for Tomcat with Fs-CoT.}
    \label{fig:added_rules}
\end{figure}

\subsection{Fs-CoT Prompt}
\label{sec:Fs-CoT}
This variant employs few-shot chain-of-thought (Fs-CoT) demonstrations, which follow the input-output format of the CP exemplar shown at the bottom of Figure~\ref{fig:dc_pap}, but \emph{exclude} grid states to prevent exceeding the context window limit of GPT-4o. We include seven input-output demonstrations (\emph{k=7}) to guide Tomcat in understanding the expected response format. Through iterative testing \cite{Brown+20:few-shot-learners}, we observed that Tomcat's performance plateaued at \emph{k=7} demonstrations, with no further gains from adding more examples. The remaining components follow those in the CP, with targeted enhancements to the common ground component, particularly in the \emph{Rules} portion within the component, as shown in Figure~\ref{fig:added_rules}. Through prompt engineering, we found that appending relevant rules helps establish a robust common ground between Tomcat and the principal. The detailed prompt templates are available in \ref{sec:DFs}.

\section{Experiments}
\label{sec:exp}
The Instruction Inference task allows us to evaluate the ToM capabilities of an agent by assessing how well it interprets instructions, infers the principal's intent to retrieve a gem, and generates responses aligned with that intent. 
To this end, we conducted experiments using the Tomcat variants and an empirical study involving 52 human participants who acted as agents collaborating with the principal. 
The participants were undergraduate and graduate students with STEM backgrounds. This study was approved by our Institutional Review Board (IRB) with an \emph{exempt} status.

Participants received a detailed problem description. This description includes an overview of the Doors, Keys, and Gems environment, the objectives, rules, two problem scenarios, and a task description. Participants were tasked with inferring the principal's intentions, generating an optimal response, and outlining the sequence of optimal actions. The problem description provided to participants was identical to Tomcat's CP framework, except for two key differences. First, the problem scenarios for the participants featured animated GIF representations of the grid states, whereas CP used static matrix representations accompanied by explanatory text. Second, whereas Tomcat was explicitly tasked with identifying the instruction type as part of its response, participants were not required to perform this step.

\subsection{Dataset and LLM}
\label{sec:dataset}
We used 16 grid configurations, adapted from \citet{Zhi-Xuan+24:CLIPS}, as the baseline for evaluating ToM capabilities on Instruction Inference. 
\textinred{We created 20 instructions (eight clear or direct, six incomplete, and six ambiguous),  and constructed 20 grid-instruction problem scenarios.}
\textinred{These scenarios were designed from the 16 grid configurations and 20 instructions to represent specific ToM tasks on instruction inference, rather than pairing every grid with every instruction.} 
We used three LLMs---GPT-4o, DeepSeek-R1, and Gemma-3-27B---for Tomcat's reasoning and response generation. We selected \textinred{these LLMs} because of their large context window. To optimize performance, we capped the maximum token count to 512, striking a balance between comprehensive outputs and computational efficiency. The temperature parameter was set to 0.2 to reduce variability in responses while allowing some nondeterminism.

\textinred{Each scenario was evaluated using three LLMs, two Tomcat variants, and 52 human participants (each completing 10 scenarios). 
Each evaluation produced three structured outputs (intent inference, natural-language response, and action-plan formulation), enabling a fine-grained analysis of ToM reasoning.}
\textinred{In total, our design yields:
$(3 \text{ LLMs} \times 2 \text{ variants} \times 20 \text{ scenarios}) + (52 \text{ humans} \times 10 \text{ scenarios each})$ $= 640$ scenario-level responses, corresponding to 1,920 evaluated reasoning components. 
This scale supports systematic model-model and model-human comparisons as outlined in Section~\ref{sec:stat-test}.
It also aligns with established ToM research practices for LLMs that emphasize carefully designed reasoning scenarios over exhaustive task enumeration. 
For example, \citet{Kosinski-24:evaluating-ToM-LLMs} uses 20 scenarios, and \citet{Strachan+24:testing-ToM-LLM} employ an average of 12 scenarios per ToM task (e.g., false-belief, faux pas, and hinting).}

\textinred{Participants reported approximately 10--15 minutes per scenario, reflecting the multistep reasoning involved in interpreting indirect instructions, inferring latent goals, and formulating action plans. This indicates that each scenario constitutes a concrete Instruction Inference task rather than a simple prompt-response item.}


\subsection{Evaluation Metrics}
\label{sec:metrics}
We employed five metrics to evaluate the ToM capabilities of Tomcat and the participants. We define an additional metric, \emph{Instruction Accuracy}, just for the Tomcat variants and not for the participants. These metrics assess the ability to recognize the principal's intentions and evaluate the quality of actions and the plans generated by both groups. Whereas actions focus exclusively on execution, plans extend to include the interpretation of the principal's next steps, reflecting the principal's mental model. This distinction is crucial for assessing ToM capabilities, as the plans capture deeper cognitive understanding. Accordingly, the metrics are categorized into \emph{actions} and \emph{plans}, as detailed below:

\begin{description}
\item[Intent Accuracy] assesses how well the generated response aligns with the principal's goal. It is scored as 1 if the goal matches and 0 otherwise.

\item[Action Feasibility] determines whether the suggested actions can be executed in the grid, regardless of efficiency. It is calculated as:
\begin{align}
&\text{Action Feasibility} = \frac{|A_{\text{feasible}}|}{|A_{\text{total}}|}, \text{ where } 
\end{align}
 $A_{\text{feasible}}$ is the set of feasible actions, and $A_{\text{total}} = A_{\text{ground-truth}} \cup A_{\text{incorrect}}$ is the union of ground-truth and incorrect or unnecessary actions.

\item[Action Optimality] measures the ratio of optimal actions correctly performed to total actions.

\item[Plan Feasibility] measures the ratio of actions within a plan that are executable in the environment, regardless of efficiency. That is:
\begin{align}
&\text{Plan Feasibility} = \frac{|A_{\text{feasible\_plan}}|}{|A_{\text{total\_plan}}|}, \text{ where }
\end{align}
$A_{\text{feasible\_plan}}$ is the set of feasible actions in the plan, and $A_{\text{total\_plan}}$ $=$\newline$A_{\text{ground-truth\_plan}}$ $\cup$ $A_{\text{incorrect\_plan}}$ is the union of ground-truth and incorrect or unnecessary actions in the plan.

\item[Plan Optimality] assesses whether the actions in the plan are in the correct sequence. If correct, 1, else 0.  

\item[Instruction Accuracy] evaluates whether Tomcat correctly identifies the instruction type (clear or unclear). If correct, 1, else 0. 
\end{description}



\subsection{Task Structure and Difficulty}
\label{sec:task-difficulty}

We characterize the 20 problem scenarios in Table~\ref{tab:complexity}. 
The first column represents the scenario number and the \fqt{Grid} column shows the grid identifier derived from \citepossessive{Zhi-Xuan+24:CLIPS} dataset. 
Grids \fqt{3}, \fqt{6}, \fqt{12}, and \fqt{13} appear twice, each paired with a different instruction, producing distinct grid-instruction scenarios. 

\paragraph{Grid Features}
We report basic structural properties of each grid layout, including the grid dimensions and the number of keys and doors present in the environment. 
These quantities determine the size of the navigation space and the number of objects present in the collaborative task.

\paragraph{Spatial Obstruction}
We quantify the navigation constraints of the grid using two heuristics: \emph{detour ratio} and \emph{corridor tightness}.

For a target key location $k$ and agent start position $m$, the detour ratio is defined as:
\begin{equation}
\rho(m,k) = 
\frac{d_{\mathrm{sp}}(m,k)}
{\max(1, d_{\mathrm{man}}(m,k))}, \text{ where }
\end{equation}
$d_{\mathrm{sp}}$ denotes the shortest-path distance on the grid (computed via BFS) and $d_{\mathrm{man}}$ denotes the Manhattan distance.
For instance, in Scenario~1 (Figure~\ref{fig:obs-moderate}, Grid~3), two red keys serve as candidate targets: $k_1$ at (1, 3) and $k_2$ at (1, 7), yielding $\rho(m,k_1) = 6/4 = 1.50$ and $\rho(m,k_2) = 2/2 = 1.00$ respectively.
The aggregate detour ratio for the scenario, $\rho_{\text{agg}}$, is computed by averaging $\rho(m,k)$ over the number of keys \emph{fetched by the agent}.
\begin{equation}
\rho_{\text{agg}} = 
\frac{1}{|T|}
\sum_{k \in T} \rho(m,k), \text{ where }
\end{equation}
$T$ is the set of target keys fetched by the agent in a scenario.
Thus, $\rho_{\text{agg}} = (1.50 + 1.00)/2 = 1.25$ for Scenario~1.

\begin{table}[t]
\centering
\rowcolors{2}{mint!10}{white}
\resizebox{\textwidth}{!}{
\begin{tabular}{lccccccccr}
\toprule
& & \multicolumn{3}{c}{\textbf{Grid Features}} & & \multicolumn{2}{c}{\textbf{Ambiguity}} & & \\
\cmidrule(lr){3-5}\cmidrule(lr){7-8}
\textbf{No.} & \textbf{Grid} & \textbf{Dims} & \textbf{Keys} & \textbf{Doors} & \textbf{Obstr.} & \textbf{Deict.} & \textbf{Impl.} & \textbf{Sub.} & \textbf{{\small\color{pumpkin}\faUsers}\,Acc.} \\
\midrule
1 & 3 & 10$\times$12 & 5 & 5 & Moderate & \ssig & \ssig & 7 & 0.84 \\
2 & 5 & 9$\times$10 & 3 & 5 & High & \nosig & \ssig & 7 & 0.92 \\
3 & 6 & 11$\times$11 & 4 & 7 & High & \nosig & \ssig & 5 & 0.68 \\
4 & 7 & 11$\times$11 & 5 & 5 & High & \nosig & \nosig & 6 & 0.92 \\
5 & 10 & 10$\times$11 & 4 & 7 & High & \nosig & \ssig & 12 & 0.88 \\
6 & 12 & 7$\times$9 & 2 & 3 & Low & \ssig & \ssig & 4 & 0.76 \\
7 & 13 & 7$\times$11 & 3 & 5 & Moderate & \nosig & \ssig & 4 & 0.84 \\
8 & 15 & 9$\times$8 & 2 & 3 & Moderate & \ssig & \ssig & 5 & 0.88 \\
9 & 18 & 8$\times$8 & 2 & 4 & Moderate & \ssig & \ssig & 5 & 0.84 \\
10 & 20 & 8$\times$8 & 2 & 4 & Moderate & \ssig & \ssig & 5 & 0.80 \\
11 & 3 & 10$\times$12 & 5 & 5 & Moderate & \ssig & \ssig & 7 & 0.74 \\
12 & 6 & 11$\times$11 & 4 & 7 & High & \nosig & \ssig & 5 & 0.70 \\
13 & 8 & 12$\times$13 & 5 & 4 & High & \nosig & \nosig & 7 & 0.81 \\
14 & 9 & 9$\times$11 & 3 & 5 & High & \nosig & \ssig & 7 & 0.56 \\
15 & 10b & 10$\times$11 & 4 & 7 & High & \nosig & \ssig & 12 & 0.93 \\
16 & 11 & 8$\times$12 & 3 & 5 & Moderate & \ssig & \ssig & 7 & 0.78 \\
17 & 12 & 7$\times$9 & 2 & 3 & Low & \ssig & \nosig & 3 & 0.81 \\
18 & 13 & 7$\times$11 & 3 & 5 & Moderate & \ssig & \ssig & 4 & 0.78 \\
19 & 17 & 8$\times$8 & 2 & 3 & Moderate & \ssig & \ssig & 5 & 0.93 \\
20 & 19 & 7$\times$10 & 3 & 5 & Moderate & \nosig & \ssig & 8 & 0.70 \\
\bottomrule
\end{tabular}
}
\caption{\textinred{Scenario Characterization: 20 unique grid-instruction problem scenarios from the dataset, showing systematic variation across grid metrics (dimensions, keys, doors), obstruction level, ambiguity types (deictic references, implicit goal inference), plan depth (subgoals), and empirical difficulty measured by human participants' accuracy in recognizing the principal's intent.}}
\label{tab:complexity}
\end{table}

To capture narrow navigation corridors induced by walls, we measure corridor tightness for each cell, $c$, along the agent's shortest path, $P(m,k)$, from its start position $m$ to a target key $k$. The corridor tightness is defined as:
\begin{equation}
\beta(m,k) =
\frac{|\{c \in P(m,k):\deg(c)\leq 2 \}| }
{|P(m,k)|},\text{ where }
\end{equation}
$|P(m,k)|$ denotes the number of cells along the agent's shortest path to the required keys, and $\deg(c)$ denotes the number of traversable neighboring cells of $c$ in the grid's 4-neighborhood. 
Cells with $\deg(c) \leq 2$ correspond to narrow passages where movement is highly constrained. 
Thus, this heuristic measures the fraction of path cells lying in corridor-like regions of the grid.
Continuing with Scenario~1, the path to $k_1$ spans 7 cells, of which 4 have $\deg(c) \leq 2$, yielding $\beta(m,k_1) = 4/7 \approx 0.57$; the path to $k_2$ spans 3 cells, of which 2 are narrow, yielding $\beta(m,k_2) = 2/3 \approx 0.67$.
The aggregate corridor tightness, $\beta_{\text{agg}}$, is computed by averaging $\beta(m,k)$ over the number of target keys.
\begin{equation}
\beta_{\text{agg}} = 
\frac{1}{|T|}
\sum_{k \in T} \beta(m,k), \text{ where }
\end{equation}
$T$ is the set of target keys fetched by the agent in a scenario.
Thus, $\beta_{\text{agg}} = (0.57 + 0.67)/2 \approx 0.62$ for Scenario~1.

Based on $\rho_{\text{agg}}$ and $\beta_{\text{agg}}$, scenarios are categorized into \emph{Low}, \emph{Moderate}, and \emph{High} obstruction levels.

\begin{equation}
O =
\begin{cases}
\text{Low}, & \rho_{\text{agg}} \leq 1.15 \ \land \ \beta_{\text{agg}} < 0.5 \\
\text{Moderate}, & \rho_{\text{agg}} \leq 1.5 \ \land \ 0.5 \leq \beta_{\text{agg}} < 0.8 \\
\text{High}, & \text{otherwise}.
\end{cases}
\end{equation}

The thresholds reflect interpretable levels of spatial constraint in the grid environment. 
Values of $\rho_{\text{agg}} \leq 1.15$ indicate near-Manhattan navigation with minimal detours, $1.15 < \rho_{\text{agg}} \leq 1.5$ indicates moderate detours around walls, and $\rho_{\text{agg}} > 1.5$ indicates substantial path detours caused by obstacles. 
Further, $\beta_{\text{agg}}$ distinguishes open paths ($\beta_{\text{agg}} < 0.5$), corridor-dominated navigation ($0.5 \leq \beta_{\text{agg}} < 0.8$), and strongly constrained paths ($\beta_{\text{agg}} \geq 0.8$), yielding the Low, Moderate, and High obstruction categories.
Scenario~1 thus falls in the Moderate category ($\rho_{\text{agg}} = 1.25$, $\beta_{\text{agg}} = 0.62$).
Figure~\ref{fig:obstruction-levels} shows a representative grid from each obstruction level.

\begin{figure}[t]
\subcaptionsetup{width=0.9\linewidth,justification=raggedright, singlelinecheck=false}
\begin{subfigure}[t]{0.32\textwidth}
\centering
\newcommand*{\makeCell}[2]{%
    \IfEqCase{#2}{%
        {m}{\node at #1 [cellNode,fill=gray!10, text=purple] {\faRobot};}
        {h}{\node at #1 [cellNode, fill=gray!10, text=orange] {\faUser};}%
        {g}{\node at #1 [cellNode, fill=gray!10, text=ncsuTeal] {\faIcon{gem}};}%
        {r}{\node at #1 [cellNode, fill=gray!10, text=red] {\faIcon{key}};}%
        {y}{\node at #1 [cellNode, fill=gray!10, text=ncsuBanana] {\faIcon{key}};}%
        {b}{\node at #1 [cellNode, fill=gray!10, text=blue] {\faIcon{key}};}%
        {B}{\node at #1 [cellNode, fill=gray!10, text=blue] {\faIcon{door-closed}};}%
        {R}{\node at #1 [cellNode, fill=gray!10, text=red] {\faIcon{door-closed}};}
        {Y}{\node at #1 [cellNode, fill=gray!10, text=ncsuBanana] {\faIcon{door-closed}};}%
        {W}{\node at #1 [cellNode, fill=gray!99] {};}%
        {.}{\node at #1 [cellNode, fill=gray!25, draw=none] {};}%
    }[\PackageError{tree}{Undefined option to tree: #2}{}]%
  }

\begin{tikzpicture}[scale=0.45]
\usetikzlibrary{calc}

\tikzstyle{cellNode}=[draw=white, inner sep=1, minimum width=12, minimum height=12, text=white, font=\footnotesize\sffamily, text centered, draw=none]
\tikzstyle{labelNode}=[fill=none, draw=none, text=gray, font=\small\sffamily]

\foreach \y [count=\n] in {
{g, ., W, W, W, W, W, W, W},
{r, b, W, W, W, W, W, W, W},
{., ., m, ., ., ., ., ., g},
{R, W, W, W, W, B, W, W, .},
{., ., h, ., ., ., ., ., R},
{W, ., W, W, ., W, W, W, .},
{W, g, W, W, ., W, W, W, g},
, 
  } {
      \foreach \x [count=\m] in \y {
       \makeCell{(\m,-\n)}{\x};
    }
}

  \pgfmathsetmacro{\z}{8} 
  \pgfmathsetmacro{\x}{6} 
  
  \foreach \a [count=\i] in {0,...,\z} {
    \node[labelNode,anchor=center] at (\i,-\x - 1.9) {\a};
  }
  \foreach \a [count=\i] in {0,...,\x} {
    \node[labelNode,anchor=east] at (0.6,-\i) {\a};
}

\draw[->,dotted, red, very thick, rounded corners = 2pt] (5,-7) |- (4,-5);

\node[anchor=north, font=\tiny\bfseries\sffamily] at (5, -\x - 2.2) {``Can you go and unlock that door?''};

\end{tikzpicture}
\caption{Scenario 17, Grid 12. Low obstruction ($\rho_{\text{agg}}$\,=\,1.00, $\beta_{\text{agg}}$\,=\,0.25) with deictic ambiguity, but no implicit goal inference.}
\label{fig:obs-low}
\end{subfigure}
\begin{subfigure}[t]{0.32\textwidth}
\centering
\newcommand*{\makeCell}[2]{%
    \IfEqCase{#2}{%
        {m}{\node at #1 [cellNode,fill=gray!10, text=purple] {\faRobot};}
        {h}{\node at #1 [cellNode, fill=gray!10, text=orange] {\faUser};}%
        {g}{\node at #1 [cellNode, fill=gray!10, text=ncsuTeal] {\faIcon{gem}};}%
        {r}{\node at #1 [cellNode, fill=gray!10, text=red] {\faIcon{key}};}%
        {y}{\node at #1 [cellNode, fill=gray!10, text=ncsuBanana] {\faIcon{key}};}%
        {b}{\node at #1 [cellNode, fill=gray!10, text=blue] {\faIcon{key}};}%
        {B}{\node at #1 [cellNode, fill=gray!10, text=blue] {\faIcon{door-closed}};}%
        {R}{\node at #1 [cellNode, fill=gray!10, text=red] {\faIcon{door-closed}};}
        {Y}{\node at #1 [cellNode, fill=gray!10, text=ncsuBanana] {\faIcon{door-closed}};}%
        {W}{\node at #1 [cellNode, fill=gray!99] {};}%
        {.}{\node at #1 [cellNode, fill=gray!25, draw=none] {};}%
    }[\PackageError{tree}{Undefined option to tree: #2}{}]%
  }

\begin{tikzpicture}[scale=0.45]
\usetikzlibrary{calc}

\tikzstyle{cellNode}=[draw=white, inner sep=1, minimum width=12, minimum height=12, text=white, font=\footnotesize\sffamily, text centered, draw=none]
\tikzstyle{labelNode}=[fill=none, draw=none, text=gray, font=\small\sffamily]

\foreach \y [count=\n] in {
{., ., ., y, ., ., ., b, W, W, W, W},
{r, W, W, r, W, W, ., r, W, W, W, g},
{W, W, W, W, W, W, m, W, W, W, W, R},
{W, W, W, W, W, W, ., W, W, W, W, .},
{g, ., ., ., B, ., h, R, ., ., ., .},
{W, W, W, W, W, W, ., W, W, W, W, W},
{W, W, W, W, W, W, ., W, W, W, W, W},
{., ., ., Y, ., ., ., W, W, W, W, W},
{B, W, W, W, W, W, ., W, W, W, W, W},
{g, W, W, W, W, W, ., ., ., ., ., g}, 
} {
      \foreach \x [count=\m] in \y {
       \makeCell{(\m,-\n)}{\x};
    }
}
\pgfmathsetmacro{\z}{11} 
\pgfmathsetmacro{\x}{9}  

  \foreach \a [count=\i] in {0,...,\z} {
    \node[labelNode,anchor=center] at (\i,-\x - 1.9) {\a};
  }
  
  \foreach \a [count=\i] in {0,...,\x} {
    \node[labelNode,anchor=east] at (0.6,-\i) {\a};
}

\draw[->,dotted, red, very thick, rounded corners = 2pt] (7,-10) -- (7,-6);

\node[anchor=north, font=\scriptsize\bfseries\sffamily] at (6.7, -\x-2.2) {``Give me a {\color{red}red key} for this door?''};

\end{tikzpicture}
\caption{Scenario 1, Grid 3. Moderate obstruction ($\rho_{\text{agg}}$\,=\,1.25, $\beta_{\text{agg}}$\,=\,0.62) with deictic ambiguity and implicit goal inference.}
\label{fig:obs-moderate}
\end{subfigure}
\begin{subfigure}[t]{0.32\textwidth}
\centering
\newcommand*{\makeCell}[2]{%
    \IfEqCase{#2}{%
        {m}{\node at #1 [cellNode,fill=gray!10, text=purple] {\faRobot};}
        {h}{\node at #1 [cellNode, fill=gray!10, text=orange] {\faUser};}%
        {g}{\node at #1 [cellNode, fill=gray!10, text=ncsuTeal] {\faIcon{gem}};}%
        {r}{\node at #1 [cellNode, fill=gray!10, text=red] {\faIcon{key}};}%
        {y}{\node at #1 [cellNode, fill=gray!10, text=ncsuBanana] {\faIcon{key}};}%
        {b}{\node at #1 [cellNode, fill=gray!10, text=blue] {\faIcon{key}};}%
        {B}{\node at #1 [cellNode, fill=gray!10, text=blue] {\faIcon{door-closed}};}%
        {R}{\node at #1 [cellNode, fill=gray!10, text=red] {\faIcon{door-closed}};}
        {Y}{\node at #1 [cellNode, fill=gray!10, text=ncsuBanana] {\faIcon{door-closed}};}%
        {W}{\node at #1 [cellNode, fill=gray!99] {};}%
        {.}{\node at #1 [cellNode, fill=gray!25, draw=none] {};}%
    }[\PackageError{tree}{Undefined option to tree: #2}{}]%
  }

\begin{tikzpicture}[scale=0.45]
\usetikzlibrary{calc}

\tikzstyle{cellNode}=[draw=white, inner sep=1, minimum width=12, minimum height=12, text=white, font=\footnotesize\sffamily, text centered, draw=none]
\tikzstyle{labelNode}=[fill=none, draw=none, text=gray, font=\small\sffamily]

\foreach \y [count=\n] in {
{g, ., R, ., ., ., ., r, ., ., .},
{., W, W, ., W, W, W, W, W, W, .},
{., W, W, ., W, W, W, W, W, W, .},
{., W, W, ., W, W, W, r, W, W, .},
{., W, W, ., W, W, W, B, W, W, .},
{., W, W, ., ., ., ., m, ., ., .},
{., W, W, R, W, W, ., W, W, W, R},
{., W, W, B, W, W, h, W, W, W, .},
{g, ., ., ., g, W, b, W, W, W, g}, 
  } {
      \foreach \x [count=\m] in \y {
       \makeCell{(\m,-\n)}{\x};
    }
}

  \pgfmathsetmacro{\z}{10} 
  \pgfmathsetmacro{\x}{8} 
  
  \foreach \a [count=\i] in {0,...,\z} {
    \node[labelNode,anchor=center] at (\i,-\x - 1.9) {\a};
  }
  \foreach \a [count=\i] in {0,...,\x} {
    \node[labelNode,anchor=east] at (0.6,-\i) {\a};
}

\draw[->,dotted, red, very thick, rounded corners = 2pt] (4,-6) -| (7,-7);

\node[anchor=north, font=\scriptsize\bfseries\sffamily] at (6, -\x-2.2) {``Can you go get the {\color{red}red key}?''};

\end{tikzpicture}
\caption{Scenario 14, Grid 9. High obstruction ($\rho_{\text{agg}}$\,=\,2.20, $\beta_{\text{agg}}$\,=\,1.00) with implicit goal inference, but no deictic ambiguity.}
\label{fig:obs-high}
\end{subfigure}
\caption{Example scenarios in the Doors, Keys, and Gems world, showing spatial obstruction levels (low, moderate, high) and ambiguity types (deictic, implicit goal inference).}
\label{fig:obstruction-levels}
\end{figure}

\paragraph{Ambiguity and Planning Depth}
We annotate each scenario for two forms of ambiguity: \emph{deictic ambiguity}, where the instruction contains spatial references such as, \fsl{this}, \fsl{that}, or \fsl{there}, and \emph{implicit goal inference}, where the agent must infer the principal's intended objective from contextual cues such as the principal's movement, the available resources (e.g., keys or doors), and the division of labor in the collaborative task.
For instance, Scenario~17 (Figure~\ref{fig:obs-low}) contains deictic ambiguity but not implicit goal inference: the instruction \fqt{Can you go and unlock \fsl{that} door?} uses the spatial reference \fsl{that}, yet the target door can be identified from the principal's movement toward the red door.
Scenario~1 (Figure~\ref{fig:obs-moderate}) contains both forms: the instruction \fqt{Give me a red key for \fsl{this} door?} uses \fsl{this} to indicate the red door near which the principal is standing, and it implies collecting a single red key; however, two red doors block the path to the target gem, so the agent must infer that two red keys in the grid are required to fulfill the principal's goal.
Scenario~14 (Figure~\ref{fig:obs-high}) contains implicit goal inference without deictic ambiguity: the instruction \fqt{Can you go get the red key?} is spatially unambiguous, but the grid contains two red keys---one accessible and one locked behind a blue door.
Since the principal is observed moving toward the blue key, the agent infers that while the principal collects the blue key, it should fetch and pass the accessible red key to the principal who then unlocks both doors to reach the gem.
We also report the number of subgoals required in the corresponding plan, which reflects the depth of the cooperative task.

\paragraph{Difficulty}
We report human participants' mean accuracy in recognizing the principal's intent for each scenario as an indicator of task difficulty.

\begin{table}[t]
\sisetup{
  text-family-to-math = true,
  text-series-to-math = true,
  round-precision=2,
  round-mode=places
}
\rowcolors{3}{white}{ncsuLightTeal!50}
\centering
\resizebox{\textwidth}{!}{
\begin{tabular}{ll*5{S[table-format = 2.2, input-symbols={()\%}]}}
\toprule
 \multirow{2}{*}{\textbf{Agents}}& \multirow{2}{*}{\textbf{Method}} & \textbf{Intent Acc.} & \textbf{Act. Feas.} & \textbf{Act. Opt.} & \textbf{Plan Feas.} & \textbf{Plan Opt.} \\
 &  & {(\qtyP{})} & {(\qtyP{})} & {(\qtyP{})} & {(\qtyP{})} & {(\qtyP{})} \\
\midrule
\whitecell{} & CP   & 50.00 & 78.75 & 55.00 & 60.92 & 25.00 \\
\cellcolor{white}\textbf{GPT-4o}  & Fs-CoT & \bfseries 85.00 & \bfseries 97.50 & \bfseries 92.50 & \bfseries 91.78 & \bfseries 75.00 \\
\midrule
  & CP  & 60.00 & 86.00 & 73.50 & 69.90 & 55.00 \\
\cellcolor{white}\textbf{DeepSeek-R1} & Fs-CoT & 80.00 & 88.50 & 73.50 & 80.45 & 70.00  \\
\midrule
 & CP   & 5.00 & 60.42 & 30.42 & 32.64 & 5.00  \\
\cellcolor{white}\textbf{Gemma-3-27B}   & Fs-CoT & 30.00 & 75.00 & 51.25 & 49.78 & 30.00 \\
\midrule
\textbf{Human} & Participants & 80.38 & 89.26 & 86.22 & 79.02 & 71.15  \\
\bottomrule
\end{tabular}
}
\caption{Performance comparison between LLMs and participants across metrics.}
\label{tab:comparison_metrics}
\end{table}

\section{Results and Discussion}
\label{sec:result}
Table~\ref{tab:comparison_metrics} reveals that Tomcat with Fs-CoT consistently outperformed its CP counterpart in all metrics, regardless of the underlying LLM. 
Among the Fs-CoT variants, GPT-4o outperformed, DeepSeek-R1 performed comparably \textinred{to}, and Gemma-3-27B \textinred{performed worse} than human participants. In contrast, Tomcat with CP consistently underperformed relative to participants in all metrics, although DeepSeek-R1 exhibited performance that was moderately close to, but still below, that of participants and its Fs-CoT variant. In general, DeepSeek-R1 was the strongest among the CP variants, whereas the GPT-4o Fs-CoT variant achieved the best overall performance. We substantiate these findings with further analyses \textinred{below}. 

\begin{table}[htb]
\centering
\sisetup{round-mode = places}
\sisetup{round-precision = 3}
\rowcolors{2}{ncsuBlue!15}{white}
\resizebox{\textwidth}{!}{
\begin{tabular}{
    l
    l 
    l 
    S[table-format=<2.2,round-minimum = 0.001]
    S[table-format=+1.3]
}
\toprule
LLMs & \textbf{Metric} & \textbf{Comparison} & \bm{$p$} & \textbf{Cohen's }\bm{$g$} \\
\midrule
 & Intent Accuracy & Fs-CoT vs CP & 0.039062 & 0.777778 \\
\cellcolor{white}\textbf{GPT-4o} & Plan Optimality      & Fs-CoT vs CP & 0.001953 & 1.000000 \\
 & Instruction Accuracy & Fs-CoT vs CP & 0.250000   & 1.000000 \\
\midrule
\whitecell{} & Intent Accuracy & Fs-CoT vs CP & 0.125 & 1.000000 \\
\textbf{DeepSeek-R1}& Plan Optimality      & Fs-CoT vs CP & 0.375 & 0.6 \\
\whitecell{} & Instruction Accuracy & Fs-CoT vs CP & 0.125   & 1.000000 \\
\midrule
 & Intent Accuracy & Fs-CoT vs CP & 0.0625 & 1.000000 \\
\cellcolor{white}\textbf{Gemma-3-27B}& Plan Optimality      & Fs-CoT vs CP & 0.0625 & 1.000000 \\
 & Instruction Accuracy & Fs-CoT vs CP & 0.375 & 0.600000 \\
\bottomrule
\end{tabular}
}
\caption{Results of McNemar's exact test with Cohen's $g$ for binary metrics.}
\label{tab:mcnemar}
\end{table}

\subsection{Statistical Tests}
\label{sec:stat-test}
To substantiate these findings, we conducted statistical significance tests for each metric between the three groups---two Tomcat variants and participants---separately for each LLM. We conducted a Kruskal-Wallis test followed by post hoc Dunn tests with Holm-Bonferroni correction for the continuous metrics: action feasibility, action optimality, and plan feasibility. In addition, we calculated the Rank-biserial correlation ($r$) to quantify the effect sizes for pairwise comparisons, and the results are summarized in Table~\ref{tab:kruskal_dunn}. We performed McNemar's exact test to compare the performance of the two Tomcat variants on the binary metrics---intent accuracy, plan optimality, and instruction accuracy---leveraging their paired data structure. To quantify effect size, we report Cohen's $g$. We conducted Chi-square tests and calculated Cram{\'e}r's $V$ as an effect size to compare the performance of the participants and the two Tomcat variants across the binary metrics, given their independent data. The results of these tests for the binary metrics are shown in Tables~\ref{tab:mcnemar} and~\ref{tab:chi}. The following sections present a comprehensive analysis of performance in all metrics, statistical significance results, qualitative analysis, and responses to research questions.

\begin{table}[htb]
\centering
\sisetup{round-mode = places}
\sisetup{round-precision = 3}
\resizebox{\textwidth}{!}{
\begin{tabular}{
    l
    l 
    l 
    S[table-format=<2.2,round-minimum = 0.001]
    S[table-format=+1.3]
}
\toprule
LLMs & \textbf{Metric} & \textbf{Comparison} & \bm{$p$} & \textbf{Cram{\'e}r's }\bm{$V$} \\
\midrule
 &  & Participants vs CP & 0.002632 &  0.129435 \\
 \rowcolor{ncsuPurple!25}
 \whitecell{} & Intent Accuracy  & Participants vs Fs-CoT   & 0.822833  &   0.009635 \\
 \cmidrule{2-5}
\whitecell{}\textbf{GPT-4o} &  & Participants vs CP  & 0.000033 &  0.178576 \\
\rowcolor{ncsuPurple!25}
\whitecell{} & Plan Optimality   & Participants vs Fs-CoT   & 0.903483  &  0.005218 \\
\midrule
 &  & Participants vs CP & 0.052586 &  0.083411 \\
 \rowcolor{ncsuPurple!25}
 \whitecell{} &  Intent Accuracy & Participants vs Fs-CoT   & 1.000000  &   0.00000 \\
 \cmidrule{2-5}
\textbf{DeepSeek-R1} &  & Participants vs CP      & 0.191762 &  0.056175\\
\rowcolor{ncsuPurple!25}
 \whitecell{} & Plan Optimality & Participants vs Fs-CoT   & 1.000000  &   0.00000 \\
\midrule
 &  & Participants vs CP & 0.000000000000 &  0.32967 \\
 \rowcolor{ncsuPurple!25}
 \whitecell{} &  Intent Accuracy   & Participants vs Fs-CoT   & 0.000000328153  &   0.219749 \\
 \cmidrule{2-5}
\textbf{Gemma-3-27B} &  & Participants vs CP      & 0.000000001796 &  0.258854\\
\rowcolor{ncsuPurple!25}
 \whitecell{} & Plan Optimality  & Participants vs Fs-CoT   & 0.000233537000  &   0.158348 \\
\bottomrule
\end{tabular}
}
\caption{Results of Chi-square test with Cram{\'e}r's $V$ for binary metrics.}
\label{tab:chi}
\end{table}

\begin{table}[htb]
\centering
\sisetup{round-mode = places}
\sisetup{round-precision = 3}
\resizebox{\textwidth}{!}{
\begin{tabular}{
l
l
S[table-format=<2.2,round-minimum = 0.001]
c
l
S[table-format=<2.2,round-minimum = 0.001]
S[table-format=+1.3]
}
\toprule
\textbf{LLMs} & \textbf{Metric} & $\bm{p}$ & \textbf{Dunn-Test} & \textbf{Comparison} & \textbf{Adjusted } $\bm{p}$ & $\bm{r}$ \\
\midrule
 &  &  &  & Fs-CoT vs CP  & 0.005  & 0.3525     \\
            &   &   &   & Participants vs CP & 0.003 & 0.245096 \\
\rowcolor{orange!15}            
 \whitecell{}    & Action Feasibility  & 0.002707  & Yes  & Participants vs Fs-CoT & 0.332 & -0.072115 \\
\cmidrule{2-7}
 &  &  &  & Fs-CoT vs CP  & 0.0001342224  & 0.545     \\
 \textbf{GPT-4o}    &  &  &   & Participants vs CP & 0.0000005463354 & 0.451538 \\
            & Action Optimality  & 0.000001  & Yes  & Participants vs Fs-CoT & 0.7524798 & -0.025 \\
\cmidrule{2-7}
 & & &   & Fs-CoT vs CP  & 0.003674832  &  0.55    \\
 &   &   &   & Participants vs CP & 0.003624437 & 0.335673 \\
\rowcolor{orange!15}            
 \whitecell{}   &  Plan Feasibility & 0.002561  &  Yes  & Participants vs Fs-CoT & 0.2772892 & -0.110192 \\
\midrule
 & Action Feasibility & 0.254209
 & No  & {--} & {--}  & {--}     \\ 
\cmidrule{2-7}
 &  &  &  & Fs-CoT vs CP  & 1.000000 & 0.000000     \\
 \textbf{DeepSeek-R1}  &   &   &   & Participants vs CP & 0.121575 & 0.176731 \\
 \rowcolor{orange!15}
  \whitecell{} & Action Optimality  & 0.017494  &  Yes & Participants vs Fs-CoT & 0.121575 & 0.176731 \\
\cmidrule{2-7}
 & Plan Feasibility & 0.44473 & No  & {--} & {--}  & {--}  \\
 \midrule
 &  &  &  & Fs-CoT vs CP & 0.4846715  & 0.175   \\
            &   &    &   & Participants vs CP & 0.00001200018 & 0.365192 \\
\rowcolor{orange!15}
 \whitecell{} & Action Feasibility  &  0.00000005793772 & Yes  & Participants vs Fs-CoT & 0.0004419224 & 0.295096 \\
\cmidrule{2-7}
 &  &  &   & Fs-CoT vs CP  & 0.1539965  & 0.285   \\
 \textbf{Gemma-3-27B} &   &  &   & Participants vs CP & 0.000000000000 & 0.637788 \\
            & Action Optimality  & 0.000000000000  & Yes   & Participants vs Fs-CoT & 0.0000009109217 & 0.465\\
\cmidrule{2-7}
 &  & &   & Fs-CoT vs CP  & 0.1653792  &  0.325    \\
            &   &   &   & Participants vs CP & 0.00000004889773 & 0.606827 \\
\rowcolor{orange!15}            
 \whitecell{} & Plan Feasibility & 0.0000000002461595  & Yes  & Participants vs Fs-CoT & 0.0003950473 & 0.407404 \\
\bottomrule
\end{tabular}
}
\caption{Kruskal-Wallis and Dunn post hoc test results for continuous metrics. Dunn test pairwise comparisons include adjusted $p$-values using Holm-Bonferroni correction and effect sizes ($r$) using Rank-biserial correlation.}
\label{tab:kruskal_dunn}
\end{table}

\subsection{Intent Accuracy}
\label{sec:intent-metric}
The GPT-4o Fs-CoT variant achieved an improvement of \qtyP{70} over its CP counterpart and \qtyP{5.74} over the participants in intent recognition accuracy. In contrast, neither of the DeepSeek-R1 Tomcat variants outperformed the participants; however, the Fs-CoT variant achieved results comparable to the participants and demonstrated a \qtyP{33.33} improvement over its CP variant. Similarly, both Gemma-3-27B Tomcat variants underperformed relative to the participants, though its Fs-CoT variant achieved a substantial improvement of \qtyP{500} over its CP counterpart. 
\begin{description}[leftmargin=0em]
\item[Significance tests:] We assessed statistical significance for intent accuracy using McNemar's exact test to compare Tomcat variants (Table~\ref{tab:mcnemar}) and Chi-square tests to compare variants with participants (Table~\ref{tab:chi}). We report results for each LLM below:

  \begin{itemize}
    \item \textbf{GPT-4o:} We observed a significant difference between Fs-CoT and CP ($p = 0.039$), with a large effect size ($g = 0.78$), favoring Fs-CoT. We found substantial significant differences between participants and CP ($p = 0.0026$) with a small effect ($V = 0.13$), indicating CP's underperformance, whereas we observed no significant difference between participants and Fs-CoT ($p = 0.823$; $V = 0.01$).
    
    \item \textbf{DeepSeek-R1:} We observed no significant difference between Fs-CoT and CP ($p = 0.125$), despite a large effect size ($g = 1$), supporting Fs-CoT. Compared to the participants, we did not observe significant differences for either variant ($p = 0.053$ for CP; $p = 1$ for Fs-CoT), though CP approached significance with a small effect size ($V = 0.08$).
    
    \item \textbf{Gemma-3-27B:} We observed no significant difference between Fs-CoT and CP ($p = 0.063$), although the result was marginally nonsignificant with a large effect size ($g = 1$), favoring Fs-CoT. However, the test revealed that both variants differed significantly from the participants ($p < 0.001$), with moderate to small effect sizes ($V = 0.33$ for CP; $V = 0.22$ for Fs-CoT).
  \end{itemize}
 \item [Qualitative analysis:] In examples such as \textsl{Can you unlock this red door?} \textinred{(scenario 9)}, where both red and blue doors must be unlocked to reach the gem, the CP variants of GPT-4o and DeepSeek-R1 often misinterpret the instruction, treating the request as an isolated task rather than part of a coordinated effort to retrieve the gem beyond the red and blue doors. In contrast, their Fs-CoT counterparts correctly infer the underlying goal and generate responses aligned with that intention. Further, the Gemma-3-27B Tomcat variants generally fail to recognize the principal's ultimate objective. For instance, in following the instruction, \textsl{I'll get the blue key. Can you pick up a red key?} \textinred{(scenario 13)}, both variants perform the literal action of passing the red key without acknowledging the principal's next steps of gem retrieval. Although the Gemma-3-27B Fs-CoT variant shows a slight improvement in intent recognition, it does not generate a coherent plan in most cases. These patterns suggest that the CP variants of GPT-4o and DeepSeek-R1, as well as both Gemma-3-27B Tomcat variants, struggle with latent goal inference when instructions are indirect or rely on shared context. However, the Fs-CoT variants of GPT-4o and DeepSeek-R1 benefit from reasoning exemplars that bridge intention and action.

 \item [Participant analysis:] Most participants \textinred{correctly} identified the principal's intent. 
 On average, three to four participants per instruction failed to do so, possibly due to weaker ToM reasoning or misinterpretation of the task. Interestingly, even participants who typically performed well made mistakes in certain instances. For example, \textinred{in scenario 14 (Figure~\ref{fig:obs-high})}, the grid contained two red keys, one easily accessible but far from the principal-agent pair, and another located behind a locked blue door. 
 \textinred{The principal issued an instruction \fsl{Can you go get the red key?}
 as it} was moving toward a blue key, implying a multistep plan to unlock both the red and blue doors in order to retrieve a gem placed beyond them. This meant the appropriate response was to retrieve the accessible red key, pass it to the principal, and support the shared objective of unlocking the doors and retrieving the gem. However, some participants instead attempted to retrieve the red key behind the blue door or focused only on gems behind a single red door, missing the principal's broader intention.
 \textinred{Consequently, the average participants' performance was the worst for this scenario (Table~\ref{tab:complexity}}). 
 In contrast, both Fs-CoT variants of GPT-4o and DeepSeek-R1 correctly inferred the intended plan and coordinated their actions accordingly. Apart from these rare lapses, participants generally exhibited strong performance in inferring the principal's latent goals even under \textinred{unclear} instructions.     

 \item [Takeaways:] These findings underscore the effectiveness of Fs-CoT for LLMs like GPT-4o and DeepSeek-R1 in enabling Tomcat to infer the principal's desired gem from \textinred{incomplete} or ambiguous instructions, unlike CP, which lacks reasoning exemplars. The significance tests indicate that the Fs-CoT variants of GPT-4o and DeepSeek-R1 perform comparably with the participants. However, the Gemma-3-27B Fs-CoT variant \textinred{performed worse} than the participants and other LLMs. This may stem from its smaller model size, which could limit its capacity for complex reasoning required for Instruction Inference.

\end{description}

\subsection{Action Feasibility and Optimality}
\label{sec:action-metric}
For action feasibility, the GPT-4o Fs-CoT variant achieved an improvement of \qtyP{23.81} over its CP counterpart and \qtyP{9.23} over the participants. The DeepSeek-R1 Tomcat variants performed comparably to the participants, though with slightly lower performance. Both Gemma-3-27B Tomcat variants failed to reach the performance level of the participants, though its Fs-CoT variant showed an improvement of \qtyP{24.13} over CP.

For action optimality, the GPT-4o Fs-CoT variant demonstrated a considerable gain of \qtyP{68.18} over CP and \qtyP{7.28} over the participants. The DeepSeek-R1 Tomcat variants performed equally and yielded lower performance than the participants. Similarly, both Gemma-3-27B Tomcat variants underperformed relative to the participants, though its Fs-CoT variant achieved a notable gain of \qtyP{68.47} over CP.
\begin{description}[leftmargin=0em]
\item[Significance tests:] We assessed statistical significance for action feasibility and optimality using the Kruskal-Wallis test, followed by Dunn's post hoc comparisons where applicable (Table~\ref{tab:kruskal_dunn}). We report results per LLM below:

\begin{itemize}
\item \textbf{GPT-4o:}
For \emph{action feasibility}, the Kruskal-Wallis test identified significant differences across groups ($p = 0.003$). The Dunn tests showed that CP underperformed both Fs-CoT ($p = 0.005$, $r = 0.353$) and participants ($p = 0.003$, $r = 0.245$), with moderate to small effect sizes, respectively. However, the test revealed no significant difference between participants and Fs-CoT ($p = 0.332$, $r = -0.07$).

Similarly, for \emph{action optimality}, the Kruskal-Wallis test revealed notable significant differences across groups ($p < 0.001$). The Dunn tests demonstrated that CP substantially underperformed relative to both Fs-CoT ($p < 0.001$, $r = 0.545$) and participants ($p < 0.001$, $r = 0.452$), with large to moderate effect sizes, respectively. In contrast, the test revealed no significant difference between participants and Fs-CoT ($p = 0.752$, $r = -0.025$).

\item \textbf{DeepSeek-R1:}
For \emph{action feasibility}, the Kruskal-Wallis test revealed no significant differences across groups ($p = 0.254$); therefore, we did not conduct the Dunn tests.

For \emph{action optimality}, the test revealed significant differences across groups ($p = 0.017$). However, post hoc comparisons did not indicate significant differences between any of the pairs. That is, CP, Fs-CoT, and participants performed comparably, though participants scored higher than the Tomcat variants. 

\item \textbf{Gemma-3-27B:}
The Kruskal-Wallis tests for both \emph{action feasibility} and \emph{action optimality} revealed significant differences across groups ($p < 0.001$ for both). Dunn tests showed that participants outperformed both CP ($p < 0.001$, $r = 0.365$ for feasibility; $p < 0.001$, $r = 0.638$ for optimality) and Fs-CoT ($p < 0.001$, $r = 0.295$ for feasibility; $p < 0.001$, $r = 0.465$ for optimality), with moderate to large effect sizes. However, we observed no significant differences between Fs-CoT and CP in either metric ($p = 0.485$, $r = 0.175$ for feasibility; $p = 0.154$, $r = 0.285$ for optimality).

\end{itemize}
 \item [Qualitative analysis:] The CP variants of GPT-4o and Gemma-3-27B often produce actions that are technically feasible but suboptimal. For example, \textinred{in scenario 4}, \fsl{I'm picking up the yellow key. Can you get a red key?}, multiple red keys are available on the grid. However, the CP variants for both LLMs retrieve a poorly positioned red key that is feasible but not efficient. In contrast, the Fs-CoT variants select the more strategically placed red key, demonstrating an improved understanding of optimality. 

By comparison, both DeepSeek-R1 Tomcat variants execute actions similarly in most scenarios. \textinred{For instance, in scenario 1 (Figure~\ref{fig:obs-moderate})}---\textsl{Give me a red key for this door?}---retrieving the gem requires two red keys. However, both variants retrieve only the nearest key and generate a follow-up action claiming the second key is inaccessible due to walls, which is incorrect. Consequently, the DeepSeek-R1 Fs-CoT variant fails to improve over CP in terms of action feasibility and optimality. Although both variants struggled in specific edge cases, the overall performance of the CP variant was already strong in these metrics, which may have limited the observable gains from Fs-CoT.
 \item [Participant analysis:] Most participants who correctly inferred the principal's intent were also able to execute the actions optimally. In contrast to the suboptimal behavior of CP variants in the mentioned LLM cases---choosing a well-positioned red key or retrieving two red keys---participants consistently selected more efficient options. In both examples where the CP variants underperformed, human participants executed the tasks optimally.

 The few failure cases among participants typically stemmed from a misinterpretation of the task setup or difficulty interpreting unclear instructions. For example, \textinred{in scenario 18} \textsl{Can you help me unlock the blue door there?}, where two blue doors were present, a small number of participants unlocked the incorrect one. This occurred even though the principal's movements clearly implied that the blue door adjacent to the red door was part of a multistep plan to retrieve a gem. Such failures were rare, originating primarily under ambiguous instructions and affecting only a handful of participants. Overall, participant performance demonstrated strong action feasibility and optimality, particularly when the principal's intent was clearly understood.
 
 \item [Takeaways:] These results suggest that the Fs-CoT variants of GPT-4o and Gemma-3-27B generate more optimal and error-free actions over its CP counterpart. This improvement can be attributed to the structured reasoning provided by Fs-CoT demonstrations and the additional rules in the common ground component. However, this benefit does not extend to DeepSeek-R1, as both of its Tomcat variants behave similarly in action execution. Further, the statistical tests indicate that the GPT-4o Fs-CoT variant and both DeepSeek-R1 Tomcat variants perform comparably to participants, whereas both Gemma-3-27B Tomcat variants underperformed.    
\end{description}

\subsection{Plan Feasibility and Optimality}
\label{sec:plan-metric}
For plan feasibility, the GPT-4o Fs-CoT variant achieved an improvement of \qtyP{50.66} over its CP counterpart and \qtyP{16.14} over the participants. The DeepSeek-R1 Fs-CoT variant had a gain of \qtyP{16.56} over CP and performed slightly better than the participants. The Gemma-3-27B Fs-CoT variant also exhibited an improvement of \qtyP{52.51} over CP, though both variants fell short of participant performance.

For plan optimality, the GPT-4o Fs-CoT variant demonstrated a substantial gain of \qtyP{200} over CP and a modest gain of \qtyP{5.41} over the participants. The DeepSeek-R1 Fs-CoT variant improved by \qtyP{27.27} over CP but remained slightly below participant performance. Further, both Gemma-3-27B Tomcat variants achieved worse results than the participants, though the Fs-CoT variant achieved a notable gain of \qtyP{500} over CP.

\begin{description}[leftmargin=0em]
\item[Significance tests:] We evaluated statistical significance for plan optimality using McNemar's exact test to compare Tomcat variants (Table~\ref{tab:mcnemar}) and Chi-square tests to compare each variant with participants (Table~\ref{tab:chi}). For plan feasibility, we used the Kruskal-Wallis test, followed by Dunn's post hoc comparisons where applicable (Table~\ref{tab:kruskal_dunn}). We report results per LLM below:

\begin{itemize}
\item \textbf{GPT-4o:}
For \emph{plan feasibility}, the Kruskal-Wallis test revealed significant differences across groups ($p = 0.003$). The Dunn tests indicated that CP underperformed both Fs-CoT ($p = 0.004$, $r = 0.55$) and participants ($p = 0.004$, $r = 0.336$), with large to moderate effect sizes, respectively. However, there was no significant difference between participants and Fs-CoT ($p = 0.277$, $r = -0.11$).

For \emph{plan optimality}, McNemar's exact test found a significant difference between Fs-CoT and CP ($p = 0.002$, $g = 1$), with a large effect size. The Chi-square test demonstrated a notable significant difference between participants and CP ($p < 0.001$, $V = 0.179$), with a small effect, but no significant difference between participants and Fs-CoT($p = 0.903$, $V = 0.005$).

\item \textbf{DeepSeek-R1:}
For \emph{plan feasibility}, the Kruskal-Wallis test revealed no significant differences across groups ($p = 0.445$); hence, Dunn tests were not performed.

For \emph{plan optimality}, McNemar's exact test indicated no significant difference between Fs-CoT and CP ($p = 0.375$, $g = 0.6$), despite a large effect size. Similarly, Chi-square results revealed no significant differences between participants and either variant ($p = 0.192$ for CP, $p = 1$ for Fs-CoT).

\item \textbf{Gemma-3-27B:}
For \emph{plan feasibility}, the Kruskal-Wallis test revealed significant differences across groups ($p < 0.001$). The Dunn tests showed that participants outperformed both CP ($p < 0.001$, $r = 0.607$) and Fs-CoT ($p < 0.001$, $r = 0.407$), with large to moderate effect sizes, respectively. We observed no significant difference between Fs-CoT and CP ($p = 0.165$, $r = 0.325$), despite a moderate effect size favoring Fs-CoT.

For \emph{plan optimality}, McNemar's exact test found no significant difference between Fs-CoT and CP ($p = 0.063$), although the result approached significance with a large effect size ($g = 1$), supporting Fs-CoT. However, the Chi-square test, showed that participants outperformed both CP ($p < 0.001$, $V = 0.259$) and Fs-CoT ($p < 0.001$, $V = 0.158$), with moderate to small effect sizes, respectively.
\end{itemize}

 \item [Qualitative analysis:] The CP variants of GPT-4o and DeepSeek-R1 often generate plans that are feasible but lack goal-directed coherence. Consider the example \textinred{in scenario 18}, \textsl{Can you help me unlock the blue door there?}, where multiple blue doors are present. Based on the principal's movements, the blue door adjacent to the red door must be unlocked to retrieve the gem. The GPT-4o CP variant passes the key to the principal without unlocking the door, whereas the DeepSeek-R1 CP variant unlocks a different blue door. In contrast, their Fs-CoT counterparts generate plans that enable the principal to proceed seamlessly through both doors and retrieve the gem.
 
 In another example \textinred{(scenario 13)}, \textsl{I'll get the blue key. Can you pick up a red key?}, multiple red keys are available, but the one closest to the blue door must be selected for efficient coordination. The GPT-4o CP variant retrieves a distant red key, failing to generate a coherent plan for gem retrieval, whereas its Fs-CoT counterpart collects the optimal key. In contrast, both DeepSeek-R1 Tomcat variants execute the action correctly by picking up the optimal red key. This suggests that the DeepSeek-R1 CP variant generates more coherent plans than the GPT-4o CP variant. These observations indicate that both Fs-CoT variants of GPT-4o and DeepSeek-R1---and, in \textinred{some} cases, the DeepSeek-R1 CP variant---model long-term dependencies and shared intentions, resulting in plans that are both feasible and optimal. 
 
 However, the Gemma-3-27B Tomcat variants in both examples execute only the immediate action without generating a plan that considers the principal's intended next steps. Although the Fs-CoT variant improves plan feasibility and optimality for Gemma-3-27B, its overall planning performance trails substantially behind the other LLMs.

 \item [Participant analysis:] In both examples discussed---unlocking the correct blue door and retrieving the red key adjacent to the blue door---participants' plans were generally feasible and aligned with the principal's objective. However, a few responses were suboptimal, primarily because they omitted references to the principal's follow-up actions for gem retrieval. Even when participants identified the correct key or door, their responses occasionally lacked the broader coordination necessary for optimal planning.

Additional failures occurred when some participants misunderstood the principal's intent entirely, leading to incorrect goal identification and less effective action choices in planning. Nevertheless, participant performance consistently surpassed the CP variants, which frequently failed to generate coherent plans. In terms of plan feasibility and optimality, the Fs-CoT variants of GPT-4o and DeepSeek-R1 closely matched participant performance.   

 \item [Takeaways:] These findings indicate that across all LLMs, Tomcat with Fs-CoT outperformed its CP counterpart in generating feasible and optimal plans. Although we observed no significant difference in the DeepSeek-R1 Tomcat variants, its Fs-CoT variant exhibited substantial gains over CP. Further, since effective planning requires anticipating the principal's future actions, Fs-CoT variants generally demonstrated stronger ToM reasoning than CP. Additionally, the significance tests show that the GPT-4o Fs-CoT variant and both DeepSeek-R1 Tomcat variants performed on par with human participants and exhibited more robust ToM reasoning, whereas both Gemma-3-27B Tomcat variants lagged behind. We attribute this comparable performance to the structured reasoning exemplars in Fs-CoT, which provide a \emph{step-by-step} process for inferring the principal's intention. The underperformance of Gemma-3-27B may be due to its smaller model size, which limits its ability to handle the complexity of ToM reasoning. 
\end{description}

\begin{table}[htb]
    \sisetup{
  text-family-to-math = true,
  text-series-to-math = true,
  round-precision=2,
  round-mode=places
}
\rowcolors{2}{mint!15}{white}
    \centering
    \resizebox{0.9\textwidth}{!}{
    \begin{tabular}{ll *4{S[table-format=1.2]}}
        \toprule
        \textbf{LLMs} & \textbf{Method} & \textbf{Precision} & \textbf{Recall} & \textbf{F1-score} & \textbf{Accuracy}\\ 
        \midrule 
        & CP & 0.600000 & \bfseries 1.000000 & 0.750000 & 0.600000   \\ 
       \cellcolor{white}\textbf{GPT-4o} & Fs-CoT  & 0.705882 & \bfseries 1.000000 & \bfseries 0.83 & \bfseries 0.750000 \\ 
        \midrule
        & CP & 0.578947 & 0.916667 & 0.709677 & 0.550000   \\ 
        \cellcolor{white}\textbf{DeepSeek-R1} & Fs-CoT  & \bfseries 0.733333 & 0.916667 & 0.814815 & \bfseries 0.750000 \\
        \midrule
         
        & CP & 0.555556 & 0.833333 & 0.666667 & 0.500000   \\ 
        \cellcolor{white}\textbf{Gemma-3-27B} & Fs-CoT  & 0.666667 & 0.833333 & 0.740741 & 0.650000 \\
        \bottomrule
    \end{tabular}
     }
    \caption{Instruction identification metrics for Tomcat.}
    \label{tab:instruction_type_accuracy}
\end{table}

\begin{table}[t]
\sisetup{
  text-family-to-math = true,
  text-series-to-math = true,
  round-precision=2,
  round-mode=places
}
\centering
\resizebox{\textwidth}{!}{
\begin{tabular}{ll l *3{S[table-format=1.2, input-symbols={()\%}]}}
\toprule
\textbf{LLMs} & \textbf{Method} & \textbf{Metric} & {\textbf{Low}\;($n$=2)} & {\textbf{Moderate}\;($n$=10)} & {\textbf{High}\;($n$=8)} \\
\midrule
 &  & Intent Accuracy  & 0.50 & 0.40 & 0.625 \\
 &  & Plan Feasibility    & 0.83 & 0.564 & 0.61 \\
 \rowcolor{ncsuBlue!15}
 \cellcolor{white}& CP & Plan Optimality & 0.50 & 0.30 & 0.125 \\
\cmidrule{2-6}
 & & Intent Accuracy    & 0.50 & 0.90 & 0.875 \\
 & & Plan Feasibility    & 0.625 & 0.931 & 0.975 \\
 \rowcolor{ncsuBlue!15}
 \cellcolor{white}\multirow{-6}{*}{\textbf{GPT-4o}}& Fs-CoT & Plan Optimality  & 0.50 & 0.70 & 0.875 \\
\midrule
 &  & Intent Accuracy  & 0.50 & 0.60 & 0.625 \\
 &  & Plan Feasibility    & 0.833 & 0.649 & 0.728 \\
 \rowcolor{ncsuBlue!15}
 \cellcolor{white}& CP & Plan Optimality & 0.50 & 0.50 & 0.625 \\
\cmidrule{2-6}
 & & Intent Accuracy    & 0.50 & 0.80 & 0.875 \\
 & & Plan Feasibility    & 0.75 & 0.730 & 0.911 \\
 \rowcolor{ncsuBlue!15}
 \cellcolor{white}\multirow{-6}{*}{\textbf{DeepSeek-R1}}& Fs-CoT & Plan Optimality  & 0.50 & 0.60 & 0.875 \\
\midrule
 &  & Intent Accuracy  & 0.50 & 0.00 & 0.00 \\
 &  & Plan Feasibility    & 0.625 & 0.269 & 0.323 \\
 \rowcolor{ncsuBlue!15}
 \cellcolor{white}& CP & Plan Optimality & 0.50 & 0.00 & 0.00 \\
\cmidrule{2-6}
 & & Intent Accuracy    & 0.50 & 0.30 & 0.25 \\
 & & Plan Feasibility    & 0.625 & 0.508 & 0.453 \\
 \rowcolor{ncsuBlue!15}
 \cellcolor{white}\multirow{-6}{*}{\textbf{Gemma-3-27B}}& Fs-CoT & Plan Optimality  & 0.50 & 0.30 & 0.25 \\
\bottomrule
\end{tabular}
}
\caption{Mean performance of intent accuracy and planning metrics of each Tomcat variant stratified by obstruction level ($n$ denotes the number of scenarios per level).}
\label{tab:perf_obstruction}
\end{table}

\subsection{Instruction Accuracy}
\label{sec:instruction-metric}
Table~\ref{tab:instruction_type_accuracy} presents Tomcat's instruction identification metrics with Fs-CoT and CP across all LLMs. Both demonstrate excellent recall, demonstrating their ability to recognize \textinred{incomplete} or ambiguous instructions. However, their precision is imperfect. For GPT-4o and DeepSeek-R1, the Fs-CoT variant occasionally misclassified clear instructions as unclear, whereas their CP counterpart underperformed further, misclassifying several clear instructions as unclear. Gemma-3-27B shows similar trends, with lower precision across both variants, though Fs-CoT still achieves higher accuracy and F1-score than CP. This misclassification hurt CP's performance on other metrics. In contrast, despite its lower precision, the Fs-CoT variants excelled in other metrics, suggesting that its enhanced ToM capabilities, bolstered by reasoning exemplars, enable it to better infer principal's intentions and optimize actions and plans. 

Further, McNemar's exact test did not reveal statistically significant differences in instruction accuracy between Fs-CoT and CP for any of the LLMs. Specifically, the results were as follows: GPT-4o ($p = 0.25$, $g = 1.00$), DeepSeek-R1 ($p = 0.125$, $g = 1.00$), and Gemma-3-27B ($p = 0.375$, $g = 0.60$). Despite the lack of significance, the large or moderate effect sizes suggest that Tomcat with Fs-CoT outperforms CP in instruction identification, which is further corroborated by Table~\ref{tab:instruction_type_accuracy}.

\subsection{Impact of Obstruction Level}
\label{sec:obstruction-analysis}

Table~\ref{tab:perf_obstruction} reports mean intent accuracy and planning metrics for each Tomcat variant, stratified by obstruction level. Because the Low obstruction level contains only two scenarios ($n = 2$), binary metric scores in that column are constrained to multiples of 0.5 and offer limited comparative value; therefore, the following discussion focuses on the Moderate and High levels.

For GPT-4o, the CP variant's intent accuracy and plan feasibility increase from Moderate to High obstruction, whereas its plan optimality declines across the same range. 
In contrast, the Fs-CoT variant's plan feasibility and plan optimality rise from Moderate to High, whereas its intent accuracy shows a marginal decline over the same range.

For DeepSeek-R1, both variants exhibit consistent increases across all three metrics from Moderate to High obstruction, with the Fs-CoT variant recording higher absolute values throughout.
For Gemma-3-27B, the CP variant records zero intent accuracy and plan optimality at both obstruction levels, whereas its plan feasibility shows a slight increase from Moderate to High. In contrast, the Fs-CoT variant exhibits a slight uniform decline across all three metrics over the same range.

The direction and magnitude of change from Moderate to High are inconsistent across models for intent recognition and planning metrics. 
Thus, no consistent monotonic trend emerges for any LLM or metric, indicating that Tomcat's performance is determined primarily by instruction ambiguity and prompting strategy rather than by the structural complexity of the grid.

\subcaptionsetup{width=0.9\linewidth}
\begin{figure}[t]

\begin{subfigure}[t]{0.49\textwidth}
\begin{tikzpicture}
\begin{axis}[
    xbar,
    enlarge y limits=0.2,
    ymax=1,
    ymin=0,
    y axis line style = {opacity = 0},
    enlarge x limits=0.3,
    ytick={0, 0.24, 0.53, 0.83},
    yticklabels={{Participants},{Gemma-3-27B}, {Deepseek-R1}, {GPT-4o}},
    yticklabel style={font=\bfseries\footnotesize},
    ytick style={draw=none},
    xlabel style={font=\bfseries},
    xlabel={Intent Accuracy},
    xmin=0.206,
    xmax=0.9,
    axis x line*=bottom,
    legend style={at={(0.4,0.99)}, anchor=north, legend columns=3},
    cycle list={
        {ncsuDarkPurple, fill=ncsuLightPurple, draw=none},
        {ncsuDarkGreen, fill=ncsuLightGreen, draw=none},
        {ncsuDarkBlue, fill=ncsuLightBlue,draw=none},
        {ncsuDarkGreen, fill=ncsuLightGreen, draw=none},
        {ncsuDarkBlue, fill=ncsuLightBlue,draw=none},
        {ncsuDarkGreen, fill=ncsuLightGreen, draw=none},
        {ncsuDarkBlue, fill=ncsuLightBlue,draw=none},
   }
]

\addlegendimage{ncsuDarkBlue, fill=ncsuLightBlue}
\addlegendimage{ncsuDarkGreen, fill=ncsuLightGreen} 
\addlegendimage{ncsuDarkPurple, fill=ncsuLightPurple} 
\legend{Tomcat\_CP, Tomcat\_Fs-CoT, Participants}

\addplot+ [
        error bars/.cd,
        x dir=both,
        x explicit,
        error bar style={line width=0.6pt}
] coordinates {
    (0.8038,0.32) +- (0.017430, 0)
};

\addplot+ [
        error bars/.cd,
        x dir=both,
        x explicit,
        error bar style={line width=0.6pt}
] coordinates {
    (0.30,0.4) +- (0.105131, 0)
};
\addplot+ [
        error bars/.cd,
        x dir=both,
        x explicit,
        error bar style={line width=0.6pt}
] coordinates {
    (0.05,0.4) +- (0.05, 0)
};
\addplot+ [
        error bars/.cd,
        x dir=both,
        x explicit,
        error bar style={line width=0.6pt}
] coordinates {
    (0.80,0.48) +- (0.091766, 0)
};
\addplot+ [
        error bars/.cd,
        x dir=both,
        x explicit,
        error bar style={line width=0.6pt}
] coordinates {
    (0.60,0.48) +- (0.11239, 0)
};
\addplot+ [
        error bars/.cd,
        x dir=both,
        x explicit,
        error bar style={line width=0.6pt}
] coordinates {
    (0.85,0.56) +- (0.081918, 0)
};
\addplot+ [
        error bars/.cd,
        x dir=both,
        x explicit,
        error bar style={line width=0.6pt}
] coordinates {
    (0.50,0.56) +- (0.114708, 0)
};

\node at (axis cs:0.61,0.875) [anchor=west, font=\footnotesize] {0.11};
\node at (axis cs:0.93,0.77) [anchor=west, font=\footnotesize] {0.08};
\node at (axis cs:0.71,0.585) [anchor=west, font=\footnotesize] {0.11};
\node at (axis cs:0.89,0.48) [anchor=west, font=\footnotesize] {0.09};
\node at (axis cs:0.1,0.3) [anchor=west, font=\footnotesize] {0.05};
\node at (axis cs:0.4,0.2) [anchor=west, font=\footnotesize] {0.08};
\node at (axis cs:0.82,0.01) [anchor=west, font=\footnotesize] {0.01};

\end{axis}
\end{tikzpicture}
\caption{Intent accuracy across LLMs and participants.}
\label{fig:ia-visual}
\end{subfigure}
\hfill
\begin{subfigure}[t]{0.49\textwidth}
\begin{tikzpicture}
\begin{axis}[
    xbar,
    enlarge y limits=0.2,
    ymax=1,
    ymin=0,
    y axis line style = { opacity = 0 },
    enlarge x limits=0.3,
    ytick={0, 0.24, 0.53, 0.83},
    yticklabels={{Participants},{Gemma-3-27B}, {Deepseek-R1}, {GPT-4o}},
    yticklabel style={font=\bfseries\footnotesize},
    ytick style={draw=none},
    grid style={dashed, ncsuDarkGray},
    xlabel style={font=\bfseries},
    xlabel={Plan Optimality},
    xmin=0.206,
    xmax=0.9,
    axis x line*=bottom,
    legend style={at={(0.4,0.99)}, anchor=north, legend columns=3},
    cycle list={
        {ncsuDarkPurple, fill=ncsuLightPurple, draw=none},
        {ncsuDarkGreen, fill=ncsuLightGreen, draw=none},
        {ncsuDarkBlue, fill=ncsuLightBlue,draw=none},
        {ncsuDarkGreen, fill=ncsuLightGreen, draw=none},
        {ncsuDarkBlue, fill=ncsuLightBlue,draw=none},
        {ncsuDarkGreen, fill=ncsuLightGreen, draw=none},
        {ncsuDarkBlue, fill=ncsuLightBlue,draw=none},
   }
]

\addlegendimage{ncsuDarkBlue, fill=ncsuLightBlue}
\addlegendimage{ncsuDarkGreen, fill=ncsuLightGreen} 
\addlegendimage{ncsuDarkPurple, fill=ncsuLightPurple} 
\legend{Tomcat\_CP, Tomcat\_Fs-CoT, Participants}
    


\addplot+ [
        error bars/.cd,
        x dir=both,
        x explicit,
        error bar style={line width=0.6pt}
] coordinates {
    (0.7115,0.32) +- (0.019887, 0)
};

\addplot+ [
        error bars/.cd,
        x dir=both,
        x explicit,
        error bar style={line width=0.6pt}
] coordinates {
    (0.30,0.4) +- (0.105131, 0)
};
\addplot+ [
        error bars/.cd,
        x dir=both,
        x explicit,
        error bar style={line width=0.6pt}
] coordinates {
    (0.05,0.4) +- (0.05, 0)
};
\addplot+ [
        error bars/.cd,
        x dir=both,
        x explicit,
        error bar style={line width=0.6pt}
] coordinates {
    (0.70,0.48) +- (0.105131, 0)
};
\addplot+ [
        error bars/.cd,
        x dir=both,
        x explicit,
        error bar style={line width=0.6pt}
] coordinates {
    (0.55,0.48) +- (0.114133, 0)
};
\addplot+ [
        error bars/.cd,
        x dir=both,
        x explicit,
        error bar style={line width=0.6pt}
] coordinates {
    (0.75,0.56) +- (0.099340, 0)
};
\addplot+ [
        error bars/.cd,
        x dir=both,
        x explicit,
        error bar style={line width=0.6pt}
] coordinates {
    (0.25,0.56) +- (0.099340, 0)
};

\node at (axis cs:0.345,0.875) [anchor=west, font=\footnotesize] {0.09};
\node at (axis cs:0.84,0.77) [anchor=west, font=\footnotesize] {0.09};
\node at (axis cs:0.66,0.585) [anchor=west, font=\footnotesize] {0.11};
\node at (axis cs:0.8,0.48) [anchor=west, font=\footnotesize] {0.10};
\node at (axis cs:0.1,0.3) [anchor=west, font=\footnotesize] {0.05};
\node at (axis cs:0.4,0.2) [anchor=west, font=\footnotesize] {0.10};
\node at (axis cs:0.73,0.01) [anchor=west, font=\footnotesize] {0.01};

\end{axis}
\end{tikzpicture}
\caption{Plan optimality across LLMs and participants.}
\label{fig:po-visual}
\end{subfigure}
\centering
\begin{subfigure}[t]{0.49\linewidth}
\centering
\begin{tikzpicture}
\begin{axis}[
    xbar,
    enlarge y limits=0.55,
    y axis line style = {opacity=0},
    enlarge x limits=0.3,
    symbolic y coords={Gemma-3-27B,Deepseek-R1,GPT-4o}, 
    ytick=data,
    yticklabel style={font=\bfseries\footnotesize, yshift=-14pt},
    ytick style={draw=none},
    xlabel style={font=\bfseries},
    xlabel={Instruction Accuracy},
    xmin=0.206,
    xmax=0.9,
    axis x line*=bottom,
    legend style={at={(0.4,0.9)}, anchor=north, legend columns=2},
    cycle list={
        {ncsuDarkGreen, fill=ncsuLightGreen, draw=none},
        {ncsuDarkBlue, fill=ncsuLightBlue, draw=none},
   }
]

\addlegendimage{ncsuDarkBlue, fill=ncsuLightBlue, draw=none}
\addlegendimage{ncsuDarkGreen, fill=ncsuLightGreen, draw=none} 
\legend{Tomcat\_CP, Tomcat\_Fs-CoT}

\addplot+ [
        bar shift=-20pt,
        error bars/.cd,
        x dir=both,
        x explicit,
        error bar style={line width=0.6pt}
] coordinates {
    (0.65,Gemma-3-27B) +- (0.109424, 0)
    (0.75,Deepseek-R1) +- (0.099340, 0)
    (0.75,GPT-4o) +- (0.099340, 0)
};

\addplot+ [
        bar shift=-8pt,
        error bars/.cd,
        x dir=both,
        x explicit,
        error bar style={line width=0.6pt}
] coordinates {
    (0.50,Gemma-3-27B) +- (0.114708, 0)
    (0.55,Deepseek-R1) +- (0.114133, 0)
    (0.60,GPT-4o) +- (0.112390, 0)
};

\node at (axis cs:0.71,GPT-4o) [anchor=west, font=\footnotesize, yshift=-7.5pt] {0.11};
\node at (axis cs:0.85,GPT-4o) [anchor=west, font=\footnotesize, yshift=-20pt] {0.09};
\node at (axis cs:0.66,Deepseek-R1) [anchor=west, font=\footnotesize,yshift=-7.5pt] {0.11};
\node at (axis cs:0.85,Deepseek-R1) [anchor=west, font=\footnotesize,yshift=-20pt] {0.09};
\node at (axis cs:0.61,Gemma-3-27B) [anchor=west, font=\footnotesize,yshift=-7.5pt] {0.11};
\node at (axis cs:0.76,Gemma-3-27B) [anchor=west, font=\footnotesize,yshift=-20pt] {0.10};

\end{axis}
\end{tikzpicture}
\caption{Instruction accuracy across LLMs.}
\label{fig:ina-visual}
\end{subfigure}

\caption{Binary metrics with standard error of mean (SEM). Participants were not tasked with classifying the instruction type.}
\label{fig:result-barplot}
\end{figure}

\subcaptionsetup{width=0.9\linewidth}
\begin{figure}[t]

\begin{subfigure}[t]{0.49\textwidth}
\begin{tikzpicture}[thick]

\foreach \y in {0,0.25,...,1.0} {
  \pgfmathsetmacro{\ytick}{\y*5}
  \draw (-0.1,\ytick) -- (0.1,\ytick);
  \node[left] at (-0.2,\ytick) {\small \y};
}

\def\xGPTa{1.0}     
\def\xGPTb{2.0}     
\def\xRa{3.6}       
\def\xRb{4.6}       
\def\xGemmaa{6.2}   
\def\xGemmab{7.2}   
\def\xP{9}        

\def\yscale{5}

\newcommand{\boxplotfull}[9]{%
  \pgfmathsetmacro{\qone}{#2*\yscale}
  \pgfmathsetmacro{\qthree}{#3*\yscale}
  \pgfmathsetmacro{\med}{#4*\yscale}
  \pgfmathsetmacro{\min}{#5*\yscale}
  \pgfmathsetmacro{\max}{#6*\yscale}
  \filldraw[fill=#7, draw=#8] (#1-0.3,\qone) rectangle (#1+0.3,\qthree);
  \draw[#8, thick] (#1-0.3,\med) -- (#1+0.3,\med);
  \draw[draw=#8] (#1,\qone) -- (#1,\min);
  \draw[draw=#8] (#1,\qthree) -- (#1,\max);
  \draw[draw=#8] (#1-0.1,\min) -- (#1+0.1,\min);
  \draw[draw=#8] (#1-0.1,\max) -- (#1+0.1,\max);
  \foreach \val in {#9} {
    \pgfmathsetmacro{\oy}{\val*\yscale}
    \draw[fill=#7, draw=#8] (#1,\oy) circle (0.06cm);
  }
}

\boxplotfull{\xGPTa}{0.5}{1.0}{1.0}{0.25}{1.0}{ncsuLightBlue}{ncsuDarkBlue}{}
\boxplotfull{\xGPTb}{1.0}{1.0}{1.0}{1.0}{1.0}{ncsuLightGreen}{ncsuDarkGreen}{0.5}
\boxplotfull{\xRa}{0.88}{1.0}{1.0}{0.88}{1.0}{ncsuLightBlue}{ncsuDarkBlue}{0.5,0.2,0.5,0.5,0.5}
\boxplotfull{\xRb}{1.0}{1.0}{1.0}{1.0}{1.0}{ncsuLightGreen}{ncsuDarkGreen}{0.5,0.2,0.5,0.5}
\boxplotfull{\xGemmaa}{0.19}{1.0}{0.75}{0}{1.0}{ncsuLightBlue}{ncsuDarkBlue}{}
\boxplotfull{\xGemmab}{0.5}{1.0}{1.0}{0}{1.0}{ncsuLightGreen}{ncsuDarkGreen}{}
\boxplotfull{\xP}{1.0}{1.0}{1.0}{1.0}{1.0}{ncsuLightPurple}{ncsuDarkPurple}{0.0,0.0,0.0,0.0,0.0,0.0,0.0,0.0,0.0,0.0,0.0,0.0,0.0,0.0,0.0,0.5,0.5,0.0,0.0,0.5,0.0,0.0,0.0,0.0,0.0,0.67,0.0,0.0,0.0,0.0,0.0,0.0,0.0,0.0,0.0,0.0,0.5,0.0,0.0,0.0,0.0,0.0,0.0,0.0,0.0,0.0,0.5,0.0,0.0,0.0,0.0,0.0,0.0,0.0,0.5,0.0,0.0,0.5,0.0,0.5,0.5}

\node[below, xshift=1.5em,yshift=-0.5em] at (\xGPTa,0) {\small \textbf{GPT-4o}};
\node[below, xshift=1.3em,yshift=-0.5em] at (\xRa,0) {\small \textbf{Deepseek-R1}};
\node[below, xshift=1.3em,yshift=-0.5em] at (\xGemmaa,0) {\small \textbf{Gemma-3-27B}};
\node[below, yshift=-0.5em] at (\xP,0) {\small \textbf{Participants}};

\node[rotate=90] at (-1.4,2.5) {\large \textbf{Action Feasibility}};
\begin{scope}[shift={(-0.5,5.8)}] 
  \draw[thin] (0.38,-0.3) rectangle (10.43,0.3);
  \draw[fill=ncsuLightBlue, draw=ncsuDarkBlue] (0.5,-0.1) rectangle +(0.7,0.3);
  \node[right=4pt] at (1,0.05) {\large Tomcat\_CP};

  \draw[fill=ncsuLightGreen, draw=ncsuDarkGreen] (3.6,-0.1) rectangle +(0.7,0.3);
  \node[right=4pt] at (4.1,0.05) {\large Tomcat\_Fs-CoT};

    \draw[fill=ncsuLightPurple, draw=ncsuDarkPurple] (7.35,-0.1) rectangle +(0.7,0.3);
  \node[right=4pt] at (7.85,0.015) {\large Participants};
\end{scope}

\end{tikzpicture}
\caption{Action feasibility across LLMs and participants.}
\label{fig:af-visual}
\end{subfigure}
\hfill
\begin{subfigure}[t]{0.49\textwidth}
\begin{tikzpicture}[thick]

\foreach \y in {0,0.25,...,1.0} {
  \pgfmathsetmacro{\ytick}{\y*5}
  \draw (-0.1,\ytick) -- (0.1,\ytick);
  \node[left] at (-0.2,\ytick) {\small \y};
}

\def\xGPTa{1.0}     
\def\xGPTb{2.0}     
\def\xRa{3.6}       
\def\xRb{4.6}       
\def\xGemmaa{6.2}   
\def\xGemmab{7.2}   
\def\xP{9}        

\def\yscale{5}

\newcommand{\boxplotfull}[9]{%
  \pgfmathsetmacro{\qone}{#2*\yscale}
  \pgfmathsetmacro{\qthree}{#3*\yscale}
  \pgfmathsetmacro{\med}{#4*\yscale}
  \pgfmathsetmacro{\min}{#5*\yscale}
  \pgfmathsetmacro{\max}{#6*\yscale}
  \filldraw[fill=#7, draw=#8] (#1-0.3,\qone) rectangle (#1+0.3,\qthree);
  \draw[#8, thick] (#1-0.3,\med) -- (#1+0.3,\med);
  \draw[draw=#8] (#1,\qone) -- (#1,\min);
  \draw[draw=#8] (#1,\qthree) -- (#1,\max);
  \draw[draw=#8] (#1-0.1,\min) -- (#1+0.1,\min);
  \draw[draw=#8] (#1-0.1,\max) -- (#1+0.1,\max);
  \foreach \val in {#9} {
    \pgfmathsetmacro{\oy}{\val*\yscale}
    \draw[fill=#7, draw=#8] (#1,\oy) circle (0.06cm);
  }
}

\boxplotfull{\xGPTa}{0.25}{1.0}{0.5}{0.0}{1.0}{ncsuLightBlue}{ncsuDarkBlue}{}
\boxplotfull{\xGPTb}{1.0}{1.0}{1.0}{1.0}{1.0}{ncsuLightGreen}{ncsuDarkGreen}{0.5,0.5,0.5}
\boxplotfull{\xRa}{0.5}{1.0}{1.0}{0.0}{1.0}{ncsuLightBlue}{ncsuDarkBlue}{}
\boxplotfull{\xRb}{0.5}{1.0}{1.0}{0.0}{1.0}{ncsuLightGreen}{ncsuDarkGreen}{}
\boxplotfull{\xGemmaa}{0.0}{0.5}{0.0}{0}{1.0}{ncsuLightBlue}{ncsuDarkBlue}{}
\boxplotfull{\xGemmab}{0.0}{1.0}{0.5}{0}{1.0}{ncsuLightGreen}{ncsuDarkGreen}{}
\boxplotfull{\xP}{1.0}{1.0}{1.0}{1.0}{1.0}{ncsuLightPurple}{ncsuDarkPurple}{0.0, 0.0, 0.0, 0.0, 0.0, 0.0, 0.0, 0.0, 0.0, 0.0, 0.0, 0.0, 0.0, 0.0, 0.0, 0.0, 0.0, 0.0, 0.0, 0.5, 0.0, 0.0, 0.33, 0.5, 0.5, 0.0, 0.0, 0.0, 0.0, 0.0, 0.67, 0.0, 0.0, 0.0, 0.0, 0.33, 0.0, 0.0, 0.0, 0.0, 0.0, 0.0, 0.5, 0.0, 0.0, 0.0, 0.0, 0.0, 0.0, 0.0, 0.5, 0.0, 0.0, 0.0, 0.0, 0.0, 0.0, 0.0, 0.0, 0.0, 0.0, 0.0, 0.0, 0.0, 0.0, 0.0, 0.0, 0.5, 0.0, 0.0, 0.33, 0.5, 0.5, 0.0, 0.5, 0.5, 0.5, 0.67, 0.67, 0.67, 0.67, 0.5}

\node[below, xshift=1.5em,yshift=-0.5em] at (\xGPTa,0) {\small \textbf{GPT-4o}};
\node[below, xshift=1.3em,yshift=-0.5em] at (\xRa,0) {\small \textbf{Deepseek-R1}};
\node[below, xshift=1.3em,yshift=-0.5em] at (\xGemmaa,0) {\small \textbf{Gemma-3-27B}};
\node[below, yshift=-0.5em] at (\xP,0) {\small \textbf{Participants}};

\node[rotate=90] at (-1.4,2.5) {\large \textbf{Action Optimality}};
\begin{scope}[shift={(-0.5,5.8)}] 
  \draw[thin] (0.38,-0.3) rectangle (10.43,0.3);
  \draw[fill=ncsuLightBlue, draw=ncsuDarkBlue] (0.5,-0.1) rectangle +(0.7,0.3);
  \node[right=4pt] at (1,0.05) {\large Tomcat\_CP};

  \draw[fill=ncsuLightGreen, draw=ncsuDarkGreen] (3.6,-0.1) rectangle +(0.7,0.3);
  \node[right=4pt] at (4.1,0.05) {\large Tomcat\_Fs-CoT};

    \draw[fill=ncsuLightPurple, draw=ncsuDarkPurple] (7.35,-0.1) rectangle +(0.7,0.3);
  \node[right=4pt] at (7.85,0.015) {\large Participants};
\end{scope}

\end{tikzpicture}
\caption{Action optimality across LLMs and participants.}
\label{fig:ao-visual}
\end{subfigure}
\centering
\begin{subfigure}[t]{0.49\linewidth}
\centering
\begin{tikzpicture}[thick]

\foreach \y in {0,0.25,...,1.0} {
  \pgfmathsetmacro{\ytick}{\y*5}
  \draw (-0.1,\ytick) -- (0.1,\ytick);
  \node[left] at (-0.2,\ytick) {\small \y};
}

\def\xGPTa{1.0}     
\def\xGPTb{2.0}     
\def\xRa{3.6}       
\def\xRb{4.6}       
\def\xGemmaa{6.2}   
\def\xGemmab{7.2}   
\def\xP{9}        

\def\yscale{5}

\newcommand{\boxplotfull}[9]{%
  \pgfmathsetmacro{\qone}{#2*\yscale}
  \pgfmathsetmacro{\qthree}{#3*\yscale}
  \pgfmathsetmacro{\med}{#4*\yscale}
  \pgfmathsetmacro{\min}{#5*\yscale}
  \pgfmathsetmacro{\max}{#6*\yscale}
  \filldraw[fill=#7, draw=#8] (#1-0.3,\qone) rectangle (#1+0.3,\qthree);
  \draw[#8, thick] (#1-0.3,\med) -- (#1+0.3,\med);
  \draw[draw=#8] (#1,\qone) -- (#1,\min);
  \draw[draw=#8] (#1,\qthree) -- (#1,\max);
  \draw[draw=#8] (#1-0.1,\min) -- (#1+0.1,\min);
  \draw[draw=#8] (#1-0.1,\max) -- (#1+0.1,\max);
  \foreach \val in {#9} {
    \pgfmathsetmacro{\oy}{\val*\yscale}
    \draw[fill=#7, draw=#8] (#1,\oy) circle (0.06cm);
  }
}

\boxplotfull{\xGPTa}{0.38}{1.0}{0.62}{0.1428571429}{1.0}{ncsuLightBlue}{ncsuDarkBlue}{}
\boxplotfull{\xGPTb}{1.0}{1.0}{1.0}{1.0}{1.0}{ncsuLightGreen}{ncsuDarkGreen}{0.8, 0.56, 0.25, 0.75}
\boxplotfull{\xRa}{0.29}{1.0}{1.0}{0.1666666667}{1.0}{ncsuLightBlue}{ncsuDarkBlue}{}
\boxplotfull{\xRb}{0.48}{1.0}{1.0}{0.125}{1.0}{ncsuLightGreen}{ncsuDarkGreen}{}
\boxplotfull{\xGemmaa}{0.22}{0.4}{0.27}{0}{0.6666666667}{ncsuLightBlue}{ncsuDarkBlue}{1.0}
\boxplotfull{\xGemmab}{0.29}{0.7}{0.4}{0}{1.0}{ncsuLightGreen}{ncsuDarkGreen}{}
\boxplotfull{\xP}{0.75}{1.0}{1.0}{0.40}{1.0}{ncsuLightPurple}{ncsuDarkPurple}{0.0, 0.0, 0.0, 0.0, 0.09, 0.14, 0.11, 0.0, 0.08, 0.0, 0.25, 0.0, 0.25, 0.0, 0.0, 0.25, 0.2, 0.2, 0.0, 0.0, 0.0, 0.0, 0.0, 0.0, 0.0, 0.0, 0.0, 0.0, 0.14, 0.33, 0.14, 0.0, 0.0, 0.0, 0.2, 0.2, 0.0, 0.0, 0.0, 0.2, 0.0, 0.0, 0.0, 0.0, 0.0, 0.33, 0.0, 0.0, 0.0, 0.0, 0.14, 0.0, 0.0, 0.0, 0.0, 0.0, 0.0, 0.0, 0.0, 0.14, 0.0, 0.0, 0.0, 0.0, 0.0, 0.14, 0.0, 0.25, 0.2, 0.0, 0.0, 0.0, 0.11, 0.0, 0.0, 0.11, 0.0, 0.0, 0.0, 0.0, 0.0, 0.14, 0.33, 0.0, 0.0, 0.0, 0.0, 0.2, 0.11, 0.11, 0.11, 0.2, 0.2, 0.14, 0.0, 0.0, 0.14, 0.14, 0.29, 0.29}

\node[below, xshift=1.5em,yshift=-0.5em] at (\xGPTa,0) {\small \textbf{GPT-4o}};
\node[below, xshift=1.3em,yshift=-0.5em] at (\xRa,0) {\small \textbf{Deepseek-R1}};
\node[below, xshift=1.3em,yshift=-0.5em] at (\xGemmaa,0) {\small \textbf{Gemma-3-27B}};
\node[below, yshift=-0.5em] at (\xP,0) {\small \textbf{Participants}};

\node[rotate=90] at (-1.4,2.5) {\large \textbf{Plan Feasibility}};
\begin{scope}[shift={(-0.5,5.8)}] 
  \draw[thin] (0.38,-0.3) rectangle (10.43,0.3);
  \draw[fill=ncsuLightBlue, draw=ncsuDarkBlue] (0.5,-0.1) rectangle +(0.7,0.3);
  \node[right=4pt] at (1,0.05) {\large Tomcat\_CP};

  \draw[fill=ncsuLightGreen, draw=ncsuDarkGreen] (3.6,-0.1) rectangle +(0.7,0.3);
  \node[right=4pt] at (4.1,0.05) {\large Tomcat\_Fs-CoT};

    \draw[fill=ncsuLightPurple, draw=ncsuDarkPurple] (7.35,-0.1) rectangle +(0.7,0.3);
  \node[right=4pt] at (7.85,0.015) {\large Participants};
\end{scope}

\end{tikzpicture}
\caption{Plan feasibility across LLMs and participants.}
\label{fig:pf-visual}
\end{subfigure}

\caption{Box plots of continuous metrics with the IQR and median collapsed into a single line, indicating near-optimal performance.}
\label{fig:result-boxplot}
\end{figure}

\begin{table}[htb]
    \centering
\rowcolors{2}{orange!10}{white}
\resizebox{\textwidth}{!}{
\begin{tabular}{@{}l l c c c@{}}
\toprule
\textbf{LLM} & \textbf{Metric} & \textbf{Fs-CoT vs. CP} & \textbf{Part. vs. CP} & \textbf{Part. vs. Fs-CoT} \\
\midrule
    & Intent Accuracy & \sig{g=.78} &\sigv{V=.13} & \nosig \\
\whitecell{} & Action Feasibility  & \sig{r=.35} & \sig{r=.25} & \nosig \\
GPT-4o & Action Optimality & \sig{r=.55} & \sig{r=.45} & \nosig \\
\whitecell{} & Plan Feasibility    & \sig{r=.55} & \sig{r=.34} & \nosig \\
\whitecell{} & Plan Optimality   & \sig{g=1.0} & \sigv{V=.18} & \nosig \\
\whitecell{}       & Instruction Acc.  & \nose{g=1.0} & -- & -- \\
\midrule
    & Intent Accuracy & \nose{g=1.0} &\nose{$p$=.053} & \nosig \\
\whitecell{} & Action Feasibility & \nosig  & \nosig & \nosig \\  
DeepSeek-R1 & Action Optimality & \nosig    & \nosig & \nosig \\
\whitecell{} & Plan Feasibility & \nosig  & \nosig & \nosig \\ 
\whitecell{} & Plan Optimality    & \nose{g=0.6} & \nosig & \nosig \\
\whitecell{}       & Instruction Acc.  & \nose{g=1.0} & -- & -- \\
\midrule
    & Intent Accuracy & \nos{g=1.0} & \sigv{V=.33} & \sigv{V=.22} \\
\whitecell{} & Action Feasibility  & \nosig  & \sig{r=.37} & \sig{r=.30} \\
Gemma-3-27B & Action Optimality & \nosig & \sig{r=.64} & \sig{r=.47} \\
\whitecell{} & Plan Feasibility & \nosig       & \sig{r=.61} & \sig{r=.41} \\
\whitecell{} & Plan Optimality  & \nos{g=1.0} & \sigv{V=.26} & \sigv{V=.16} \\
\whitecell{}       & Instruction Acc.  & \nose{g=0.6} & -- & -- \\
\bottomrule
\end{tabular}
}
    \caption{\textinred{Statistical outcomes across all metrics. Here, \ssig\ denotes $p < 0.05$, \nosig\ indicates no significant differences (or that the test could not be conducted), \snos\ marks cases where both $p$-value approaches significance ($0.05 < p < 0.1$) and the effect was large, and \snose\ denotes cases where only one of these conditions (near-significance or large effect) was met.}}
    \label{tab:stat-summary}
\end{table}

\subsection{Metrics Visualization}
In Figure~\ref{fig:result-barplot}, the standard error of the mean (SEM) for Tomcat with CP is generally higher than for both Tomcat with Fs-CoT and the participants across all LLMs, indicating greater variability and less reliable average performance. Notably, for the Gemma-3-27B Tomcat variants, the SEM for Fs-CoT for intent accuracy (Figure~\ref{fig:ia-visual}) and plan optimality (Figure~\ref{fig:po-visual}) is slightly higher than that of CP, suggesting that despite its improved accuracy, Fs-CoT exhibits more inconsistent outcomes on this LLM. For GPT-4o, the SEM for both variants is identical in plan optimality (Figure~\ref{fig:po-visual}), indicating that Fs-CoT achieves better central performance than CP, yet their variability remains comparable. In contrast, the consistently lower SEMs observed for the Fs-CoT variants of GPT-4o and DeepSeek-R1, as well as for the participants, reflect more stable performance, making their reported means more robust.

In Figure~\ref{fig:af-visual}, for action feasibility, the CP variants of GPT-4o and DeepSeek-R1 show a relatively narrow interquartile range (IQR) with higher medians, indicating low variability and consistently high performance. In contrast, the Gemma-3-27B CP variant exhibits a wider IQR and greater variability, reflecting inconsistent performance. The Fs-CoT variants of GPT-4o and DeepSeek-R1, along with the participants, achieve near-ceiling performance, as shown by their high medians and minimal variability, with only a few minor outliers. However, the Gemma-3-27B Fs-CoT variant shows a noticeably wider IQR and increased variability than the other Fs-CoT variants.

In Figure~\ref{fig:ao-visual}, for action optimality, the GPT-4o CP variant and both Gemma-3-27B Tomcat variants exhibit wider IQRs with lower medians, indicating greater variability and overall weaker performance. In contrast, the GPT-4o Fs-CoT variant, both DeepSeek-R1 Tomcat variants, and the participants display higher medians and fewer outliers. Notably, the IQRs for the GPT-4o Fs-CoT variant and the participants are tightly collapsed, suggesting that most values are concentrated around near-optimal performance. The DeepSeek-R1 Tomcat variants exhibit narrow IQRs with high medians, reflecting stable and high performance. 

In Figure~\ref{fig:pf-visual}, for plan feasibility, the GPT-4o CP variant exhibits a wider IQR and a lower median, indicating variability and weaker performance. In contrast, its Fs-CoT variant shows a higher median and a collapsed IQR, reflecting greater consistency and improvement. The DeepSeek-R1 CP variant shows a wider IQR than GPT-4o CP but a higher median, suggesting better overall performance despite variability. Its Fs-CoT variant, however, displays a tighter IQR and a high median, indicating stable and better performance. The Gemma-3-27B CP variant has a narrower IQR but a lower median than its Fs-CoT counterpart, suggesting that while the latter achieves better performance, it also exhibits higher variability. Participants show a somewhat wider IQR than GPT-4o Fs-CoT but maintain a high median with minimal outliers, indicating consistently strong performance.

\subsection{Answering the Research Questions}
We answer the following research questions by the summary statistics \textinred{of performance} in Table~\ref{tab:comparison_metrics} \textinred{and statistical significance tests in Table~\ref{tab:stat-summary}}, and analyzing the visualized metrics, with findings reported separately for each LLM.

\begin{description}
\item [RQ$_1$:] How do Tomcat's ToM capabilities on Instruction Inference compare to human performance?

\textbf{GPT-4o:} Tomcat with CP underperformed participants in key metrics such as plan optimality and intent accuracy, which are critical indicators of ToM reasoning. In contrast, Tomcat with Fs-CoT demonstrated ToM capabilities on par with the participants, as indicated by the statistical significance testing.

\textbf{DeepSeek-R1:} Tomcat with CP did not match participant performance in planning and intent recognition. Although the statistical tests revealed no significant differences, intent accuracy approached significance ($p = 0.053$), and participants showed gains over CP in planning metrics, suggesting weaker ToM reasoning for CP. For other metrics, however, CP performed comparably with participants. In contrast, Tomcat with Fs-CoT achieved substantial improvements over CP in both intent accuracy and planning, and performed on par with participants across all ToM-relevant metrics.   

\textbf{Gemma-3-27B:} Tomcat with CP underperformed participants across all metrics, particularly in intent recognition and planning. Despite gains in these metrics, the Fs-CoT variant also failed to reach participant-level performance. Statistical tests confirmed significant differences between participants and both variants, and although Fs-CoT improved over CP, the improvement remained insufficient. These findings suggest that Gemma-3-27B lacks the model capacity required for reliable ToM reasoning, even with structured exemplars.

\item [RQ$_2$:] \textinred{Does Tomcat with Fs-CoT outperform Tomcat with CP on Instruction Inference?}

\textbf{GPT-4o:} Fs-CoT demonstrations substantially enhanced Tomcat's ToM capabilities and performance on Instruction Inference. By leveraging these few-shot exemplars, Tomcat effectively interpreted \textinred{incomplete} and ambiguous instructions, inferred human intentions, and optimized actions and plans, achieving notable improvements over Tomcat with CP, as evidenced by its performance across the metrics and significance testing.

\textbf{DeepSeek-R1:} Fs-CoT demonstrations improved Tomcat's performance, particularly in intent recognition, planning, and instruction identification, where it achieved notable gains over CP. Although significance tests did not reveal statistical differences between Fs-CoT and CP, large effect sizes for intent and instruction accuracy, along with consistent improvements in planning metrics, suggest that Fs-CoT enabled more reliable interpretation of \textinred{unclear} instructions. Overall, Fs-CoT improved DeepSeek-R1 Tomcat's ToM capabilities, aligning its performance with that of the participants. Notably, the CP variant of DeepSeek-R1 was the strongest among all CP variants, particularly in the action metrics, where both variants performed similarly.

\textbf{Gemma-3-27B:} Fs-CoT demonstrations improved Tomcat's performance on Instruction Inference, achieving substantial gains over CP across all metrics. Although the statistical tests revealed no significant differences between the two variants, intent accuracy and plan optimality approached significance ($p = 0.063$), and plan feasibility yielded a moderate effect size, all of which support improved ToM performance with Fs-CoT. However, these gains were less notable than the other LLMs, largely because CP performed poorly. However, despite these strong improvements, Tomcat with Fs-CoT still did not achieve participant-level performance. This outcome likely reflects limitations in Gemma-3-27B's smaller model size, which may constrain its capacity for complex ToM reasoning even when guided by structured exemplars.
\end{description}

\subsection{Theoretical Assumptions of LLMs and Fs-CoT}
\label{sec:theoretical-assumptions}
\textinred{To interpret the performance variations observed across the LLMs, we outline theoretical assumptions grounded in their respective post-training optimization objectives and reasoning behaviors.}

\paragraph{DeepSeek-R1: Self-Directed Reasoning through GRPO} 
\textinred{DeepSeek-R1 is optimized via Reinforcement Learning (Group Relative Policy Optimization) \cite{Guo+25:DeepSeekR1} using outcome-based rewards. It is trained to converge on an internally derived reasoning trajectory. \citet{Guo+25:DeepSeekR1} find that introducing few-shot exemplars acts as \fsl{contextual noise} for reasoning-intensive tasks, forcing the model to reconcile its optimized internal path with an external template. This likely creates a \fqt{reasoning collision} that degrades performance, suggesting that zero-shot prompting may act as a practical performance ceiling for DeepSeek-R1. However, our empirical results offer a more nuanced perspective. Although we observed no statistical differences between the Fs-CoT and CP variants, the Fs-CoT variant showed moderate to substantial absolute gains across all metrics, except \fsl{action feasibility}. These findings suggest that the claim by \citet{Guo+25:DeepSeekR1}---\fqt{Few-shot prompting consistently degrades its performance}---may not hold universally in complex, task-oriented environments like instruction inference, as we observed no performance degradation using Fs-CoT.}  
\paragraph{GPT-4o: Structural Alignment via RLHF} 
\textinred{Conversely, GPT-4o exhibits a direct correlation between performance and prompt alignment. As a product of dense Supervised Finetuning (SFT) and Reinforcement Learning from Human Feedback (RLHF) \cite{Ouyang+22:TrainingLMsHF}, GPT-4o is strongly optimized for instruction following. It treats the few-shot context as a command structure or a formatting instruction, which aligns with its training to act as a mimicking assistant. The exemplars serve as guiding constraints that effectively reduce the search space for an optimal response \cite{Min+22:in-context-learning}. Consistent with this, we observed that GPT-4o achieved the highest scores across all metrics when provided with Fs-CoT prompting over its counterpart.}

\paragraph{Gemma-3-27B: Simulation via Knowledge Distillation} 
\textinred{Gemma's performance is likely driven by teacher-student mimicry. Using knowledge distillation from larger models (e.g., Gemini), its reward functions \cite{Rame+24:WARM} encourage behavior that approximates the reasoning patterns of the \fqt{Teacher} model. Unlike DeepSeek-R1's self-discovery, Gemma is optimized to replicate high-quality reasoning patterns provided during post-training. Consequently, Fs-CoT prompting succeeds by providing the teacher-like context Gemma was specifically trained to simulate. In this study, we observed the success of the Gemma-3-27B Fs-CoT variant over its CP variant, suggesting that explicit reasoning exemplars effectively activate these distilled capabilities.}

\textinred{Table~\ref{tab:theoretical_summary} breaks down the optimization objectives and impact of using Fs-CoT.}  

\begin{table}[htb]
\centering
\rowcolors{2}{white}{mint!15}
\resizebox{\textwidth}{!}{
\begin{tabular}{R{0.2\textwidth} R{0.25\textwidth} R{0.25\textwidth} R{0.3\textwidth}}
\toprule
\fbf{Model} & \fbf{Opt. Objective} & \fbf{Opt. Signal} & \fbf{Effect of Fs-CoT} \\ 
\midrule
\whitecell{}\fbf{DeepSeek-R1} & Self-Discovery & Outcome-based & Neutral (Reasoning Interference) \\
\whitecell{}\fbf{GPT-4o} & Alignment & Preference-based & Positive (Instruction Alignment) \\
\whitecell{}\fbf{Gemma-3-27B} & Simulation & Distillation-based & Positive (Pattern Activation) \\ \bottomrule
\end{tabular}
}
\caption{\textinred{Optimization (abbreviated as \fqt{Opt.}) objectives, signals and the effect of Fs-CoT across all evaluated LLMs.}}
\label{tab:theoretical_summary}
\end{table}

\section{Discussion}
\label{sec:discussion}
Our evaluation demonstrated that Fs-CoT, an instantiation of Tomcat that integrates structured reasoning exemplars, consistently outperformed its CP counterpart across multiple LLMs on most metrics, highlighting the benefits of exemplar-based ToM reasoning for instruction understanding. Across instruction identification, action, and planning metrics, Fs-CoT improved Tomcat's ability to model shared intentions and respond appropriately. The Fs-CoT variants of GPT-4o and DeepSeek-R1 achieved performance comparable to that of human participants, particularly in intent recognition and planning, where reasoning about the principal's future steps is crucial. 

Notably, the DeepSeek-R1 CP variant performed closely to both its Fs-CoT counterpart and the participants. \textinred{This is largely owing to the DeepSeek-R1 model being aligned to reason independently in its post-training phase}. For Gemma-3-27B, Fs-CoT yielded substantial improvements over CP, but its performance remained below human levels. This gap likely stems from its smaller model capacity, which may constrain its ability to generalize from exemplars and reason about complex, multistep tasks. These findings confirm the value of embedding structured ToM exemplars in collaborative agents and highlight the importance of model capacity in supporting Instruction Inference.

\subsection{Threats to Validity}
Despite the promising results, some factors may limit the generalizability of our findings. First, a small number of participants appeared to misunderstand the task requirements. Although such a variation is expected in user studies and reflects real-world variance, it may have introduced noise in our human baseline. Second, only two annotators evaluated the LLM completions. Although they followed a strict rubric and the criteria was objective, involving additional annotators could have reduced potential bias and increased confidence in the evaluation results.

\subsection{Directions in Addressing Limitations}
\label{sec:discuss-limitations}
We identify some limitations that open avenues for further investigation.
  \paragraph{Participant sampling} The study involved 52 participants, divided into two groups to balance cognitive load. Expanding to more diverse populations and ensuring exposure to all instructions would improve reliability.
  \paragraph{Model diversity} We tested only three LLMs. Incorporating a broader range of models, would yield more comprehensive insights into ToM generalization.
  \paragraph{\textinred{Domain diversity}} \textinred{Our experiment is limited to one domain of human-agent collaboration. Future work can benefit from incorporating diverse domains of human-agent collaboration.}
  \paragraph{Alternative techniques} Beyond exemplars, other ToM enhancement strategies such as reward-based finetuning, chain-of-reasoning distillation, or belief-tracking modules, remain unexplored in our setup.
  \paragraph{Environmental variation} Generalization could be tested by designing environments with alternative layouts, object types, or collaborative constraints.


\subsection{Directions in Addressing Opportunities}
\label{sec:discuss-opportunities}
Our results offer a promising outlook on human-agent collaboration. Tomcat with Fs-CoT demonstrates the potential of AI agents to serve as capable collaborators by accurately interpreting \textinred{incomplete} or ambiguous instructions, inferring human intent, and generating feasible and optimal actions and plans, often on par with human participants. These findings motivate future work on extending Instruction Inference by incorporating pragmatics to address a broader spectrum of instruction types, including nonverbal cues, humor, sarcasm, irrelevance, invalidity, and incompleteness. Further, to ensure robust generalization across LLMs of varying capacities, especially lower capacity models like Gemma-3-27B, future research should explore model-specific adaptations and scaling strategies. 

\textinred{The development of such capabilities aligns with the projected evolution of LLMs from transactional chatbots to relational cognitive agents. As post-training scaling laws begin to emphasize multistep reasoning, agentic planning, and metacognition, future architectures are expected to incorporate explicit recursive reasoning modules \cite{Qi+23:Recursive-Socratic-LLM, Qu+24:Recursive-SelfImprove-LLM, Zhang+25:Recursive-LLM} for self-reflection and reinforcement learning strategies for finetuning ToM reasoning \cite{Plaat+25:MSR-Survey}. Such advancements will likely enable even smaller-scale models to move beyond literal pattern matching toward functional ToM, where they can maintain persistent models of a user's mental state across extended collaborative sessions.} \textinred{Additionally, our findings regarding DeepSeek-R1 invite a re-evaluation of the Fs-CoT strategy. Contrary to the developers' claims that few-shot prompting degrades performance, we observed no such degradation, suggesting that DeepSeek-R1's reasoning capabilities may be more robust to contextual examples than previously reported.} Finally, we aim to evaluate Tomcat in real-world collaborative domains featuring richer dialogue, multiagent coordination, and dynamic environments.    

\subsection{Reproducibility}
The code, dataset, task description, and results for our experiments are available at: \url{https://github.com/fardinsaad/Tomcat-LLM}.

\section*{Acknowledgments}
This research was partially supported by the National Science Foundation (grant IIS-2116751).

\bibliographystyle{elsarticle-num-names}
\bibliography{Fardin,Munindar}

\newpage
\appendix
\section{Prompts}
\subsection{Common Ground Component}
\label{sec:CGC}
\begin{figure}[htb]
    \begin{promptbox}[colback=green!10]{robot}{General Chain-of-Thought and Background}
    You assist a human in a cooperative planning domain called Doors, Keys, and Gems, set in a gridworld. The human attempts to retrieve a specific gem, and you assist the human. You help by unlocking doors using the correct keys or passing the keys to the human. Think of it as a team game where a human gives callouts to their teammate (you), expecting collaboration to achieve their goal. The challenge for you is correctly understanding human instructions, which may require some inferencing due to a lack of specificity in the instructions. \\\\
   \textbf{\textit{Background:}} This collaborative game is played in a grid containing the following objects: keys, doors, gems, walls, empty spaces, a human, and you. The colors of keys and doors can be red, blue, or yellow. \\\\
   \textbf{Figure 1: Grid Configuration}
   \begin{MyVerbatim}
[[`.' `.' `.' `y' `.' `.' `.' `b' `W' `W' `W' `W']
 [`r' `W' `W' `r' `W' `W' `.' `r' `W' `W' `W' `g']
 [`W' `W' `W' `W' `W' `W' `m' `W' `W' `W' `W' `R']
 [`W' `W' `W' `W' `W' `W' `.' `W' `W' `W' `W' `.']
 [`g' `.' `.' `.' `B' `.' `.' `R' `.' `.' `.' `.']
 [`W' `W' `W' `W' `W' `W' `.' `W' `W' `W' `W' `W']
 [`W' `W' `W' `W' `W' `W' `.' `W' `W' `W' `W' `W']
 [`.' `.' `.' `Y' `.' `.' `.' `W' `W' `W' `W' `W']
 [`B' `W' `W' `W' `W' `W' `.' `W' `W' `W' `W' `W']
 [`g' `W' `W' `W' `W' `W' `h' `.' `.' `.' `.' `g']]

   \end{MyVerbatim}
\end{promptbox}
    \caption{Common Ground component part-1.}
    \label{fig:c1-1}
\end{figure}

\newpage

\begin{figure}[htb]
    \begin{promptbox}[colback=green!10]{robot}{Objectives and Rules}
   \textbf{\textit{Human's objective:}} The human's goal is to retrieve a specific gem by providing only a single instruction. However, the human's instruction could be ambiguous or lack specificity, adding a challenge for you to correctly infer the gem desired by the human. For instance, consider the configuration in Figure 1. While some gems, like the gem at (9, 11), may be easily accessible (i.e., don't require a key), the human may instead be aiming for a more challenging gem, such as the red gem placed behind two doors, and can provide instruction as "Get the red key." Here, the human is directing you to collect a red key to unlock the red door and get the gem. But the gem is blocked by two red doors. Thus, you must collect and pass two red keys to the human. You need to infer this even when not explicitly stated based on the context (i.e., the movement of the human). \\\\
   \textbf{\textit{Your objective:}} Your primary goal is to assist the human in retrieving the desired gem by fetching the necessary keys or unlocking doors while prioritizing minimal movement. Upon receiving the instruction, you infer which gem the human intends to collect and take only the essential actions—either collecting and passing keys or unlocking doors—to enable the human to access the gem as efficiently and quickly as possible. As a result, your movements are optimized to ensure that the number of steps taken is minimal. \\\\
   \textbf{\textit{Rules of the Game:}} \\
   - For each grid configuration (problem), the human provides a single instruction called a callout. \\
- The instructions (callouts) can be of two types: (1) either directing you to collect a key (or keys) and passing them to the human or (2) instructing you to unlock a door (or doors). \\
- Only the key of the same color as the door can unlock the door. However, once a key is used to unlock a door, it cannot be reused for any other door of the same color. For instance, one red key can only unlock one red door. \\
- If you collect one or more keys, they are immediately passed to the human, depending on the grid configuration. However, if you are blocked by a door (or doors) and cannot pass the key directly, you will either collect the necessary key(s) to unlock the door or instruct the human to retrieve the key if it is beyond your reach. This ensures you can overcome obstacles and continue assisting the human efficiently. \\
- If the callout instructs to unlock a door or doors, you must first collect the corresponding key or keys before proceeding to unlock the door (or doors). In these instances, you do not pass the key to the human. \\
- Gems are randomly placed within the grid. \\
- You prioritize efficient, optimal, and obstacle-free routes to achieve the human's objective as quickly and effectively as possible. When instructed to collect key(s) or unlock the door(s), you don't gather all the keys or unlock all the doors on the grid. Instead, you perform only the actions necessary 
to assist the human in retrieving the gem with minimal steps. \\
- The objects, such as gems, keys, and doors, have no associated cost (i.e., they are all the same). Your primary goal is to acquire the necessary keys or unlock the appropriate doors, ensuring the human can obtain the desired gem most efficiently, i.e., requiring minimal movements to execute the instruction. \\
- The human's instruction may not always be explicit, requiring you to infer the intended action based on the context and the grid configuration. The instructions may be ambiguous or lacking specificity.
 
\end{promptbox}
    \caption{Common Ground component part-2.}
    \label{fig:c1-2}
\end{figure}

\FloatBarrier

\subsection{Response Generation Component}
\label{sec:RGC}
\begin{figure}[htb]
    \begin{promptbox}[colback=red!20]{robot}{Response Generation: Grid Structure}
   The Current Grid Layout for the Keys, Doors, and, Gems:
   \begin{MyVerbatim}
[[`.' `.' `.' `y' `.' `.' `.' `b' `W' `W' `W' `W']
 [`r' `W' `W' `r' `W' `W' `.' `r' `W' `W' `W' `g']
 [`W' `W' `W' `W' `W' `W' `m' `W' `W' `W' `W' `R']
 [`W' `W' `W' `W' `W' `W' `.' `W' `W' `W' `W' `.']
 [`g' `.' `.' `.' `B' `.' `.' `R' `.' `.' `.' `.']
 [`W' `W' `W' `W' `W' `W' `.' `W' `W' `W' `W' `W']
 [`W' `W' `W' `W' `W' `W' `.' `W' `W' `W' `W' `W']
 [`.' `.' `.' `Y' `.' `.' `.' `W' `W' `W' `W' `W']
 [`B' `W' `W' `W' `W' `W' `.' `W' `W' `W' `W' `W']
 [`g' `W' `W' `W' `W' `W' `h' `.' `.' `.' `.' `g']]
   \end{MyVerbatim}
   Below are the locations of all objects on the grid, displayed as (row, column) coordinates, along with their total count: \\
My position (Labeled as 'm'): (2, 6) \\
Human (Labeled as 'h'): (9, 6) \\
Red keys (Labeled as 'r'): (1, 0), (1, 3), (1, 7) \texttt{-->} Total Red keys: 3 \\
Yellow key (Labeled as 'y'): (0, 3) \texttt{-->} Total Yellow key: 1 \\
Blue key (Labeled as 'b'): (0, 7) \texttt{-->} Total Blue key: 1\\
Red doors (Labeled as 'R'): (2, 11), (4, 7) \texttt{-->} Total Red doors: 2\\
Yellow door (Labeled as 'Y'): (7, 3) \texttt{-->} Total Yellow door: 1\\
Blue doors (Labeled as 'B'): (4, 4), (8, 0) \texttt{-->} Total Blue doors: 2\\
Gems (Labeled as 'g'): (1, 11), (4, 0), (9, 0), (9, 11) \texttt{-->} Total Gems: 4\\
Walls (Labeled as 'W'): (0, 8), (0, 9), (0, 10), (0, 11), (1, 1), (1, 2), (1, 4), (1, 5), (1, 8), (1, 9), (1, 10), (2, 0), (2, 1), (2, 2), (2, 3), (2, 4), (2, 5), (2, 7), (2, 8), (2, 9), (2, 10), (3, 0), (3, 1), (3, 2), (3, 3), (3, 4), (3, 5), (3, 7), (3, 8), (3, 9), (3, 10), (5, 0), (5, 1), (5, 2), (5, 3), (5, 4), (5, 5), (5, 7), (5, 8), (5, 9), (5, 10), (5, 11), (6, 0), (6, 1), (6, 2), (6, 3), (6, 4), (6, 5), (6, 7), (6, 8), (6, 9), (6, 10), (6, 11), (7, 7), (7, 8), (7, 9), (7, 10), (7, 11), (8, 1), (8, 2), (8, 3), (8, 4), (8, 5), (8, 7), (8, 8), (8, 9), (8, 10), (8, 11), (9, 1), (9, 2), (9, 3), (9, 4), (9, 5) \texttt{-->} Total Walls: 73 \\
Empty spaces (Labeled as '.'): (0, 0), (0, 1), (0, 2), (0, 4), (0, 5), (0, 6), (1, 6), (3, 6), (3, 11), (4, 1), (4, 2), (4, 3), (4, 5), (4, 6), (4, 8), (4, 9), (4, 10), (4, 11), (5, 6), (6, 6), (7, 0), (7, 1), (7, 2), (7, 4), (7, 5), (7, 6), (8, 6), (9, 7), (9, 8), (9, 9), (9, 10) \texttt{-->} Total Empty spaces: 31 
\end{promptbox}
    \caption{Response Generation component part-1.}
    \label{fig:c2-1}
\end{figure}

\newpage

\begin{figure}[H]
    \begin{promptbox}[colback=red!20]{robot}{Response Generation: Type, Response, Actions}
   Based on your understanding of the game rules, current grid layout, human actions, instruction, and the labeled examples provided below, generate the instruction type (clear or unclear), response, and optimal actions for the following human actions and instruction, delimited by triple backticks:\\
\texttt{```}\\
\texttt{\textbf{Human Action:} The human moves to the right from  their current position at (5,0) until they reach at (5,4) and gives the instruction.\\
\textbf{Instruction:} Can you pass me the blue key?\\
\textbf{Type:} \textless Identify the instruction type and provide chain-of-thought reasoning for the type based on the current grid configuration\textgreater\\
\textbf{Response:} \textless Generate chain-of-thought reasoning for 
the response based on the current grid configuration\textgreater\\
\textbf{Actions:} \textless Provide optimal actions from the 
response based on the instruction and the current 
grid configuration\textgreater\\}
\texttt{```} 
\end{promptbox}
    \caption{Response Generation component part-2.}
    \label{fig:c2-2}
\end{figure}

\FloatBarrier
\newpage
\subsection{Demonstrations: CP}
\label{sec:DCP}
\begin{figure}[htb]
    \begin{promptbox}[colback=ncsuLightBlue!60]{robot}{Demonstration Exemplars: CP Problem-1 (Initial Grid)}
    Use the following problems and examples, delimited by triple quotes, to understand how to generate the appropriate type, response, and actions for each instruction and human actions.\\
\texttt{'''}\\
   ---\textbf{Problem (1)} demonstrating initial, observed, completed grids with figures and the example. \\
\textbf{Figure 2:} Initial Grid Configuration
   \begin{MyVerbatim}
[[`r' `.' `.' `.' `m' `W' `W' `g']
 [`y' `.' `W' `W' `.' `W' `W' `.']
 [`W' `W' `W' `W' `.' `W' `W' `.']
 [`.' `R' `.' `.' `.' `.' `h' `.']
 [`.' `W' `.' `W' `W' `W' `W' `.']
 [`.' `W' `.' `W' `W' `W' `W' `Y']
 [`Y' `W' `.' `W' `W' `W' `W' `.']
 [`g' `W' `g' `W' `W' `W' `W' `g']]
   \end{MyVerbatim}
   Object Positions for Figure 2: \\
My position (Labeled as 'm'): (0, 4)\\
Human (Labeled as 'h'): (3, 6)\\
Red key (Labeled as 'r'): (0, 0) \texttt{-->} Total Red key: 1\\
Yellow key (Labeled as 'y'): (1, 0) \texttt{-->} Total Yellow key: 1\\
Red door (Labeled as 'R'): (3, 1) \texttt{-->} Total Red door: 1\\
Yellow doors (Labeled as 'Y'): (5, 7), (6, 0) \texttt{-->} Total Yellow doors: 2\\
Gems (Labeled as 'g'): (0, 7), (7, 0), (7, 2), (7, 7) \texttt{-->} Total Gems: 4\\
Walls (Labeled as 'W'): (0, 5), (0, 6), (1, 2), (1, 3), (1, 5), (1, 6), (2, 0), (2, 1), (2, 2), (2, 3), (2, 5), (2, 6), (4, 1), (4, 3), (4, 4), (4, 5), (4, 6), (5, 1), (5, 3), (5, 4), (5, 5), (5, 6), (6, 1), (6, 3), (6, 4), (6, 5), (6, 6), (7, 1), (7, 3), (7, 4), (7, 5), (7, 6) \texttt{-->} Total Walls: 32\\
Empty spaces (Labeled as '.'): (0, 1), (0, 2), (0, 3), (1, 1), (1, 4), (1, 7), (2, 4), (2, 7), (3, 0), (3, 2), (3, 3), (3, 4), (3, 5), (3, 7), (4, 0), (4, 2), (4, 7), (5, 0), (5, 2), (6, 2), (6, 7) \texttt{-->} Total Empty spaces: 21
 \\\\
   \textbf{--Initial Grid:} This grid in Figure 2 displays the initial positions of you, human, keys (red, yellow), doors (red, yellow), and gems.
\end{promptbox}
    \caption{Demonstration exemplar component for Tomcat with CP part-1}
    \label{fig:c3-1}
\end{figure}

\newpage

\begin{figure}[h]
    \begin{promptbox}[colback=ncsuLightBlue!60]{robot}{Demonstration Exemplars: CP Problem-1 (Observed Grid)}
   \textbf{Figure 3:} Observed Grid Configuration
   \begin{MyVerbatim}
[[`r' `.' `.' `.' `m' `W' `W' `g']
 [`y' `.' `W' `W' `.' `W' `W' `.']
 [`W' `W' `W' `W' `.' `W' `W' `.']
 [`.' `R' `h' `.' `.' `.' `.' `.']
 [`.' `W' `.' `W' `W' `W' `W' `.']
 [`.' `W' `.' `W' `W' `W' `W' `Y']
 [`Y' `W' `.' `W' `W' `W' `W' `.']
 [`g' `W' `g' `W' `W' `W' `W' `g']]
   \end{MyVerbatim}
   Object Positions for Figure 3: \\
My position (Labeled as 'm'): (0, 4) \\
Human (Labeled as 'h'): (3, 2)\\
Red key (Labeled as 'r'): (0, 0) \texttt{-->} Total Red key: 1\\
Yellow key (Labeled as 'y'): (1, 0) \texttt{-->} Total Yellow key: 1\\
Red door (Labeled as 'R'): (3, 1) \texttt{-->} Total Red door: 1\\
Yellow doors (Labeled as 'Y'): (5, 7), (6, 0) \texttt{-->} Total Yellow doors: 2\\
Gems (Labeled as 'g'): (0, 7), (7, 0), (7, 2), (7, 7) \texttt{-->} Total Gems: 4\\
Walls (Labeled as 'W'): (0, 5), (0, 6), (1, 2), (1, 3), (1, 5), (1, 6), (2, 0), (2, 1), (2, 2), (2, 3), (2, 5), (2, 6), (4, 1), (4, 3), (4, 4), (4, 5), (4, 6), (5, 1), (5, 3), (5, 4), (5, 5), (5, 6), (6, 1), (6, 3), (6, 4), (6, 5), (6, 6), (7, 1), (7, 3), (7, 4), (7, 5), (7, 6) \texttt{-->} Total Walls: 32\\
Empty spaces (Labeled as '.'): (0, 1), (0, 2), (0, 3), (1, 1), (1, 4), (1, 7), (2, 4), (2, 7), (3, 0), (3, 3), (3, 4), (3, 5), (3, 6), (3, 7), (4, 0), (4, 2), (4, 7), (5, 0), (5, 2), (6, 2), (6, 7) \texttt{-->} Total Empty spaces: 21 \\\\
   \textbf{--Observed Grid:} The human moves left toward the red door in Figure 3 from position (3,6) to (3,2) and gives the instruction, ``Can you pass me the red key?'' From this, you infer the human's intentions based on their movement and location--specifically, which gem they are aiming to collect.
\end{promptbox}
    \caption{Demonstration exemplar component for Tomcat with CP part-2}
    \label{fig:c3-2}
\end{figure}

\newpage

\begin{figure}[ht]
    \begin{promptbox}[colback=ncsuLightBlue!60]{robot}{Demonstration Exemplars: CP Problem-1 (Completed Grid)}
    \small
   \textbf{Figure 4:} Completed Grid Configuration

   \begin{MyVerbatim}
[[`r' `.' `.' `.' `m' `W' `W' `g']
 [`y' `.' `W' `W' `.' `W' `W' `.']
 [`W' `W' `W' `W' `.' `W' `W' `.']
 [`.' `.' `.' `m' `.' `.' `.' `.']
 [`.' `W' `.' `W' `W' `W' `W' `.']
 [`.' `W' `.' `W' `W' `W' `W' `Y']
 [`.' `W' `.' `W' `W' `W' `W' `.']
 [`h' `W' `g' `W' `W' `W' `W' `g']]
   \end{MyVerbatim}
   Object Positions for Figure 4:\\
My position (Labeled as 'm'): (3, 3)\\
Human (Labeled as 'h'): (7, 0)\\
Yellow door (Labeled as 'Y'): (5, 7) \texttt{-->} Total Yellow door: 1\\
Gems (Labeled as 'g'): (0, 7), (7, 2), (7, 7) \texttt{-->} Total Gems: 3\\
Walls (Labeled as 'W'): (0, 5), (0, 6), (1, 2), (1, 3), (1, 5), (1, 6), (2, 0), (2, 1), (2, 2), (2, 3), (2, 5), (2, 6), (4, 1), (4, 3), (4, 4), (4, 5), (4, 6), (5, 1), (5, 3), (5, 4), (5, 5), (5, 6), (6, 1), (6, 3), (6, 4), (6, 5), (6, 6), (7, 1), (7, 3), (7, 4), (7, 5), (7, 6) \texttt{-->} Total Walls: 32\\
Empty spaces (Labeled as '.'): (0, 0), (0, 1), (0, 2), (0, 3), (0, 4), (1, 0), (1, 1), (1, 4), (1, 7), (2, 4), (2, 7), (3, 0), (3, 1), (3, 2), (3, 4), (3, 5), (3, 6), (3, 7), (4, 0), (4, 2), (4, 7), (5, 0), (5, 2), (6, 0), (6, 2), (6, 7) \texttt{-->} Total Empty spaces: 26
 \\\\
   \textbf{--Completed Grid:} You infer that the human intends to collect the gem beyond the red door. However, you notice that after the red door, there is also a yellow door blocking the path to the gem. In response, in Figure 4, you collect both the red and yellow keys and pass them to the human. The human then uses the corresponding keys to unlock both doors and successfully retrieves the gem.
\end{promptbox}
    \caption{Demonstration exemplar component for Tomcat with CP part-3}
    \label{fig:c3-3}
\end{figure}
\newpage
\begin{figure}[ht]
    \begin{promptbox}[colback=ncsuLightBlue!60]{robot}{Demonstration Exemplars: CP Problem-1 (Example)}
   \textbf{Example from this grid configuration.} \\
\textbf{Human Action:} The human moves left from their current position at (3,6) to (3,2), which is adjacent to the red door at (3,1). Upon arriving at (3,2), they provide the instruction. \\
\textbf{{Instruction:}} Can you pass me the red key? \\
\textbf{Type:} Unclear. Based on the instruction and the grid, I infer that the human wants to collect the gem at (7,0), beyond a red door. However, before reaching the gem, a yellow door is blocking the path. Therefore, I must collect both the red and yellow keys, and since the instruction doesn't mention picking up the yellow key, this instruction lacks specificity and is thus unclear. \\
\textbf{Response:} There are two yellow doors, one red door, one red key, and one yellow key on the grid. The instruction asks to collect a red key, so the desired gem the human wants to collect is placed beyond the red door at (3,1). However, a yellow door at (6,0) blocks the path to the gem at (7,0). Thus, I will collect both the red and yellow keys. To minimize movement, I will first pick up the red key at (0,0) followed by the yellow key at (1,0), as this strategy reduces the number of steps from my current position at (0,4). I will then pass them to the human at (3,2) to unlock the doors and collect the gem. \\
\textbf{Actions:} \\
1) Collect: red\_key at (0,0).\\
2) Collect: yellow\_key at (1,0). \\
3) Pass: red\_key and yellow\_key to the human at (3,2). \\
4) Unlock: human unlocks the Red\_door at (3,1) and the Yellow\_door at (6,0).\\
5) Retrieve: human retrieves gem at (7,0).

\end{promptbox}
    \caption{Demonstration exemplar component for Tomcat with CP part-4}
    \label{fig:c3-4}
\end{figure}

\newpage
\begin{figure}[H]
    \begin{promptbox}[colback=ncsuLightBlue!60]{robot}{Demonstration Exemplars: CP Problem-2 (Initial Grid)}
   ---\textbf{Problem (2)} demonstrating initial, observed, completed grids with figures and the example. \\
\textbf{Figure 5:} Initial Grid Configuration

   \begin{MyVerbatim}
[[`W' `W' `b' `W' `W' `W' `r' `W' `W']
 [`W' `r' `.' `r' `W' `b' `.' `b' `W']
 [`W' `W' `.' `W' `W' `W' `.' `W' `W']
 [`W' `W' `.' `.' `m' `.' `.' `W' `W']
 [`W' `W' `W' `W' `.' `W' `W' `W' `g']
 [`h' `.' `.' `.' `.' `.' `B' `B' `.']
 [`W' `W' `W' `W' `.' `W' `W' `W' `g']
 [`W' `W' `W' `W' `.' `W' `W' `W' `W']
 [`W' `W' `W' `W' `R' `W' `W' `W' `W']
 [`W' `W' `W' `W' `R' `W' `W' `W' `W']
 [`g' `.' `.' `.' `.' `.' `.' `.' `g']]
   \end{MyVerbatim}
   Object Positions for Figure 5: \\
My position (Labeled as 'm'): (3, 4)\\
Human (Labeled as 'h'): (5, 0)\\
Red keys (Labeled as 'r'): (0, 6), (1, 1), (1, 3) \texttt{-->} Total Red keys: 3\\
Blue keys (Labeled as 'b'): (0, 2), (1, 5), (1, 7) \texttt{-->} Total Blue keys: 3\\
Red doors (Labeled as 'R'): (8, 4), (9, 4) \texttt{-->} Total Red doors: 2\\
Blue doors (Labeled as 'B'): (5, 6), (5, 7) \texttt{-->} Total Blue doors: 2\\
Gems (Labeled as 'g'): (4, 8), (6, 8), (10, 0), (10, 8) \texttt{-->} Total Gems: 4\\
Walls (Labeled as 'W'): (0, 0), (0, 1), (0, 3), (0, 4), (0, 5), (0, 7), (0, 8), (1, 0), (1, 4), (1, 8), (2, 0), (2, 1), (2, 3), (2, 4), (2, 5), (2, 7), (2, 8), (3, 0), (3, 1), (3, 7), (3, 8), (4, 0), (4, 1), (4, 2), (4, 3), (4, 5), (4, 6), (4, 7), (6, 0), (6, 1), (6, 2), (6, 3), (6, 5), (6, 6), (6, 7), (7, 0), (7, 1), (7, 2), (7, 3), (7, 5), (7, 6), (7, 7), (7, 8), (8, 0), (8, 1), (8, 2), (8, 3), (8, 5), (8, 6), (8, 7), (8, 8), (9, 0), (9, 1), (9, 2), (9, 3), (9, 5), (9, 6), (9, 7), (9, 8) \texttt{-->} Total Walls: 59\\
Empty spaces (Labeled as '.'): (1, 2), (1, 6), (2, 2), (2, 6), (3, 2), (3, 3), (3, 5), (3, 6), (4, 4), (5, 1), (5, 2), (5, 3), (5, 4), (5, 5), (5, 8), (6, 4), (7, 4), (10, 1), (10, 2), (10, 3), (10, 4), (10, 5), (10, 6), (10, 7) \texttt{-->} Total Empty spaces: 24
 \\\\
   \textbf{--Initial Grid:} This grid in Figure 5 displays the initial positions of you, human, keys (red, blue), doors (red, blue), and gems.
\end{promptbox}
    \caption{Demonstration Exemplars component for Tomcat with CP part-4}
    \label{fig:c3-11}
\end{figure}

\newpage

\begin{figure}[H]
    \begin{promptbox}[colback=ncsuLightBlue!60]{robot}{Demonstration Exemplars: CP Problem-2 (Observed Grid)}
   \textbf{Figure 6:} Observed Grid Configuration

   \begin{MyVerbatim}
[[`W' `W' `b' `W' `W' `W' `r' `W' `W']
 [`W' `r' `.' `r' `W' `b' `.' `b' `W']
 [`W' `W' `.' `W' `W' `W' `.' `W' `W']
 [`W' `W' `.' `.' `m' `.' `.' `W' `W']
 [`W' `W' `W' `W' `.' `W' `W' `W' `g']
 [`.' `.' `.' `.' `h' `.' `B' `B' `.']
 [`W' `W' `W' `W' `.' `W' `W' `W' `g']
 [`W' `W' `W' `W' `.' `W' `W' `W' `W']
 [`W' `W' `W' `W' `R' `W' `W' `W' `W']
 [`W' `W' `W' `W' `R' `W' `W' `W' `W']
 [`g' `.' `.' `.' `.' `.' `.' `.' `g']]
   \end{MyVerbatim}
   Object Positions for Figure 6:\\
My position (Labeled as 'm'): (3, 4)\\
Human (Labeled as 'h'): (5, 4)\\
Red keys (Labeled as 'r'): (0, 6), (1, 1), (1, 3) \texttt{-->} Total Red keys: 3\\
Blue keys (Labeled as 'b'): (0, 2), (1, 5), (1, 7) \texttt{-->} Total Blue keys: 3\\
Red doors (Labeled as 'R'): (8, 4), (9, 4) \texttt{-->} Total Red doors: 2\\
Blue doors (Labeled as 'B'): (5, 6), (5, 7) \texttt{-->} Total Blue doors: 2\\
Gems (Labeled as 'g'): (4, 8), (6, 8), (10, 0), (10, 8) \texttt{-->} Total Gems: 4\\
Walls (Labeled as 'W'): (0, 0), (0, 1), (0, 3), (0, 4), (0, 5), (0, 7), (0, 8), (1, 0), (1, 4), (1, 8), (2, 0), (2, 1), (2, 3), (2, 4), (2, 5), (2, 7), (2, 8), (3, 0), (3, 1), (3, 7), (3, 8), (4, 0), (4, 1), (4, 2), (4, 3), (4, 5), (4, 6), (4, 7), (6, 0), (6, 1), (6, 2), (6, 3), (6, 5), (6, 6), (6, 7), (7, 0), (7, 1), (7, 2), (7, 3), (7, 5), (7, 6), (7, 7), (7, 8), (8, 0), (8, 1), (8, 2), (8, 3), (8, 5), (8, 6), (8, 7), (8, 8), (9, 0), (9, 1), (9, 2), (9, 3), (9, 5), (9, 6), (9, 7), (9, 8) \texttt{-->} Total Walls: 59\\
Empty spaces (Labeled as '.'): (1, 2), (1, 6), (2, 2), (2, 6), (3, 2), (3, 3), (3, 5), (3, 6), (4, 4), (5, 0), (5, 1), (5, 2), (5, 3), (5, 5), (5, 8), (6, 4), (7, 4), (10, 1), (10, 2), (10, 3), (10, 4), (10, 5), (10, 6), (10, 7) \texttt{-->} Total Empty spaces: 24
 \\\\
   \textbf{--Observed Grid:} The human moves to the right from their current position in Figure 6 from (5,0) to (5,4), and provides an instruction, "Can you pass me the red keys?" From this, you infer the human's intentions based on their movement and location—specifically, which gem they are aiming to collect. 
\end{promptbox}
    \caption{Demonstration exemplar component for Tomcat with CP part-5}
    \label{fig:c3-22}
\end{figure}

\newpage
\begin{figure}[H]
    \begin{promptbox}[colback=ncsuLightBlue!60]{robot}{Demonstration Exemplars: CP Problem-2 (Completed Grid)}
   \textbf{Figure 7:} Completed Grid Configuration
   \begin{MyVerbatim}
[[`W' `W' `b' `W' `W' `W' `r' `W' `W']
 [`W' `.' `.' `,' `W' `b' `.' `b' `W']
 [`W' `W' `.' `W' `W' `W' `.' `W' `W']
 [`W' `W' `.' `.' `.' `.' `.' `W' `W']
 [`W' `W' `W' `W' `m' `W' `W' `W' `g']
 [`.' `.' `.' `.' `,' `.' `B' `B' `.']
 [`W' `W' `W' `W' `.' `W' `W' `W' `g']
 [`W' `W' `W' `W' `.' `W' `W' `W' `W']
 [`W' `W' `W' `W' `.' `W' `W' `W' `W']
 [`W' `W' `W' `W' `.' `W' `W' `W' `W']
 [`h' `.' `.' `.' `.' `.' `.' `.' `g']]
   \end{MyVerbatim}
   Object Positions for Figure 7:\\
My position (Labeled as 'm'): (4, 4)\\
Human (Labeled as 'h'): (10, 0)\\
Red key (Labeled as 'r'): (0, 6) \texttt{-->} Total Red key: 1\\
Blue keys (Labeled as 'b'): (0, 2), (1, 5), (1, 7) \texttt{-->} Total Blue keys: 3\\
Red doors (Labeled as 'R'): (8, 4), (9, 4) \texttt{-->} Total Red doors: 2\\
Blue doors (Labeled as 'B'): (5, 6), (5, 7) \texttt{-->} Total Blue doors: 2\\
Gems (Labeled as 'g'): (4, 8), (6, 8), (10, 8) \texttt{-->} Total Gems: 3\\
Walls (Labeled as 'W'): (0, 0), (0, 1), (0, 3), (0, 4), (0, 5), (0, 7), (0, 8), (1, 0), (1, 4), (1, 8), (2, 0), (2, 1), (2, 3), (2, 4), (2, 5), (2, 7), (2, 8), (3, 0), (3, 1), (3, 7), (3, 8), (4, 0), (4, 1), (4, 2), (4, 3), (4, 5), (4, 6), (4, 7), (6, 0), (6, 1), (6, 2), (6, 3), (6, 5), (6, 6), (6, 7), (7, 0), (7, 1), (7, 2), (7, 3), (7, 5), (7, 6), (7, 7), (7, 8), (8, 0), (8, 1), (8, 2), (8, 3), (8, 5), (8, 6), (8, 7), (8, 8), (9, 0), (9, 1), (9, 2), (9, 3), (9, 5), (9, 6), (9, 7), (9, 8) \texttt{-->} Total Walls: 59\\
Empty spaces (Labeled as '.'): (1, 1), (1, 2), (1, 3), (1, 6), (2, 2), (2, 6), (3, 2), (3, 3), (3, 4), (3, 5), (3, 6), (5, 0), (5, 1), (5, 2), (5, 3), (5, 4), (5, 5), (5, 8), (6, 4), (7, 4), (10, 1), (10, 2), (10, 3), (10, 4), (10, 5), (10, 6), (10, 7) \texttt{-->} Total Empty spaces: 27
 \\\\
   \textbf{--Completed Grid:} You infer that the human intends to retrieve the gem located beyond the two red doors on the grid. Observing that there are three red keys available, you strategically choose to collect the two keys positioned on the upper left of the grid, as they require the fewest steps from your current location. After gathering the keys, you pass them to the human, who then uses them to unlock both doors and successfully retrieve the gem.
\end{promptbox}
    \caption{Demonstration exemplar component for Tomcat with CP part-6}
    \label{fig:c3-33}
\end{figure}

\newpage
\begin{figure}[H]
    \begin{promptbox}[colback=ncsuLightBlue!60]{robot}{Demonstration Exemplars: CP Problem-2 (Example)}
    \textbf{Example from this grid configuration.} \\
\textbf{Human Action:} The human moves to the right from their current position at (5,0) to (5,4), where they provide the instruction before continuing their movement. \\
\textbf{Instruction:} Can you pass me the red keys? \\
\textbf{Type:} Clear. Based on the instruction and the grid, I infer that the human wants to collect the gem at (10,0) or at (10,8), which is beyond two red doors. Since the human needs two red keys to unlock the doors and reach the gem, the instruction is clear.\\
\textbf{Response:} There are three red keys, three blue keys, two red doors, and two blue doors on the grid. The instruction asks to collect red keys to retrieve the gem at (10,0) or (10,8). I will collect the two closest red keys at (1,1) and (1,3), as this minimizes the steps required to retrieve them. Since the human continues moving after the instruction, they are likely heading downward from (5,4) to (7,4), adjacent to the red door at (8,4). I will pass the keys to the human at either (5,4) or (7,4), depending on their movements, enabling them to unlock the red doors at (8,4) and (9,4) to access the desired gem. \\
\textbf{Actions:} \\
1) Collect: red\_key at (1,1).\\
2) Collect: red\_key at (1,3). \\
3) Pass: red\_keys to the human's future position either at (5,4) or (7,4). \\
4) Unlock: human unlocks the Red\_doors at (8,4) or (9,4).\\
5) Retrieve: human retrieves gem at either (10,0) or (10,8). \\
\texttt{'''}
\end{promptbox}
    \caption{Demonstration exemplar component for Tomcat with CP part-7}
    \label{fig:c3-34}
\end{figure}

\FloatBarrier

\newpage
\subsection{Demonstration: Fs-CoT}
\label{sec:DFs}
\begin{figure}[H]
    \begin{promptbox}[colback=ncsuLightBlue!60, fontupper=\color{black}]{robot}{Demonstration Exemplars: Fs-CoT Exemplar-1 (k=7)}
     Use the following problems and examples, delimited by triple quotes, to understand how to generate the appropriate type, response, and actions for each instruction and human actions.\\
\texttt{'''} \\
    \textbf{Human Action:} The human moves downward from their current position at (6,4) to (8,4), where they provide the instruction. \\
\textbf{Instruction:} Can you pass me the blue key? \\
\textbf{Type:} Unclear. Based on the human's movement, instruction and the current grid configuration, I infer that the human moved to (8,4) to collect the gem at (8,9) located beyond the blue door and red door at (8,6) and (8,8) respectively since they asked for a blue key. To retrieve the gem the human needs to unlock both these doors. Since the human does not mention passing the red key, the instruction is unclear.\\
\textbf{Response:} The instruction asks to collect a blue key and based on the human movements and the grid configuration the desired gem they want to collect is placed at (8,9). But this gem is beyond the blue door and red door at (8,6) and (8,8) respectively. So I will collect one blue key and one red key located at (3,8) and (5,8). To minimize movement, I will first pick up the blue key at (3,8) followed by the red key at (5,8), as this strategy reduces the number of steps from my current position at (1,8) to the human at (8,4). Once both keys are collected, I will pass them to the human, allowing them to unlock blue door and red door at (8,6) and (8,8) respectively and retrieve the gem at (8,9). \\
\textbf{Actions:} \\
1) Collect: blue\_key at (3,8).\\
2) Collect: red\_key at (5,8). \\
3) Pass: blue\_key and red\_key to the human at (8,4). \\
4) Unlock: human unlocks the Blue\_door at (8,6) and the Red\_door at (8,8).\\
5) Retrieve: human retrieves gem at (8,9).
\end{promptbox}
    \caption{Demonstration exemplar component for Tomcat with Fs-CoT part-1}
    \label{fig:c3_2-1}
\end{figure}

\newpage
\begin{figure}[H]
    \begin{promptbox}[colback=ncsuLightBlue!60]{robot}{Demonstration Exemplars: Fs-CoT Exemplar-2 (k=7)}
    \textbf{Human Action:} The human moves left from their current position at (0,5) to (0,3), where they provide the instruction before continuing their movement. \\
\textbf{Instruction:} On my way to pick up the blue key, can you find a yellow key? \\
\textbf{Type:} Clear. Based on the human's movement, instruction and the current grid configuration, I infer that the human moved to (0,3) to pick up the blue key at (1,0) and requested the yellow key to retrieve the gem at (2,10) located beyond both the blue and yellow doors at (4,7) and (3,10) respectively. Since both the keys are required to retrieve the gem at (2,10), the instruction is clear.\\
\textbf{Response:} The instruction asks to collect a yellow key while the human is picking up a blue key, so the desired gem the human wants to collect is placed beyond a blue and a yellow door. To get the gem at (2,10) the human needs to go through a blue door at (4,7) and a yellow door at (3,10), thus the gem at (2,10) must be the desired gem. Since the human is picking up the blue key at (1,0) based on their movement towards the left direction, I will collect the yellow key. The human after collecting the blue key will move to the blue door to unlock the blue door at (4,7), thus their future position can be (4,10) as they will keep moving towards the gem at (2,10). To minimize movement, I will pick up the yellow key at (8,10), as this strategy reduces the number of steps from my current position at (10,9). Once the key is collected, I will pass it to the human at (4,10) or their future position, after they unlock the blue door at (4,7). The human can then unlock the yellow door at (3,10) to retrieve the gem at (2,10). \\
\textbf{Actions:} \\
1) Collect: yellow\_key at (8,10).\\
2) Unlock: human unlocks Blue\_door at (4,7).\\
3) Pass: yellow\_key to the human at (4,10) or their future position.\\
4) Unlock: human unlocks the Yellow\_door at (3,10).\\
5) Retrieve: human retrieves gem at (2,10).
\\
\end{promptbox}
    \caption{Demonstration exemplar component for Tomcat with Fs-CoT part-2}
    \label{fig:c3_2-2}
\end{figure}

\newpage
\begin{figure}[H]
    \begin{promptbox}[colback=ncsuLightBlue!60]{robot}{Demonstration Exemplars: Fs-CoT Exemplar-3 (k=7)}
\textbf{Human Action:} The human moves right from their current position at (3,3) to (3,5), where they provide the instruction before continuing their movement. \\
\textbf{Instruction:} I'll get the blue key. Can you pick up a red key? \\
\textbf{Type:} Clear. Based on the human's movement, instruction and the current grid configuration, I infer that the human moved to (3,5) to pick up the blue key at (3,8) and requested the red key to collect either the gem at (11,8) or the gem at (11,12) located behind the red and blue doors at (9,10) and (10,10) respectively. Since the red and blue keys are required for the human to reach either one of their desired gems, the instruction is clear.\\
\textbf{Response:} The instruction asks to collect a red key while the human is picking up a blue key, so the desired gem the human wants to collect is placed beyond a blue and a red door. To get the gem at either (11,8) or (11,12) the human needs to go through a red door at (9,10) and a blue door at (10,10), thus the gems at either (11,8) or (11,12) must be one of the desired gems. Since the human is picking up the blue key at (3,8) based on their movement towards the right direction, I will collect the red key. The human after collecting the blue key at (3,8) will move towards the red and blue doors at (9,10) and (10,10), thus their future position can be at (8,10), located just before both the doors. To minimize movement, I will pick up the red key located at (8,6), as this strategy reduces the number of steps from my current position at (6,3). Once the key is collected, I will pass it to the human at (8,10) or their future position, allowing them to unlock the both the doors and retrieve either one of the gems. \\
\textbf{Actions:} \\
1) Collect: red\_key at (8,6).\\
2) Pass: red\_key to the human at (8,10) or their future position.\\
3) Unlock: human unlocks Red\_door at (9,9) and Blue\_door at (10,10).\\
4) Retrieve: human retrieves a gem at either (11,8) or (11,12).\\
\end{promptbox}
    \caption{Demonstration exemplar component for Tomcat with Fs-CoT part-3}
    \label{fig:c3_2-3}
\end{figure}

\newpage
\begin{figure}[H]
    \begin{promptbox}[colback=ncsuLightBlue!60]{robot}{Demonstration Exemplars: Fs-CoT Exemplar-4 (k=7)}
    \small
\textbf{Human Action:} The human moves right from their current position at (5,3) to (5,6) and then moves downward to (6,6), where they provide the instruction before continuing their movement. \\
\textbf{Instruction:} Can you go get the red key? \\
\textbf{Type:} Unclear. Based on the human's movement, instruction and the current grid configuration, I infer that the human moved to (6,6) to pick up the blue key at (8,6) even though it was not mentioned in the instruction, and requested the red key to collect either the gem at (8,0) or (8,4) located behind the red and blue doors at (6,3) and (7,3) respectively. Since the human does not mention anything about a blue key, the instruction is unclear.\\
\textbf{Response:} The instruction asks to collect a red key while the human is likely picking up a blue key, so the desired gem the human wants to collect is placed beyond a blue and a red door. To get the gem at either (8,0) or (8,4) the human needs to go through a red door at (6,3) and a blue door at (7,3), thus the gem at either (8,0) or (8,4) must be the desired gem. Since the human is likely picking up the blue key at (8,6) based on their movement towards the downward direction, I will collect the red key. The human after collecting the blue key at (8,6) will move towards the red and blue doors at (6,3) and (7,3), thus their future position can be at (5,3), located just before both the doors. There are two red keys on the grid at (0,7) and at (3,7). Since the red key at (3,7) is blocked by a blue door, I will collect the red key at (0,7) such that it takes minimal steps. Once the red key is collected, I will pass it to the human at (5,3) or their future position, allowing them to unlock the doors and collect either one of the gems. \\
\textbf{Actions:} \\
1) Collect: red\_key at (0,7).\\
2) Pass: red\_key to the human at (5,3) or their future position.\\
3) Unlock: human unlocks Red\_door at (6,3) and Blue\_door at (7,3).\\
4) Retrieve: human retrieve a gem at either (8,0) or (8,4).
\end{promptbox}
    \caption{Demonstration exemplar component for Tomcat with Fs-CoT part-4}
    \label{fig:c3_2-4}
\end{figure}
\newpage
\begin{figure}[H]
    \begin{promptbox}[colback=ncsuLightBlue!60]{robot}{Demonstration Exemplars: Fs-CoT Exemplar-5 (k=7)}
\textbf{Human Action:} The human moves upward from their current position at (3,5) to (0,5), where they provide the instruction before continuing their movement. \\
\textbf{Instruction:} I will pick up this red key. Can you find a yellow one? \\
\textbf{Type:} Unclear. Based on the human's movement, instruction and the current grid configuration, I infer that the human moved to (0,5) to pick up the red key at (0,0) and requested for a yellow key to collect either one of the gems at (7,2), (7,4), or (7,6), respectively. Because to reach these gems it requires to unlock a yellow door first and then a red door. Since there are many possibilities to reach gems which will require both yellow and red keys, it is impossible to infer which gem the human wants, making this an unclear instruction.\\
\textbf{Response:} The instruction asks for a yellow key while the human is picking up a red key, so the desired gem the human wants to collect is placed beyond a yellow and a red door. To get the gems at (7,2), (7,4), or (7,6), the human needs to go through a yellow and a red door at. Therefore, based on the human's future position their desired gem can be any one of them. A red door at (6,2) is placed after a yellow door at (5,2). Two red doors at (6,4) and (6,6) are placed after a yellow door at (4,5). Therefore, the human's future position will be before the two yellow doors at (4,2) or (3,5), or it can be someplace else. To minimize movement, I will pick up the yellow key at (9,0) to reduce the number of steps from my current position at (9,5) to the human at (3,5) or (4,2) or their future position. Once I collect the yellow key, I will pass it to the human at (3,5) or (4,2), allowing them to unlock the doors and receive the gem. \\
\textbf{Actions:} \\
1) Collect: yellow\_key at (9,0).\\
2) Pass: yellow\_key to the human at (3,5) or (4,2) or their future position.\\
3) Unlock: human unlocks the Yellow\_door at (5,2) or (4,5) and the Red\_door at (6,2) or (6,4), (6,6).\\
4) Retrieve: human retrieves the gem at either (7,2), (7,4) or (7,6).

\end{promptbox}
    \caption{Demonstration exemplar component for Tomcat with Fs-CoT part-5}
    \label{fig:c3_2-5}
\end{figure}
\newpage
\begin{figure}[H]
    \begin{promptbox}[colback=ncsuLightBlue!60]{robot}{Demonstration Exemplars: Fs-CoT Exemplar-6 (k=7)}
\textbf{Human Action:} The human moves down from their current position at (4,4) to (5,4), and then proceeds to move left to collect the red key at (5,0). After collecting the key, they continue moving right until they reach (5,2), where they provide the instruction before continuing their movement. \\
\textbf{Instruction:} Can you help me unlock the blue door there? \\
\textbf{Type:} Clear. Based on the human's movement, instruction and the current grid configuration, I infer that the human picked up the red key at (5,0), moved to (5,2), and requested me to unlock blue door to retrieve the gem at (0,0), which is located beyond the blue and red doors at (2,4) and (2,0) respectively. To retrieve the gem the human needs to pass both these doors. Since the human picked up the red key and I just have to collect the blue key to unlock the blue door blocking the human's path to the gem, the instruction is clear.\\
\textbf{Response:} The instruction asks to unlock the blue door while the human has already picked up the red key at (5,0). The human's desired gem based on their actions and the grid configuration must be beyond a blue and a red door. The gem at (0,0) is the desired gem as it is placed beyond a blue door at (2,4) and a red door at (2,0). To minimize movement, I will collect the only accessible blue key at (0,8) and then move to unlock the blue door at (2,4) which is blocking the path to the gem at (0,0). The human can then unlock the red door at (2,0) to retrieve the gem at (0,0). \\
\textbf{Actions:} \\
1) Collect: blue\_key at (0,8).\\
2) Unlock: Blue\_door at (2,4).\\
3) Unlock: human unlocks Red\_door at (2,0).\\
4) Retrieve: human retrieves gem at (0,0).
\end{promptbox}
    \caption{Demonstration exemplar component for Tomcat with Fs-CoT part-6}
    \label{fig:c3_2-6}
\end{figure}

\newpage
\begin{figure}[H]
    \begin{promptbox}[colback=ncsuLightBlue!60]{robot}{Demonstration Exemplars: Fs-CoT Exemplar-7 (k=7)}
\textbf{Human Action:} The human moves left from their current position at (6,2) to (6,1) and then upward to (5,1). Upon reaching (5,1), adjacent to the red door at (4,1), they provide the instruction. \\
\textbf{Instruction:} Can you unlock this red door? \\
\textbf{Type:} Unclear. Based on the human's movement, instruction and the current grid configuration, I infer that the human moved to (5,1) and asked to unlock the red door at (4,1) to retrieve the gem at (1,1) located behind a blue and red door at (2,1) and (4,1) respectively. To retrieve the gem the human needs to have both doors unlocked. However, since there is no mention about unlocking the blue door, the instruction is unclear.\\
\textbf{Response:} The instruction specifies unlocking a red door, indicating that the desired gem must be behind one. Based on the grid configuration, there are two gems on the grid, both blocked by a red and a blue door. Given that the human has moved to (5,1), adjacent to the red door at (4,1), they are likely targeting the gem at (1,1), which lies beyond the adjacent red door. However, the gem at (1,1) is also blocked by a blue door at (2,1), in addition to the red door at (4,1). To assist the human, I will collect both the red and blue keys. To minimize movement, I will first collect the red key at (0,5), followed by the blue key at (2,7), as this approach reduces the total steps from my current position at (2,5). After collecting both keys, I will proceed to unlock the blue door at (2,1) and then the red door at (4,1), enabling the human to access the gem at (1,1). \\
\textbf{Actions:} \\
1) Collect: red\_key at (0,5).\\
2) Collect: blue\_key at (2,7).\\
3) Unlock: Blue\_door at (2,1) and Red\_door at (4,1).\\
4) Retrieve: human retrieves gem at (1,1). \\
\texttt{'''}
\end{promptbox}
    \caption{Demonstration exemplar component for Tomcat with Fs-CoT part-7}
    \label{fig:c3_2-7}
\end{figure}

\end{document}